\pdfoutput=1

\documentclass[11pt]{article}

\usepackage{acl}

\usepackage{times}
\usepackage{latexsym}

\usepackage[T1]{fontenc}

\usepackage[utf8]{inputenc}

\usepackage{microtype}

\usepackage{soul}
\usepackage{xspace}
\usepackage{multirow}
\usepackage{amsmath}
\usepackage{amssymb}
\usepackage{bbding}
\usepackage{graphicx}
\usepackage{textcomp}
\usepackage{pifont}
\usepackage[caption=false]{subfig}
\DeclareMathOperator*{\argmax}{arg\,max}

\def\tasksymbol{$\mathcal{T\mkern-5mu}$\xspace}
\def\md{LMTurk\xspace}
\def\mdr{LMTurker\xspace}
\def\mdrs{LMTurkers\xspace}
\usepackage{xcolor}
\definecolor{blue}{HTML}{4466A5}
\definecolor{green}{HTML}{4E9E5E}
\definecolor{orange}{HTML}{D87B4D}

\def\figref#1{Figure~\ref{fig:#1}}
\def\figlabel#1{\label{fig:#1}\label{p:#1}}

\def\tabref#1{Table~\ref{tab:#1}}

\def\tablabel#1{\label{tab:#1}\label{p:#1}}

\def\secref#1{\S\ref{sec:#1}}
\def\seclabel#1{\label{sec:#1}}

\newcounter{notecounter}
\newcommand{\enotesoff}{\long\gdef\enote##1##2{}}
\newcommand{\enoteson}{\long\gdef\enote##1##2{{
\stepcounter{notecounter}
{\large\bf \hspace{1cm}\arabic{notecounter} $<<<$ ##1: ##2 $>>>$\hspace{1cm}}}}}
\enoteson
\enotesoff

\newcommand{\MF}[1]{}

\title{LMTurk: Few-Shot Learners as Crowdsourcing Workers in
  a Language-Model-as-a-Service Framework}

\author{
  Mengjie Zhao\textsuperscript{†}
  \ \ Fei Mi\textsuperscript{‡}
  \ \ Yasheng Wang\textsuperscript{‡}
  \ \ {\bf Minglei Li\textsuperscript{\normalfont{*}}}\\
  {\bf Xin Jiang\textsuperscript{‡}}
  \ \ {\bf Qun Liu\textsuperscript{‡}}
  \ \ {\bf Hinrich Sch\"{u}tze\textsuperscript{†}}\\
  \textsuperscript{†}CIS, LMU Munich
  \ \ \textsuperscript{‡}Huawei Noah's Ark Lab
  \ \ \textsuperscript{*}Huawei Technologies Co., Ltd.\\
  {\tt {\small mzhao@cis.lmu.de,}}
  {\tt {\small \{mifei2,wangyasheng,jiang.xin,qun.liu\}@huawei.com}}
}

\begin{document}
\maketitle

\begin{abstract}
  Vast efforts have been devoted to
  creating high-performance few-shot learners, i.e.,
  large-scale pretrained language models (PLMs)
  that perform well 
  with little downstream task training data.
  Training PLMs
  has incurred significant cost, but
  utilizing the few-shot learners is still
  challenging
  due to their enormous size.
  This work focuses on a crucial question:
  How to make effective use of these few-shot learners?
  We propose LMTurk, a novel approach that
  treats few-shot learners
  as crowdsourcing workers.
  The rationale is that
  crowdsourcing workers are in fact
  few-shot learners: They are shown
  a few illustrative examples
  to learn about a task and then
  start annotating.
  LMTurk employs few-shot learners
  built upon PLMs
  as workers.
  We show that the
  resulting annotations can be
  utilized to train
  models that
  solve the task well and are small enough to be deployable in
  practical scenarios.
  Active learning is integrated into LMTurk to
  reduce the amount of queries made to PLMs,
  minimizing the computational cost of running
  PLM inference passes.
  Altogether, LMTurk is
  an important step towards making
  effective use of
  current PLMs.\footnote{Resources are available at: \url{github.com/lmturk}}
\end{abstract}

\section{Introduction}

Equipped with prolific linguistic features
\citep{
  liulinguistic,tenney2018what,
  belinkov-glass-2019-analysis,BERTologypaper}
and rich world knowledge \citep{
  petroni-etal-2019-language,poerner-etal-2020-e,
  kassner-etal-2021-multilingual},
large-scale pretrained language models (PLMs) have been shown
to be versatile:
They are now basic
building blocks \citep{Fundationmodelpaper}
of systems solving diverse
NLP tasks in many languages
\citep{wang-etal-2018-glue,
  superglue,
  pmlr-v119-hu20b,
  xu-etal-2020-clue,
  Khashabi2020ParsiNLUAS,
  Park2021KLUEKL,  
  tacl_a_00416}.

Recent work shows that PLMs are effective \emph{few-shot learners}
\citep{GPT3paper,schick2020s,gao-etal-2021-making,Tam2021ImprovingAS}
through \emph{priming} \citep{GPT3paper,Tsimpoukelli2021MultimodalFL}
or \emph{prompting} \citep{lisaprefix,Ptuningpaper,Prompttuningpaper,xlmrprompting}.
Developing
few-shot learners is crucial
because current NLP systems 
require much more data than humans \citep{yin-etal-2020-universal}.
Few-shot learners tend to perform well;
however,
they still fall behind systems
trained with abundant data.  
Furthermore,
the enormous size of
PLMs
hinders their
deployment in
practice.
For example, it is challenging to fit
the 11 billion
T5-XXL \citep{t5paper} model
on a single regular GPU.

\begin{figure}[t]
\centering
\includegraphics[width=\linewidth]{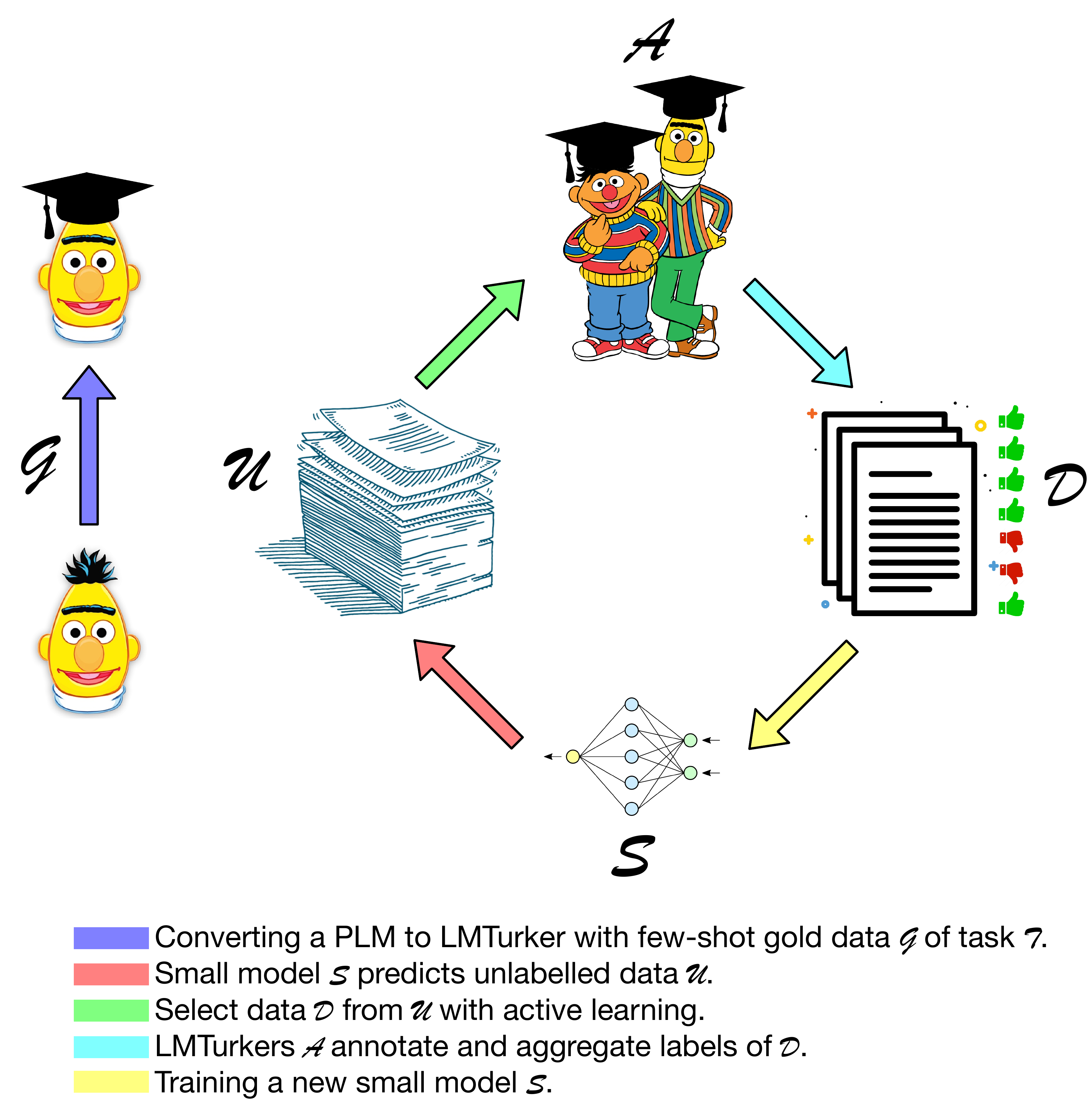}
\caption{
  \md overview; best viewed in color.
  We few-shot adapt PLMs to task $\mathcal{T}$ (left) and
  then use them as crowdsourcing workers in active learning.
  We show that these PLM workers are effective in training a small model
{\cal $\mathcal{S}$} through a customized active learning loop (right).
\md is a novel way to take advantage of large-scale PLMs:
It creates models small enough to be deployed in resource-limited
real-world settings.}
\figlabel{overallframe}
\end{figure}

Our goal in this paper is to devise
methods that make
\emph{more effective use of
  current few-shot learners}.
This is crucial because
an increasing number of
gigantic few-shot learners are trained;
how to use them effectively is thus
an important question.
In particular,
we want an alternative to
hard-to-deploy huge models.
At the same time, we want to take full advantage of the PLMs'
strengths:
Their versatility
ensures 
wide applicability across tasks;
their vast store of knowledge
about language and the world
(learned in pretraining)
manifests in the
data efficiency of
few-shot learners, reducing
labor and time consumption
in data annotation.

In this work, we propose
\textbf{\md},
\textbf{L}anguage
\textbf{M}odel as
mechanical \textbf{Turk}.
Our basic idea (see \figref{overallframe}) is that, for an NLP task \tasksymbol,
\emph{we treat few-shot learners as non-expert workers},
resembling crowdsourcing workers that annotate resources for
human language technology.
We are inspired by the fact that  we can view a
crowdsourcing worker
as a type of few-shot learner:  A few examples
demonstrating $\mathcal{T}$ teach her
enough about $\mathcal{T}$ to
conduct effective annotation.
For example, \citet{snow-etal-2008-cheap} train workers
with a few examples of annotating emotion;
\citet{he-etal-2015-question} conduct short training sessions
for workers before annotation;
\citet{lee2021annotation}  train
workers with
learning curricula.

\citet{snow-etal-2008-cheap} pioneered crowdsourcing
in NLP \citep{howe2006rise,howe2008crowdsourcing}, motivated
by the high cost of TreeBank annotation \citep{marcus-etal-1993-building,
miller-etal-1993-semantic}.
Crowdsourcing organizes
human workers over the Web
to annotate data.
Workers need not be experts to be effective, resulting in
reduced \emph{per-label cost}.
Active learning \citep{hachey-etal-2005-investigating,felder2009active} can be incorporated
\citep{laws-etal-2011-activeb} to further decrease annotation cost, by lowering
\emph{the number of labels}
to be annotated.
\md treats PLM-based
few-shot learners as
non-expert workers that
produce training sets, which are
then used to
train a small
machine learning model
$\mathcal{S}$ specialized for
\tasksymbol.
This
scenario is analogous to
active learning.
We achieve two benefits:
(i) low annotation cost
because humans only need to annotate a few shots of data;
(ii) solving practical
NLP tasks with small models
that are more real-world deployable.

\md
resonates with
\citet{laws-etal-2011-activeb}'s
earlier idea of
combining crowdsourcing and
active learning.
They consider
human workers
as ``noisy annotators''
while we explore the utilization
of modern NLP few-shot learners
(built upon machine learning models)
as workers --
which have the advantage of
being free,
instantly interactive,
fast, responsive, and non-stopping.

Our \textbf{contributions}: (i) We propose \md, a method
that uses few-shot learners as crowdsourcing workers.
\figref{overallframe} shows the overview of \md.
(ii) We vary an array of important design
choices, identifying strengths and weaknesses of \md.
(iii) Unlike much work on active learning  in
a synthetic oracle setting,
we develop methods for handling the
varying quality
of annotation  that does not come from an oracle.
(iv) We extensively evaluate \md on five
datasets, showing that \md can guide a small model
$\mathcal{S}$ to progressively improve on \tasksymbol.
$\mathcal{S}$ can then be deployed in practical scenarios.
(v) This is the
first work
showing that few-shot learners
give rise to effective NLP models
through crowdsourcing and active
learning -- with the benefits of low annotation cost and
practical deployability.

\section{Related Work}
\textbf{Few-shot learners in NLP}.
Significant progress
has been made in
developing \citep{devlin-etal-2019-bert,ELMopaper,XLNetpaper,GPT3paper},
understanding \citep{liulinguistic,tenney2018what,
  belinkov-glass-2019-analysis,hewitt-liang-2019-designing,
  hewitt-manning-2019-structural,zhao-etal-2020-quantifying,
  BERTologypaper},
and utilizing \citep{
  adapterpaper,maskingpaper,
  GPT3paper,lisaprefix,
  PETpaper,Prompttuningpaper,mi2021cins}
PLMs.
\citet{GPT3paper},
\citet{PETpaper},
and \citet{Ptuningpaper}
show that PLMs can serve as
data-efficient few-shot learners,
through priming or prompting \citep{liu2021pre}.
For example, GPT3 achieves
near state-of-the-art
performance on COPA \citep{roemmele2011choice}
with only 32 annotated data.

However,
little to no
work discusses
or explores the
actual \emph{practical utility}
of these few-shot learners.
We aim to develop
effective methods of
utilizing them
in practical scenarios.

\textbf{Crowdsourcing} has a
long history
in human language technology
\citep{crowdsourcingagg,callison-burch-2009-fast,
  trautmann2020fine}; specialized
workshops were
organized
\citep{crowdsourws,emnlp-2019-aggregating}.
It has numerous applications
\citep{crowdsourcingsurvey},
but we focus on its
application as
voting systems.
To reduce \emph{per-label} cost,
crowdsourcing organizes
non-expert human workers
distributed across the Web for annotation,
instead of employing linguistic experts
\citep{jamison-gurevych-2015-noise,
  bhardwaj-etal-2019-carb,
  nangia-etal-2021-ingredients}.
\citet{snow-etal-2008-cheap}
show that
averaging ten crowdsourced labels
matches an expert-level label for
recognizing textual entailment \citep{rtedata}.
\citet{paun-etal-2018-comparing}
show that incorporating structure in
annotation models is important.
Measuring label disagreements
is also crucial \citep{dumitrache2021empirical}.

\md
utilizes NLP few-shot learners
as non-expert workers.
The few-shot training data
can be viewed as the examples
shown
to humans before annotating.
The process is
free, fast, responsive, and non-stopping.

\textbf{Active learning} (AL; \citet{cohn1996active,settles2009active})
strives to reduce \emph{the number of examples}
to be annotated
via identifying 
informative examples
with acquisition functions.
\citet{settles-craven-2008-analysis}
evaluate AL algorithms
for sequence labeling. 
\citet{ALCNN,shen-etal-2017-deep,siddhant-lipton-2018-deep}
apply  AL to deep neural networks.
\citet{simpson-gurevych-2018-finding} devise
a scalable Bayesian
preference learning method for
identifying convincing arguments.
\citet{lee-etal-2020-empowering} propose to consider
user feedback in AL systems.
\citet{ein-dor-etal-2020-active} explore AL
for BERT.
\citet{schroder2020survey} review
text classification with AL.
\citet{liang-etal-2020-alice,margatina2021active}
integrate contrastive learning into AL.
\citet{zhang-plank-2021-cartography-active} identify
examples with datamap \citep{datasetmap}.

We incorporate AL in
\md to
reduce the amount of examples
to be annotated by PLMs,
reducing the
computational cost
of running several
inference passes.
This contributes to a more
environmentally
friendly \citep{strubell-etal-2019-energy,
  schwartz2020green,patterson2021carbon} 
scenario.

Perhaps  closest to our work,
\citet{yoo2021gpt3mix}
conduct data augmentation
via priming GPT3 and
\citet{wang2021want} mix
human- and GPT3-annotated data, focusing
on cost analysis.
GPT3 is utilized in
a Language-Model-as-a-Service form by OpenAI,
which is not free.\footnote{\url{https://beta.openai.com/pricing}}
Also, strategies of priming GPT3 may not
generalize well to other PLMs.
For example, priming strategies have to adapt to GPT3's maximum
sequence length.  However, maximum sequence length -- as a
hyperparameter -- could vary across PLMs.
In this work,
we prompt
publicly available free PLMs.
This also makes the process more flexible; for example,
the PLM can be updated
with gradient descent.

\section{\md}
\subsection{Training few-shot learners}
We first adapt a PLM to
task $\mathcal{T}$ with
a few-shot human-labeled gold dataset
$\mathcal{G}$ = \{$\mathcal{G}_{train}; \mathcal{G}_{dev}$\}
of \tasksymbol.
This procedure mimics
one of the initial but crucial
steps in crowdsourcing:
A few example annotations
are shown to the workers,
demonstrating \tasksymbol; workers
learn about the task and then start
annotating \citep{
snow-etal-2008-cheap,
he-etal-2015-question,
roit-etal-2020-controlled,
trautmann2020fine,
lee2021annotation}.

We achieve this adaptation
through P-Tuning \citep{Ptuningpaper}.
Taking movie review classification
as an example,
the goal is to associate
a binary label $y$ from \{-1, +1\}
to an input sentence
$\mathbf{x} = (x_1, ..., x_n)$ where
$x_i$ refers to a token.
Unlike finetuning and
its variants \citep{devlin-etal-2019-bert,adapterpaper,maskingpaper}
that train a classifier head,
P-Tuning reformulates a sentence into a
cloze-style query;
the PLM is then requested to
respond to the query
with an answer selected from
a list of candidates.
Concretely, an input pair

\begin{itemize}
\item[] \small{($\mathbf{x}$, y) = (``watching it leaves you giddy.'', -1)}
\end{itemize}

\noindent is reformulated to:

\begin{itemize}
\item[] \small{``\underline{[v]} watching it leaves you giddy. \underline{It} \underline{is} [MASK] \underline{.}''}
\end{itemize}

\noindent in which the \underline{underlined} tokens are
prompting words that give the model
a hint about \tasksymbol.
``[v]'' -- whose trainable embedding vector is
randomly initialized -- is a
prompting token injecting extra free parameters.
The PLM is then requested to pick a word
from \{``bad'', ``good''\} to
fill in the position of ``[MASK]''.
A mapping
\{``bad''$\,\to\,$-1, ``good''$\,\to\,$+1\}
is used to
transform the selected answer
to a label such that standard evaluation
measures like accuracy can be computed.
Prompting has been shown to effectively
adapt a PLM to $\mathcal{T}$
\emph{with only a few annotations};
see \citep{liu2021pre} for a
comprehensive review of prompting.
We refer to a PLM adapted to $\mathcal{T}$ as
an \textbf{\mdr} $A$.

We select
prompting words and mappings based on the
small development set $\mathcal{G}_{dev}$.
\secref{promptingdetails} provides
details on prompting and datasets.

\subsection{Aggregating annotations}
\seclabel{subsec:aggregateannotations}
Individual workers are subject to annotation biases
\citep{snow-etal-2008-cheap};
therefore, crowdsourcing often collects labels from
several workers \citep{crowdsourcingsurvey}
for an
example $\mathbf{x}$ and then
aggregates them
for quality control
\citep{crowdsourcingagg}.
It is straightforward to obtain
a group of \mdrs
$\mathcal{A}=\{A_1, ..., A_k\}$,
by adapting the PLM to $\mathcal{T}$
with $k$ different prompts.
A querying sentence $\mathbf{x}$
is then annotated by every \mdr,
resulting in a list of labels
$\mathbf{y} = [y_1, ..., y_k]$.
We evaluate different methods
aggregating $\mathbf{y}$ to a single
label $\hat{y}$.

\textbf{BestWorker}.
Among the $k$ \mdrs,
we pick the one
performing best on the dev set $\mathcal{G}_{dev}$.

\textbf{MajorityVoting}.
We select the
most frequent
label in
$\mathbf{y} = [y_1, ..., y_k]$
as $\hat{y}$.

To estimate
an \mdr's confidence on label $y_i$,
we compare the
logits\footnote{
  Calibration can be conducted to
  further improve the estimation
  \citep{guo2017calibration}.
  We leave this to future work.
} computed by the  PLM:
$${\scriptstyle y_i \ =\  \argmax(\text{logit}(y^1),...,\ \text{logit}(y^N)) ,}$$
\noindent where
$N$ refers to the label set size,
e.g., $N$=2 for $y$ from \{-1, +1\}.
We then can evaluate several methods of
aggregating annotations according to 
PLM logits.

\textbf{LogitVoting}.
We average the logits from all  $k$ \mdrs
$\{A_1, ..., A_k\}$
to compute $\hat{y}$:
$${\scriptstyle \hat{y}\ =\ \argmax(\frac{1}{k}\sum_{i=1}^k\text{logit}(y^1_i),...,\frac{1}{k}\sum_{i=1}^k\text{logit}(y^N_i))}.$$

\textbf{WeightedLogitVoting}.
We use \mdrs' performance on $\mathcal{G}_{dev}$ 
to weight their logits and then aggregate the
predictions:
\begin{eqnarray*}
&{\scriptstyle \hat{y}\ =\ \argmax(\sum_{i=1}^kw_i\text{logit}(y^1_i),...,\sum_{i=1}^kw_i\text{logit}(y^N_i))} \\
&{\scriptstyle w_i\ =\ f(A_i, \mathcal{G}_{dev}) / \sum_{i=1}^k f(A_i, \mathcal{G}_{dev})}  
\end{eqnarray*}

\noindent where $f(A_i, \mathcal{G}_{dev})$ is the
performance of the 
$i$th \mdr $A_i$ on $\mathcal{G}_{dev}$.

We collect and aggregate annotations
from five \mdrs, 
i.e., we use $k$=5 in our experiments.

\subsection{Training a small model $\mathcal{S}$}
\seclabel{methodtrainingsmall}
After adapting \mdrs to $\mathcal{T}$
through prompting with the
few-shot gold dataset $\mathcal{G}$,
we next train a small model
$\mathcal{S}$
specialized to solve \tasksymbol.
Though large PLMs
are versatile and strong performers,
training and inference 
are faster and more
efficient for small models: They are more
deployable in resource-restricted
scenarios, e.g., on
edge devices \citep{jiao-etal-2020-tinybert}.

We mimic
pool-based active learning (AL; \citet{settles2009active})
to train $\mathcal{S}$.
The motivation is to
avoid frequent 
querying of \mdrs
$\mathcal{A}$ because
energy and time consumption
of PLM inference  is
costly
when
the number of queries
and $|\mathcal{A}|$ are large.

Concretely, pool-based AL
assumes a large
collection of unlabeled data
$\mathcal{U}$ = \{$\mathbf{x}_1,...,\mathbf{x}_M$\}
for \tasksymbol.
$\mathcal{S}$ is first trained with
$\mathcal{G}$ = \{$\mathcal{G}_{train}; \mathcal{G}_{dev}$\}.
After that,  a group of examples
$\mathcal{B}$ from $\mathcal{U}$ is sampled (c.f. \secref{alacquisition}),
which \mdrs  annotate.
Next, the annotated and aggregated examples 
$\mathcal{B'}$ are concatenated with
$\mathcal{G}$ to train $\mathcal{S}$.
The procedure is repeated iteratively,
such that the training data for
$\mathcal{S}$ keeps expanding.
We denote as $\mathcal{S}^j$ the model trained after
the $j$th iteration.
Note that $\mathcal{S}$ is trained from
scratch in each iteration \citep{cohn1994improving}.

\subsubsection{AL acquisition function}
\seclabel{alacquisition}
At the beginning of the $j$th iteration, 
a straightforward strategy
of sampling $\mathcal{B}$ from $\mathcal{U}$
is \textbf{random sampling}.
AL promises to select
a more informative $\mathcal{B}$
such that the trained
$\mathcal{S}^j$ performs better,
under the same budget.
These strategies -- or
\emph{acquisition functions} -- 
rely on $\mathcal{S}^{j-1}$,
i.e.,
$\mathcal{S}$
from the previous iteration:
$\mathcal{S}^{j-1}$ is employed to infer
$\mathcal{U}$
to obtain labels and logits
$\mathcal{P}^{j-1} = \{(y_1,\mathbf{c}_1),...,(y_M,\mathbf{c}_M)\}$;
each $\mathbf{c}_i$ contains
the logits of the  $N$ labels;
$y_i=\argmax(\mathbf{c}_i)$.
We explore
two common
AL acquisition functions:
Entropy \citep{roy2001toward}
and
LeastConfident \citep{lewis1994sequential}.

\textbf{Entropy} selects
from $\mathcal{P}^{j-1}$
examples with the
largest prediction entropy,
computed using $\mathbf{c}$.
Large entropy of an example $\mathbf{x}$
implies that $\mathcal{S}^{j-1}$
is unsure about which label to select;
$\mathbf{x}$ is then a query made to \mdrs
to obtain its label $\hat{y}$.
$(\mathbf{x}, \hat{y})$ is subsequently
added to $\mathcal{G}_{train}$ for training $\mathcal{S}^{j}$.

\textbf{LeastConfident}
selects from $\mathcal{P}^{j-1}$
examples for which the maximum
logit in $\mathbf{c}$ is the smallest.
Selected examples are then annotated 
and added to $\mathcal{G}_{train}$
for training $\mathcal{S}^{j}$.

Our AL setup is fairly standard, both in terms of
acquisition functions
and iterative enlargement by new
sampled data $\mathcal{B}$ at iteration
$j$ labeled by $\mathcal{S}^{j-1}$.

\subsubsection{Considering annotation quality}
\seclabel{instancetresholdingintro}
As in any realistic AL scenario, annotations are not perfect:
\mdrs
do not score perfectly on \tasksymbol.
As a result,
\emph{annotation quality of \mdrs
needs to be taken into consideration
before training} $\mathcal{S}^{j}$.
Denoting the training data of
$\mathcal{S}^{j}$ as $\mathcal{D}^{j}$,
we explore
a strategy of processing
$\mathcal{D}^{j}$, based on 
\mdr logits $\mathbf{l}$.

\textbf{InstanceTresholding}.
We preserve examples
$(\mathbf{x}, \hat{y}, \mathbf{l}) \in \mathcal{D}^{j}$
for which entropy computed on
$\mathbf{l}$
is smallest.
$\mathcal{G}^{train}$ is always preserved because it
is human-labeled gold data.
Note that this is
different from the strategy of
sampling $\mathcal{B}$, where
we select from
$\mathcal{P}^{j-1}$
examples to which $\mathcal{S}^{j-1}$
is most unsure (computed with $\mathbf{c}$).
We evaluate\footnote{
Motivated by \citet{wang-etal-2017-instance},
we also investigate the effectiveness of
weighting training examples. However,
we do not observe noticeable improvements of
task performance. We list more details  in
Appendix \secref{appendix:instanceweighting}.}
the effectiveness
of processing $\mathcal{D}^j$
before training $\mathcal{S}^j$
in \secref{poolfilter}.

\subsection{Summary of \md}
\md can be viewed as
intermediate  between
self training \citep{yarowsky-1995-unsupervised,abney-2004-understanding,lee2013pseudo,mi2021self}
and AL.
Unlike self training, \md employs
\emph{external} models provide labels to
$\mathcal{S}$.
Different from the artificial setup used in many AL
experiments, the provided labels
\emph{do not have oracle quality};
so $\mathcal{S}$
must use the annotations more carefully.
We next conduct experiments
investigating the effectiveness of \md.

\section{Datasets and Setup}
\subsection{Dataset}
We evaluate \md on five datasets:
Binary (SST2) and
fine-grained (five classes)
sentiment classification (SST5)
with the Stanford Sentiment TreeBank
\citep{socher-etal-2013-recursive};
news article topic classification with the
AG's News Corpus (AGNews; \citet{agdataset});
recognizing textual entailment (RTE; \citet{rtedata});
assessing linguistic acceptability
(CoLA; \citet{warstadt-etal-2019-neural}).
Appendix \secref{appendix:checklist} reports
dataset statistics.
SST2/SST5 and AGNews are widely used
in crowdsourcing and AL \citep{
  laws-etal-2011-activeb,
  ein-dor-etal-2020-active,
  margatina2021active,
  zhang-plank-2021-cartography-active}.
RTE and CoLA
assess
the models' ability to understand
textual entailment
and linguistic phenomena -- as opposed to
text categorization.
We report Matthew’s correlation coefficient
for CoLA and accuracy for the others \citep{wang-etal-2018-glue}.

\textbf{Few-shot datasets}.
Recall \md uses a small
human-annotated dataset
$\mathcal{G} = \{\mathcal{G}_{train}; \mathcal{G}_{dev}\}$.
Denoting $n$ as the number of shots \emph{per class},
we sample $\mathcal{G}^n_{train}$ and $\mathcal{G}^n_{dev}$
for each of $n \in \{8, 16, 32\}$.
For SST2, RTE, and CoLA, we use the train and dev
sets of  GLUE \citep{wang-etal-2018-glue};
$\mathcal{G}^n_{train}$ and $\mathcal{G}^n_{dev}$ are
sampled from the train set;
the dev set is used as the test set.
For SST5 and AGNews, we use
the official datasets;
$\mathcal{G}^n_{train}$ 
($\mathcal{G}^n_{dev}$) is sampled from the
train (dev) set;
we report  performance on the test set.
We repeat the sampling process
with three random seeds.

\subsection{Training setup}
\seclabel{promptingdetails}
\citet{GPT3paper} show that
large model size is necessary
for strong few-shot performance.
We use ALBERT-XXLarge-v2
\citep{albert} -- of size 223M parameters --
as our large PLM, which is
adapted to be an \mdr $A$ of
$\mathcal{T}$ with $\mathcal{G}$.
With parameter reuse, ALBERT-XXLarge-v2 
outperforms
larger models
like the 334M
BERT-large \citep{devlin-etal-2019-bert}.
In contrast,
$\mathcal{S}$
must be small
to be deployable
in practical scenarios.
We use TinyBERT-General-4L-312D
\citep{jiao-etal-2020-tinybert},
which has 14.5M parameters.

We train -- with prompting --
the large PLM with $\mathcal{G}$
for 100 batch steps using  batch size 16,
AdamW \citep{adamw} and 
learning rate
5e-4 with linear decay.
We prompt the large PLM five times
to obtain five \mdrs;
Appendix \secref{appendix:promptingdetails}
shows
prompting details.
At each iteration,
we finetune  $\mathcal{S}$
for 20 epochs using  batch size 32,
Adam \citep{adampaper} and 
learning rate
5e-5.
Each experiment is run
with three different random seeds.
We use PyTorch \citep{torch} and  HuggingFace
\citep{wolf-etal-2020-transformers}.

\begin{table}[t!]
\scriptsize\centering\renewcommand{\arraystretch}{1.3}\setlength{\tabcolsep}{3pt}
\begin{tabular}{c|c|c|c}
      &\citet{PETpaper,schick2020s} & \citet{gao-etal-2021-making}   & Ours        \\ \hline
SST2  &n/a                          & 93.0$\pm$0.6    &         93.08$\pm$0.62  \\
SST5  &n/a                          & 49.5$\pm$1.7    &         46.70$\pm$0.93 \\
RTE   &69.8                         & 71.1$\pm$5.3    &         70.88$\pm$1.70  \\
AGN.  &86.3$\pm$0.0                 &  n/a            &         87.71$\pm$0.07 \\
CoLA  &n/a                          & 21.8$\pm$15.9   &         19.71$\pm$1.89  \\
\end{tabular}
\caption{
  \mdrs achieve comparable few-shot performance with the
  literature. We refer to \emph{PET} results in
  \citet{PETpaper,schick2020s} and results
  of \emph{Prompt-based FT (auto) + demonstrations} 
  in \citet{gao-etal-2021-making}.
}
\tablabel{fewshotperfcomp}
\end{table}


\begin{figure}[t]
  \centering
	\subfloat{
\includegraphics[width=.8\linewidth,height=0.25\textwidth]{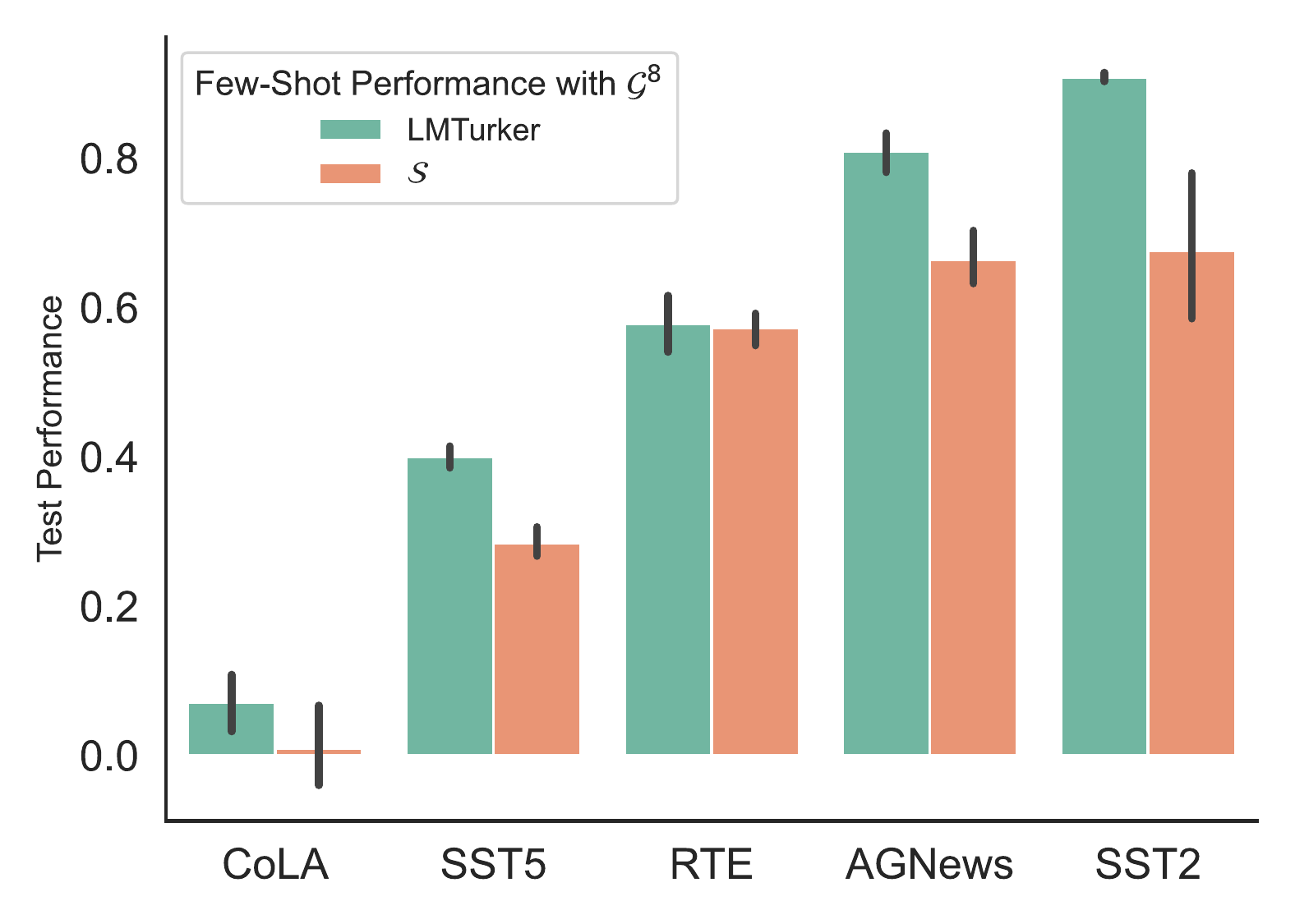}
      }\\
      \subfloat{
 \includegraphics[width=.8\linewidth,height=0.25\textwidth]{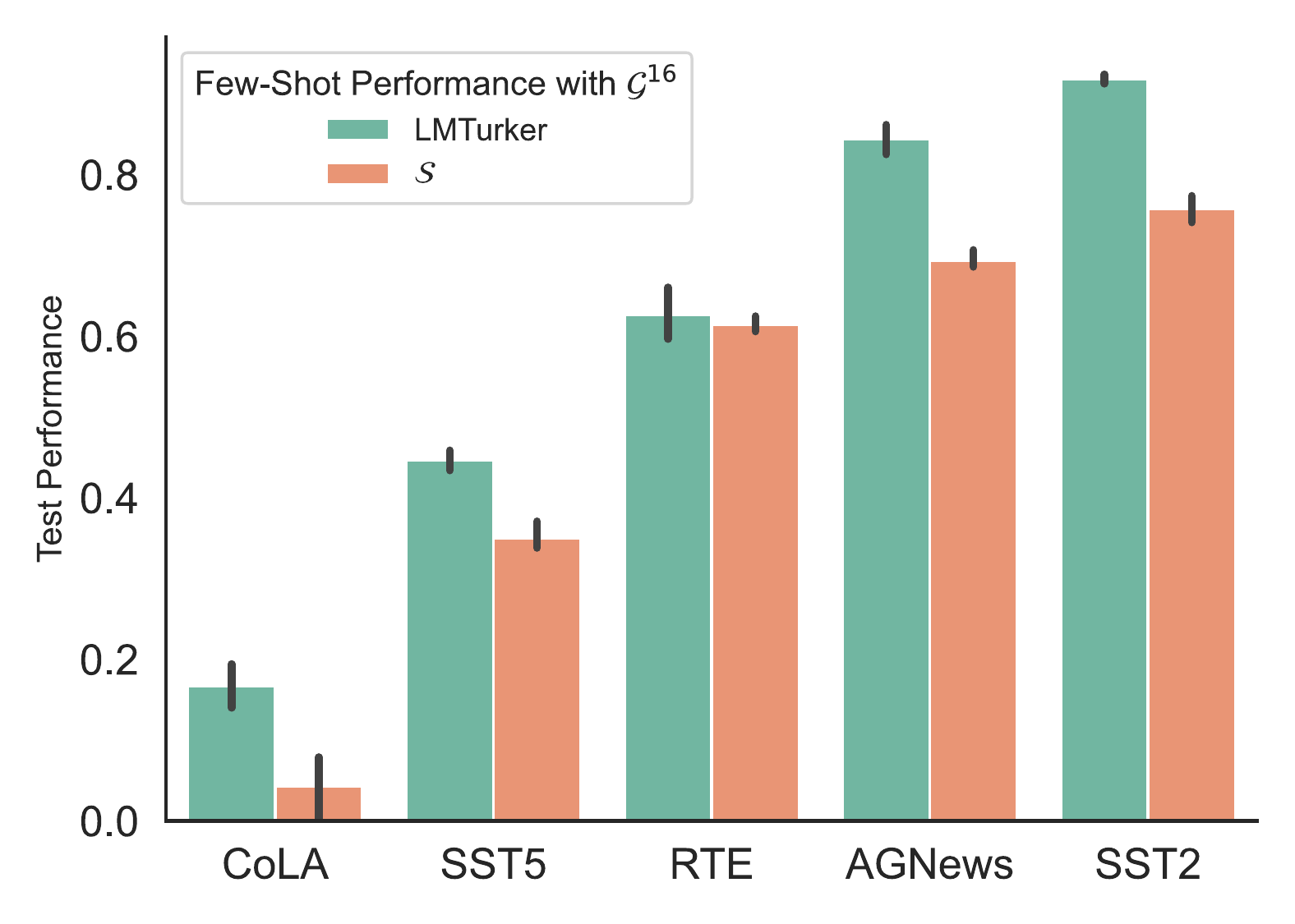}
       }
       \\
       \subfloat{
 \includegraphics[width=.8\linewidth,height=0.25\textwidth]{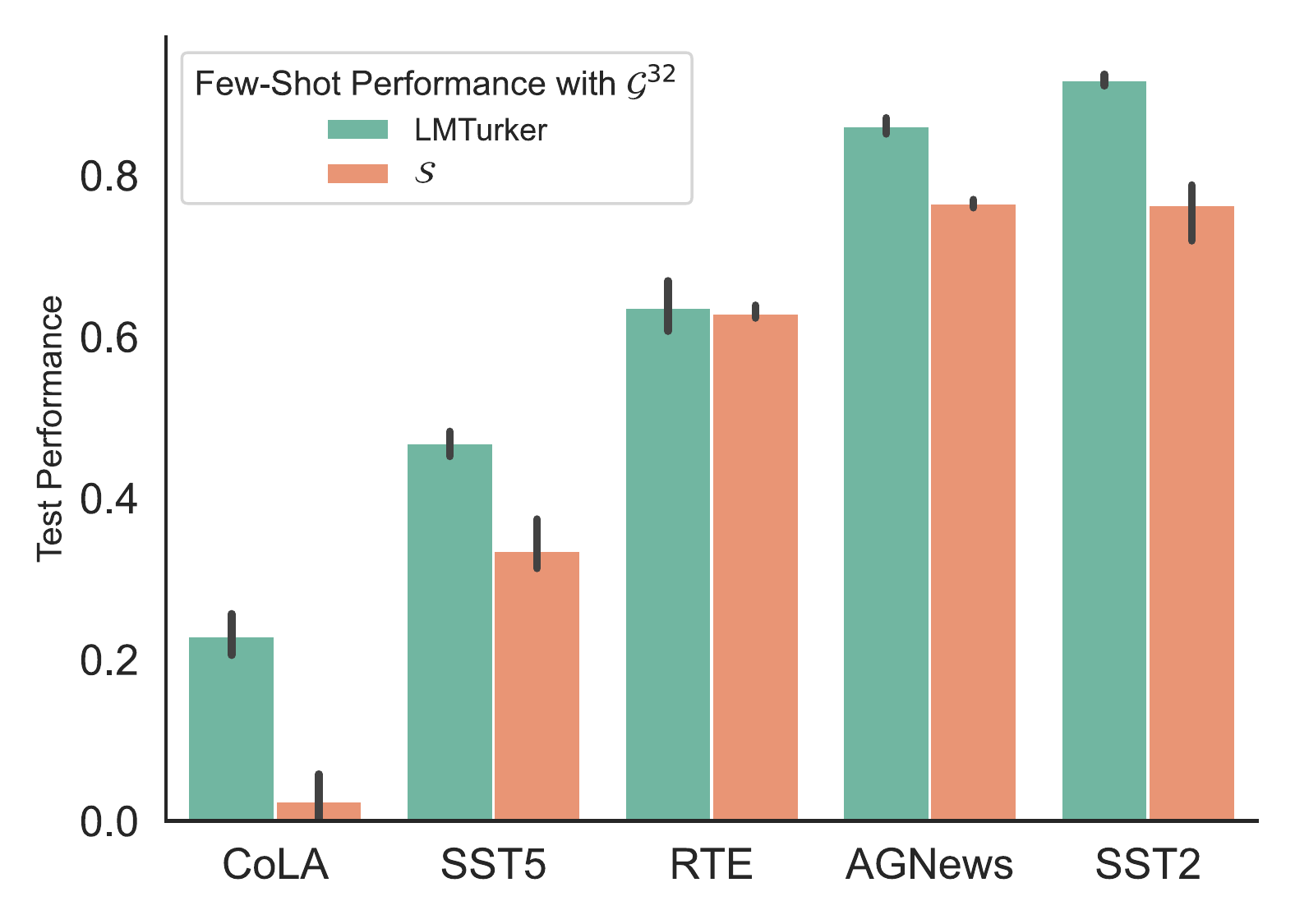}
 	}
\caption{
\emph{Few-shot} test set performance of
\mdrs and $\mathcal{S}$.
We use the few-shot gold datasets
$\mathcal{G}^8$ (top),
$\mathcal{G}^{16}$ (middle),
and $\mathcal{G}^{32}$ (bottom).}
\figlabel{fewshotperf}
\end{figure}

\section{Experiment}
\subsection{Few-shot performance (non-iterative)}
We compare few-shot performance
of \mdrs
and
the small model $\mathcal{S}$
when \emph{only $\mathcal{G}$ is used}.
\mdr performance
is comparable to prior work
\citep{PETpaper,schick2020s,gao-etal-2021-making}
as shown in \tabref{fewshotperfcomp}.

\figref{fewshotperf}
compares performance of \mdrs
and $\mathcal{S}$.
Appendix \secref{appendix:numeric}
\tabref{appendix:fewshotnumeric}
reports numeric values.
\mdrs perform clearly
better than $\mathcal{S}$
on CoLA, SST5, AGNews, and SST2; e.g.,
for SST2, for train/dev size 16,
\mdr
accuracy is 93.08\% vs.\ 75.83\% for
$\mathcal{S}$.
\mdrs' superiority
over $\mathcal{S}$  
on RTE is
modest.
As an inference task, RTE is more challenging than
classification (e.g., AGNews).
We hypothesize that current few-shot learners
require more data than $\mathcal{G}^{32}$ to
process difficult
tasks  better than
$\mathcal{S}$.  
Scaling up to even larger
PLMs is also a promising
direction \citep{GPT3paper,Prompttuningpaper}.

Overall,
\mdrs
outperform $\mathcal{S}$  
with clear margins,
evidencing that their annotations
can serve as supervisions for training
$\mathcal{S}$.
We next conduct iterative training
to improve performance of
$\mathcal{S}$ on $\mathcal{T}$
with supervisions from \mdrs.

\begin{figure}[h!]
\centering
\hspace{-.2cm}\vspace{-.25cm}\subfloat{
\includegraphics[width=.5\linewidth,height=0.2\textwidth]{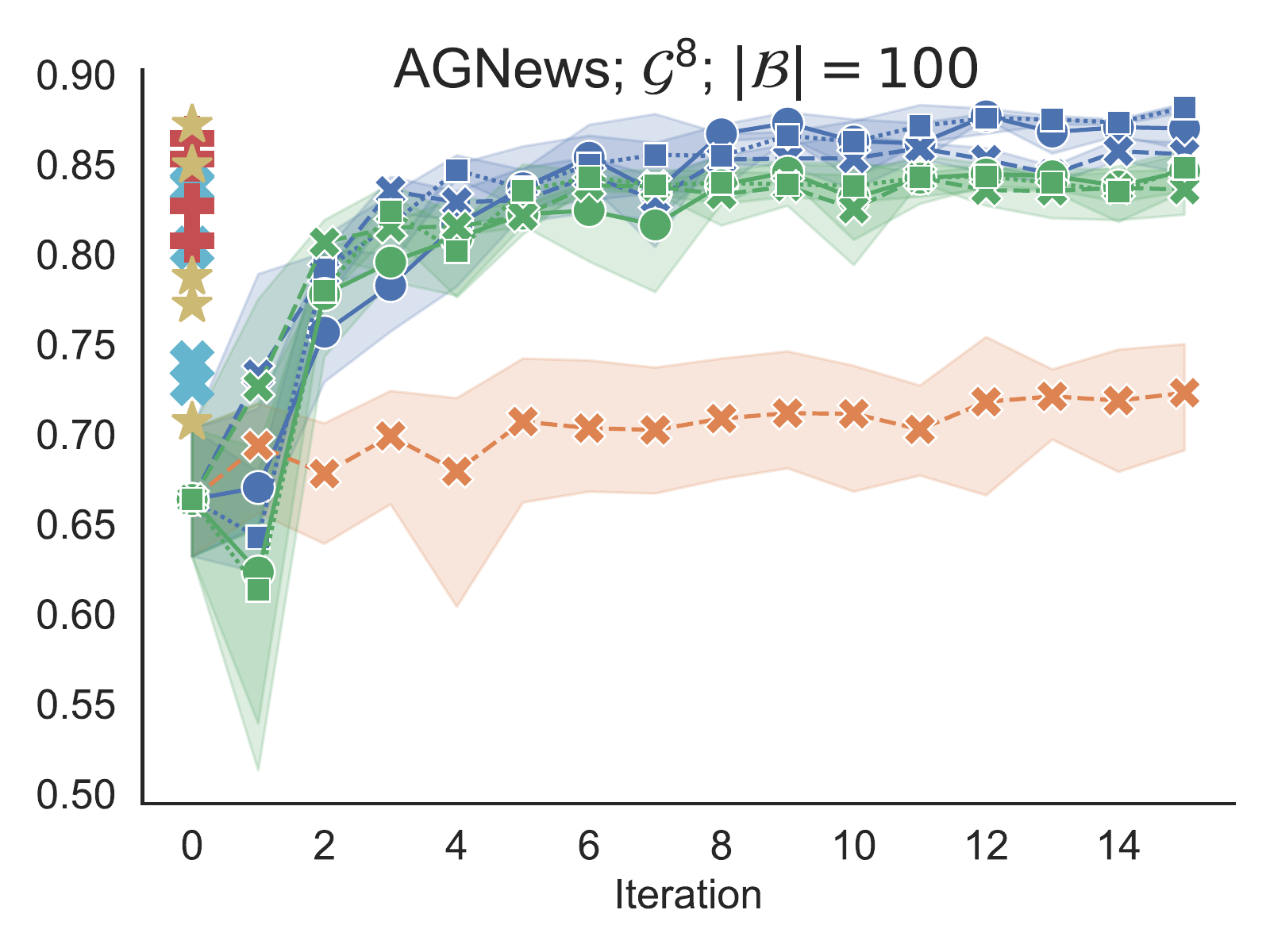}
}
\hspace{-.2cm}\vspace{-.25cm}\subfloat{
\includegraphics[width=.5\linewidth,height=0.2\textwidth]{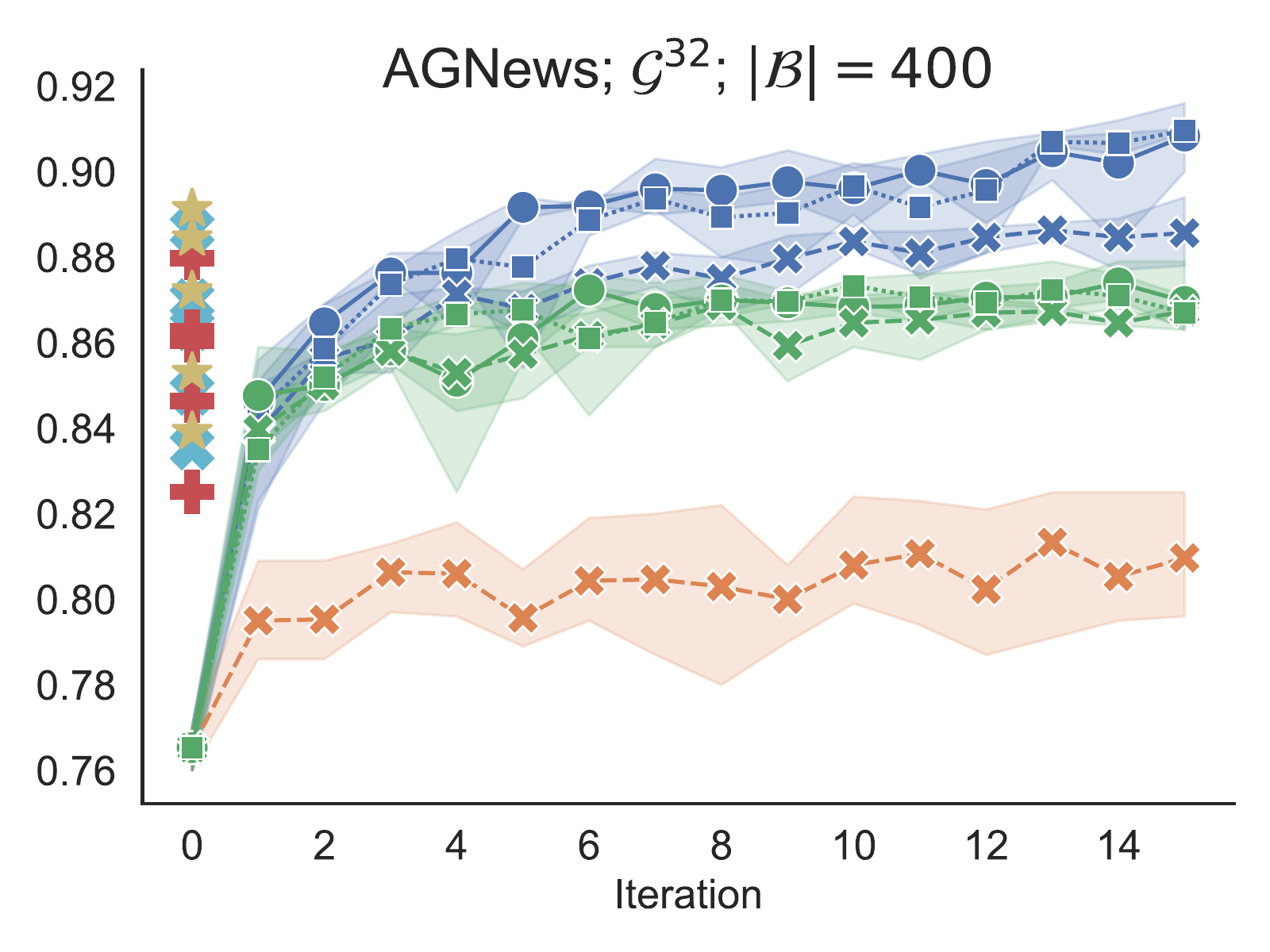}
}
\\
\hspace{-.2cm}\vspace{-.25cm}\subfloat{
\includegraphics[width=.5\linewidth,height=0.2\textwidth]{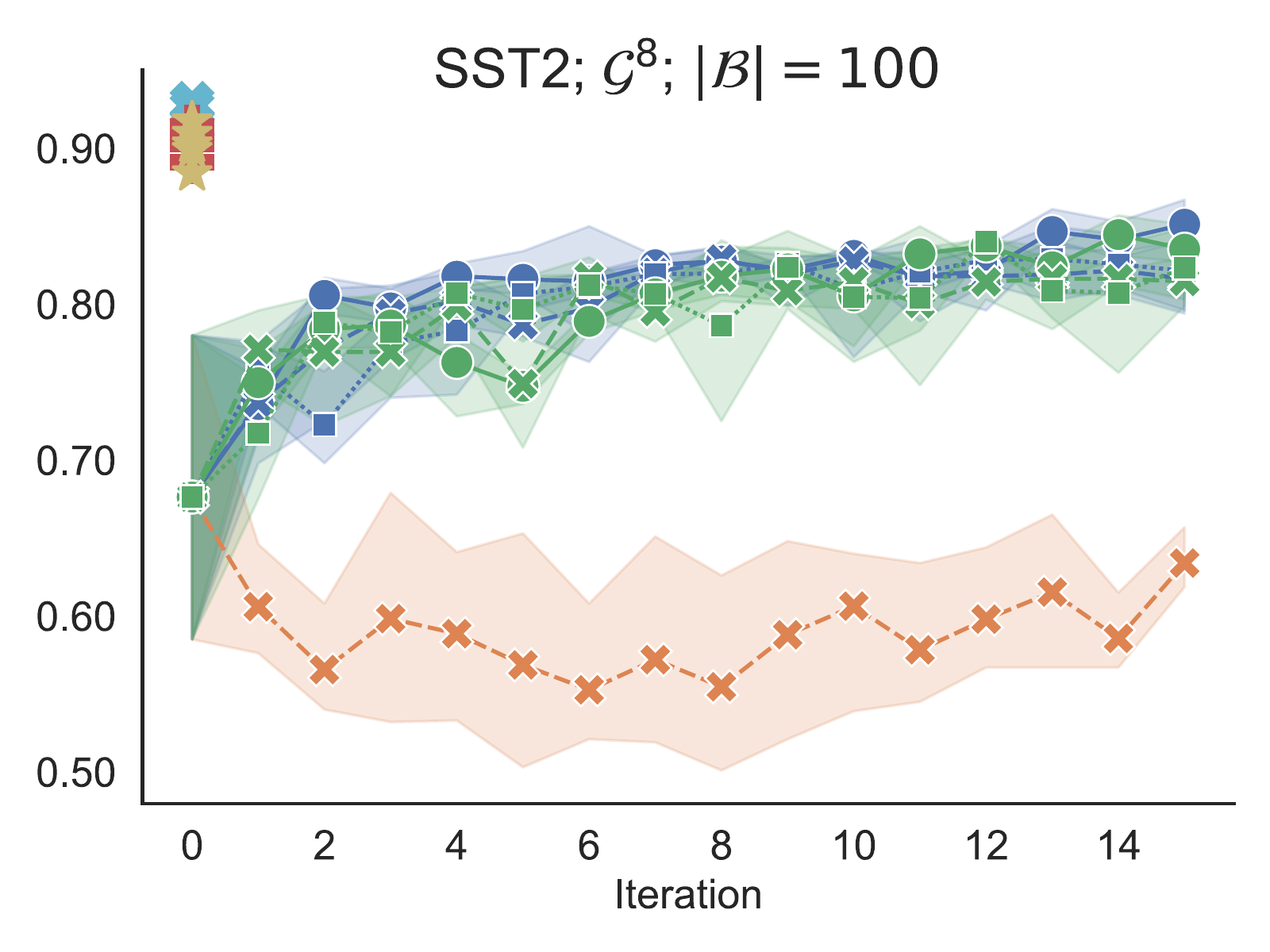}
}
\hspace{-.2cm}\vspace{-.25cm}\subfloat{
\includegraphics[width=.5\linewidth,height=0.2\textwidth]{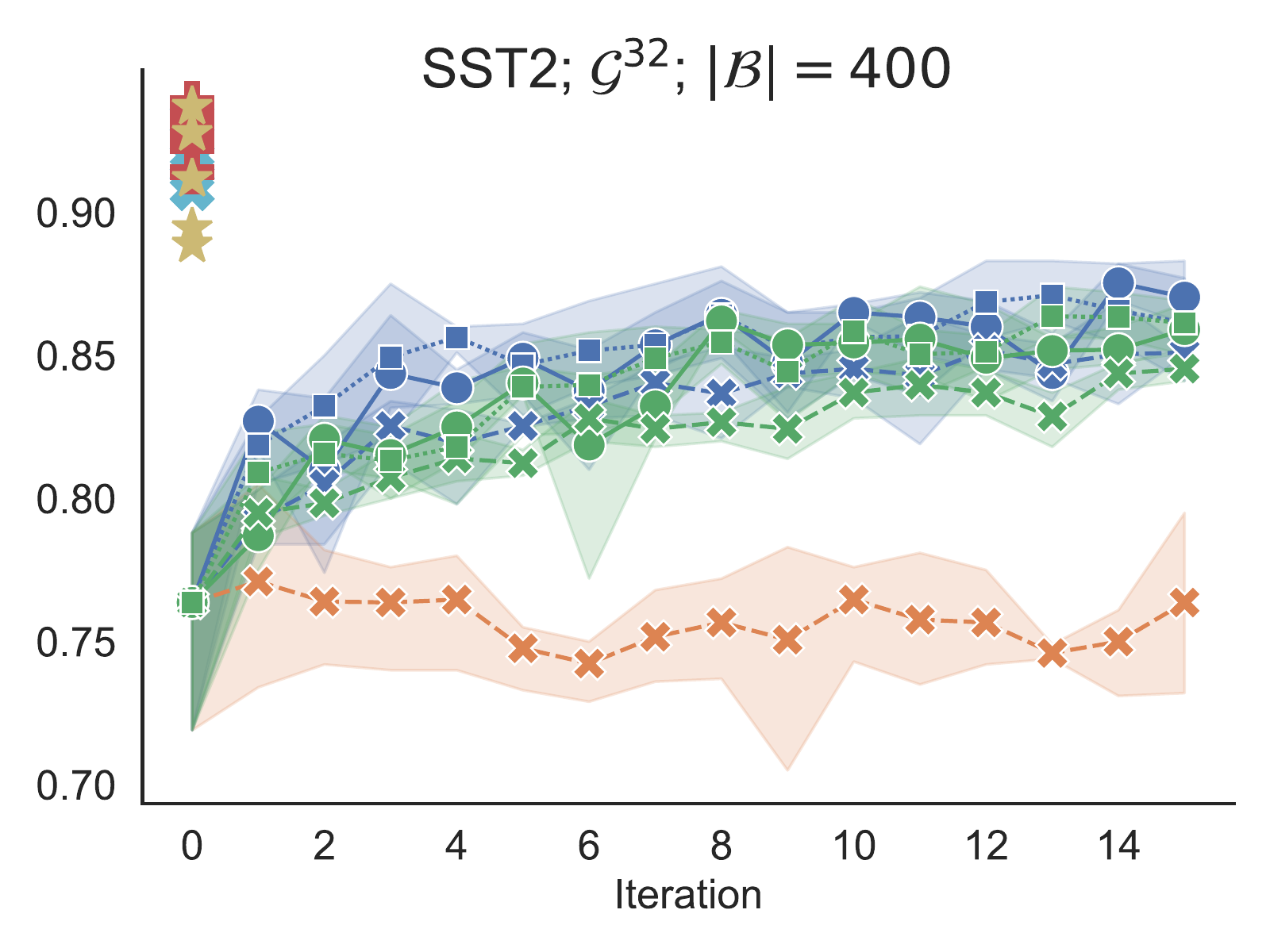}
}
\\
\hspace{-.2cm}\vspace{-.25cm}\subfloat{
\includegraphics[width=.5\linewidth,height=0.2\textwidth]{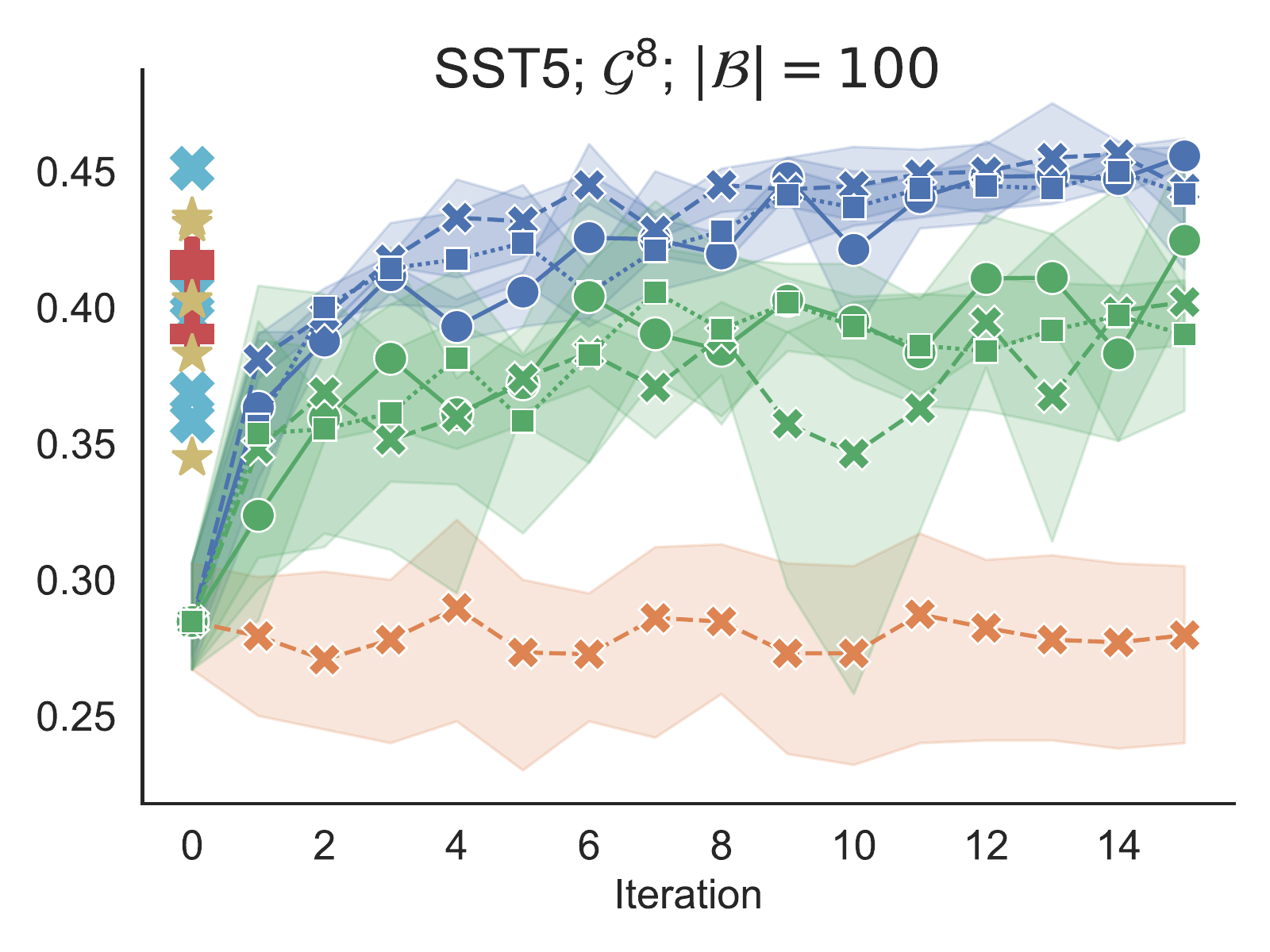}
}
\hspace{-.2cm}\vspace{-.25cm}\subfloat{
\includegraphics[width=.5\linewidth,height=0.2\textwidth]{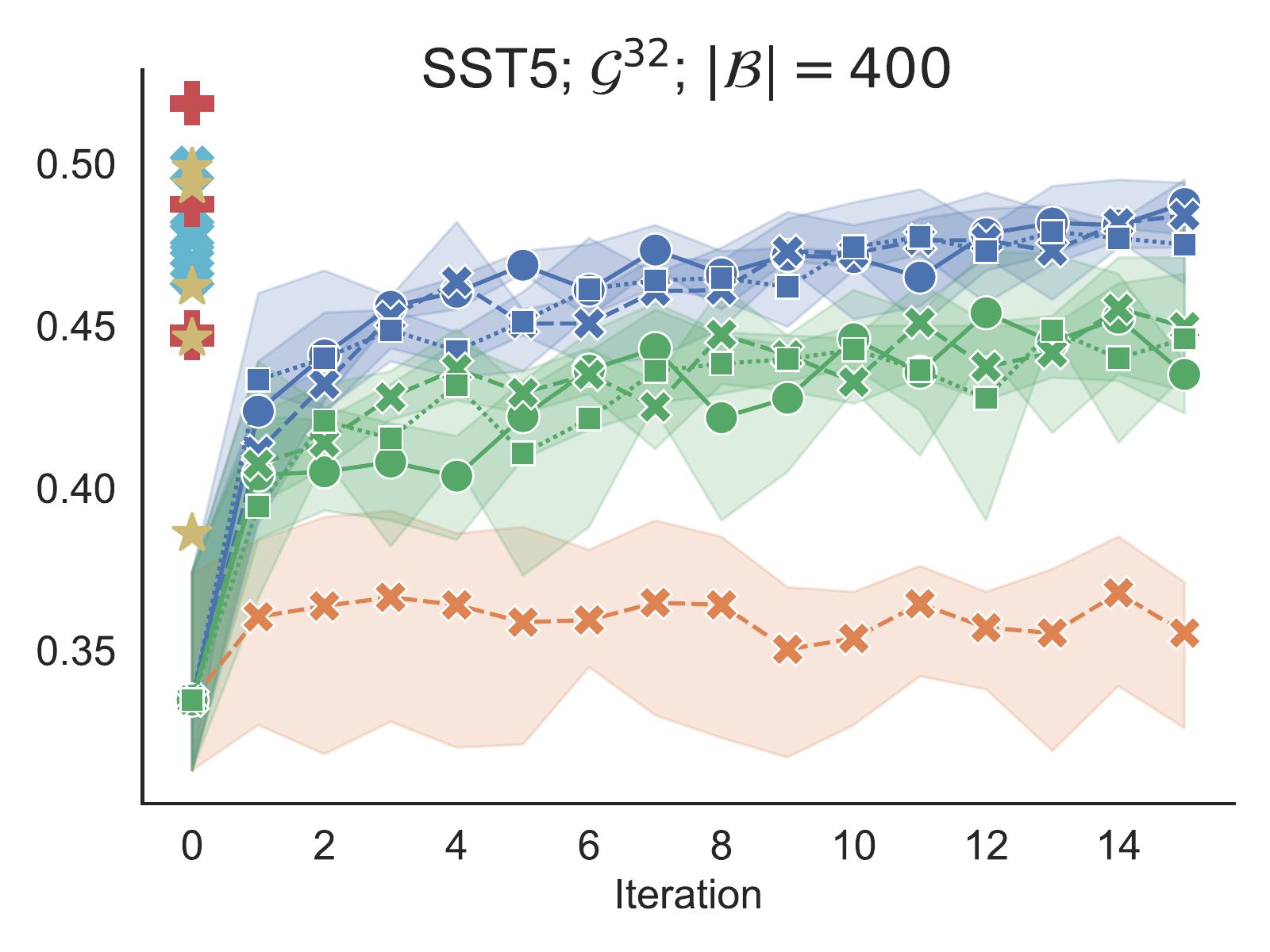}
}
\\
\hspace{-.2cm}\vspace{-.25cm}\subfloat{
\includegraphics[width=.5\linewidth,height=0.2\textwidth]{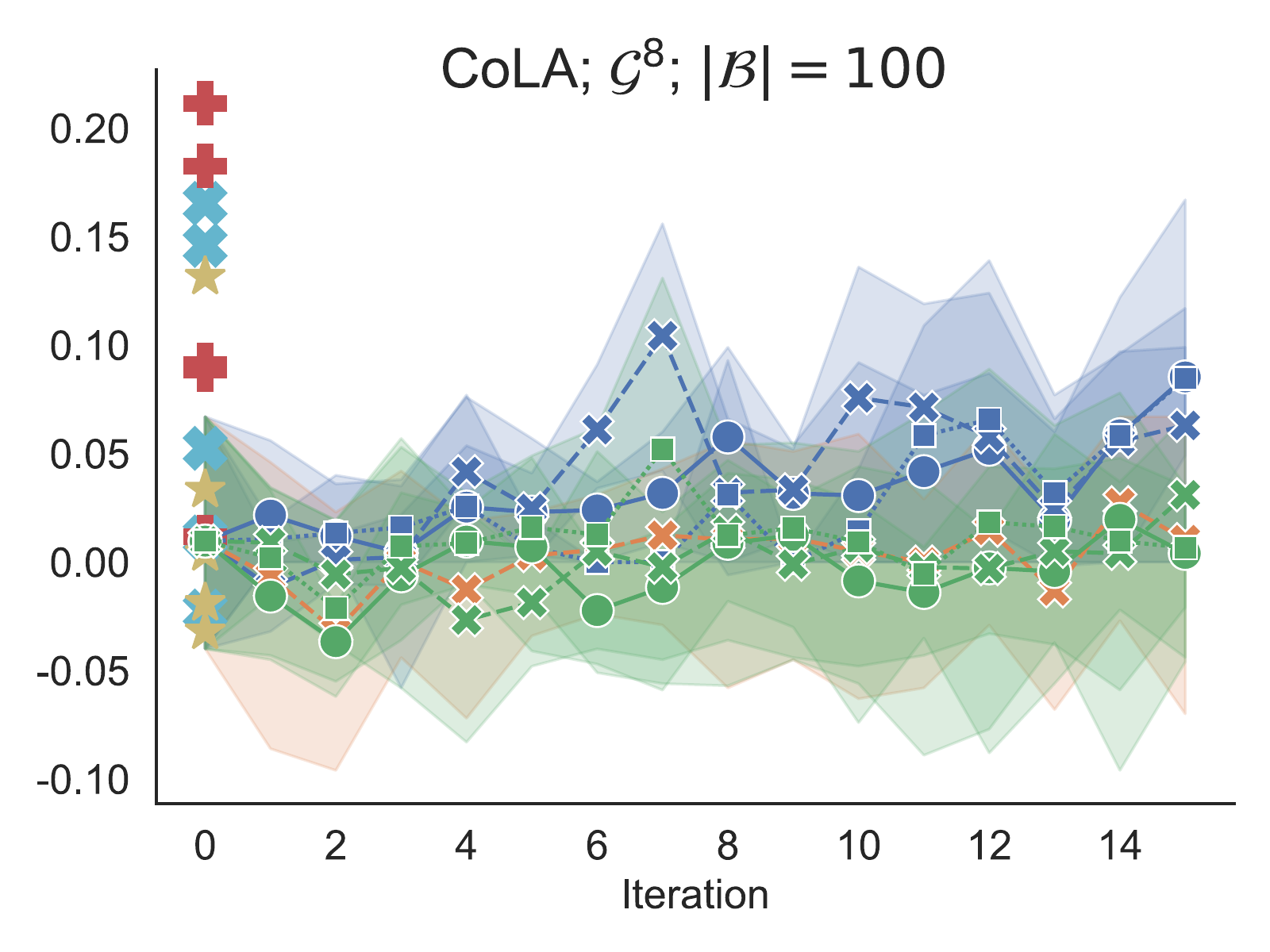}
}
\hspace{-.2cm}\vspace{-.25cm}\subfloat{
\includegraphics[width=.5\linewidth,height=0.2\textwidth]{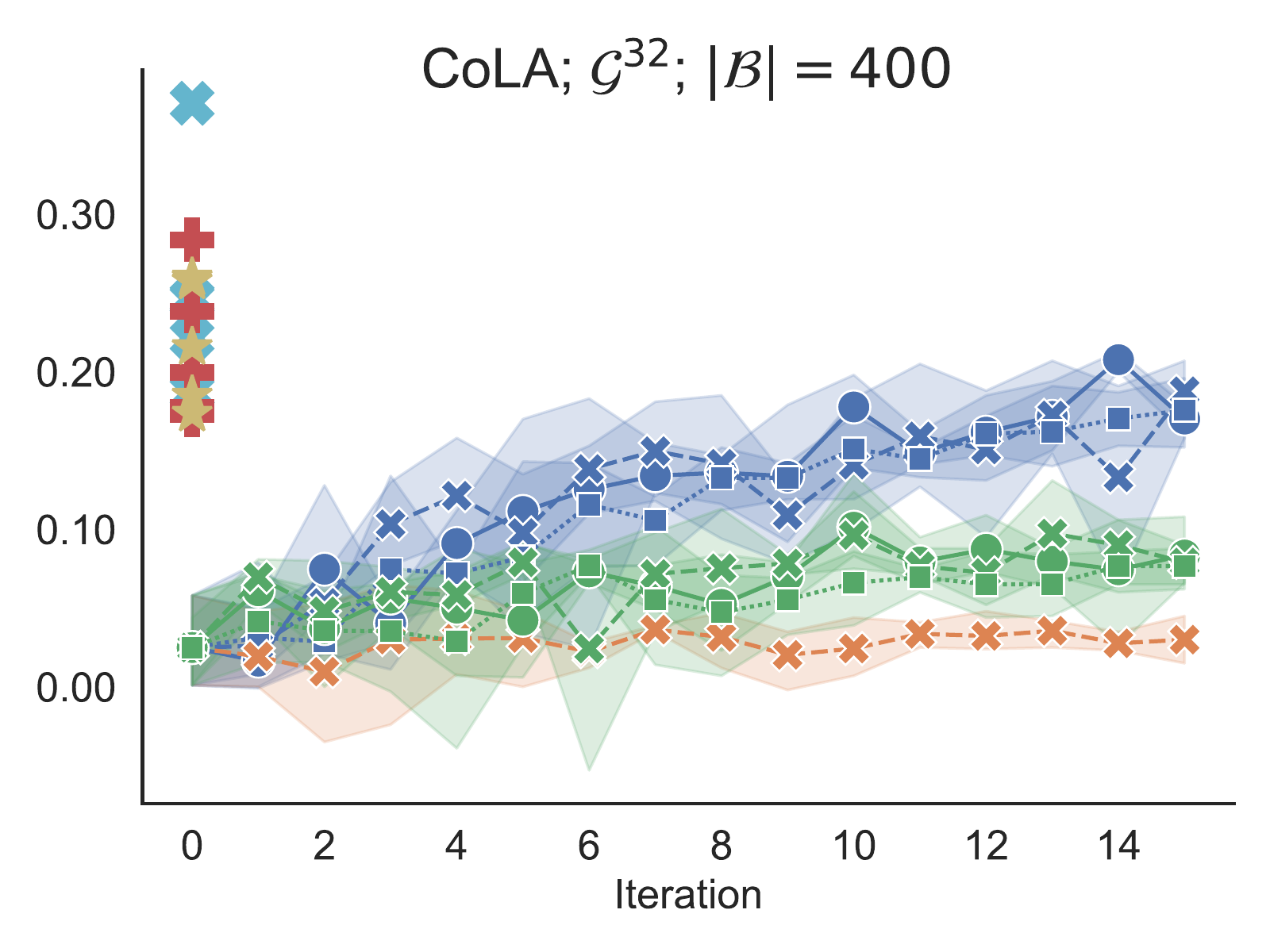}
}
\\
\hspace{-.2cm}\vspace{-.2cm}\subfloat{
\includegraphics[width=.5\linewidth,height=0.2\textwidth]{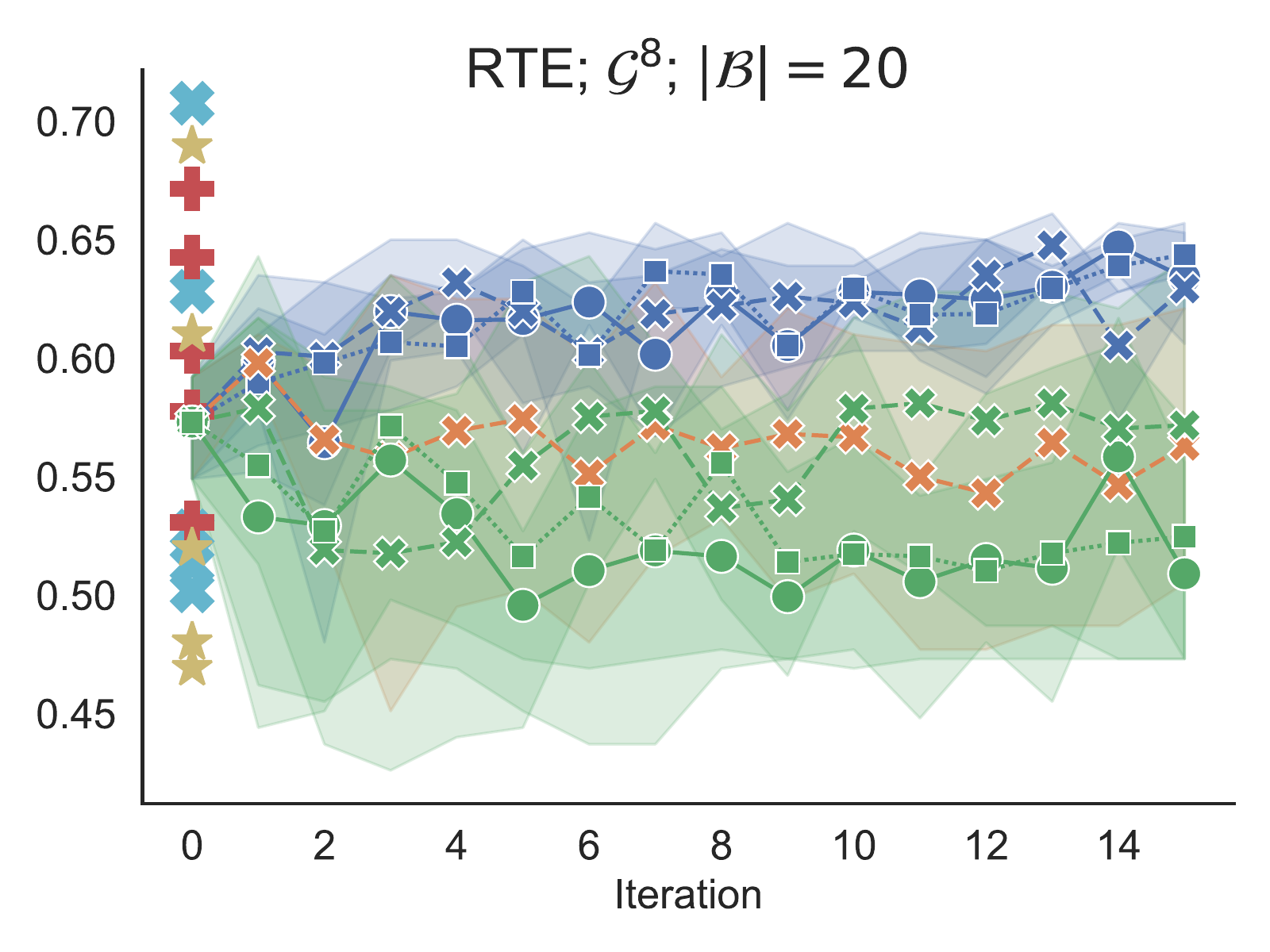}
}
\hspace{-.2cm}\vspace{-.2cm}\subfloat{
\includegraphics[width=.5\linewidth,height=0.2\textwidth]{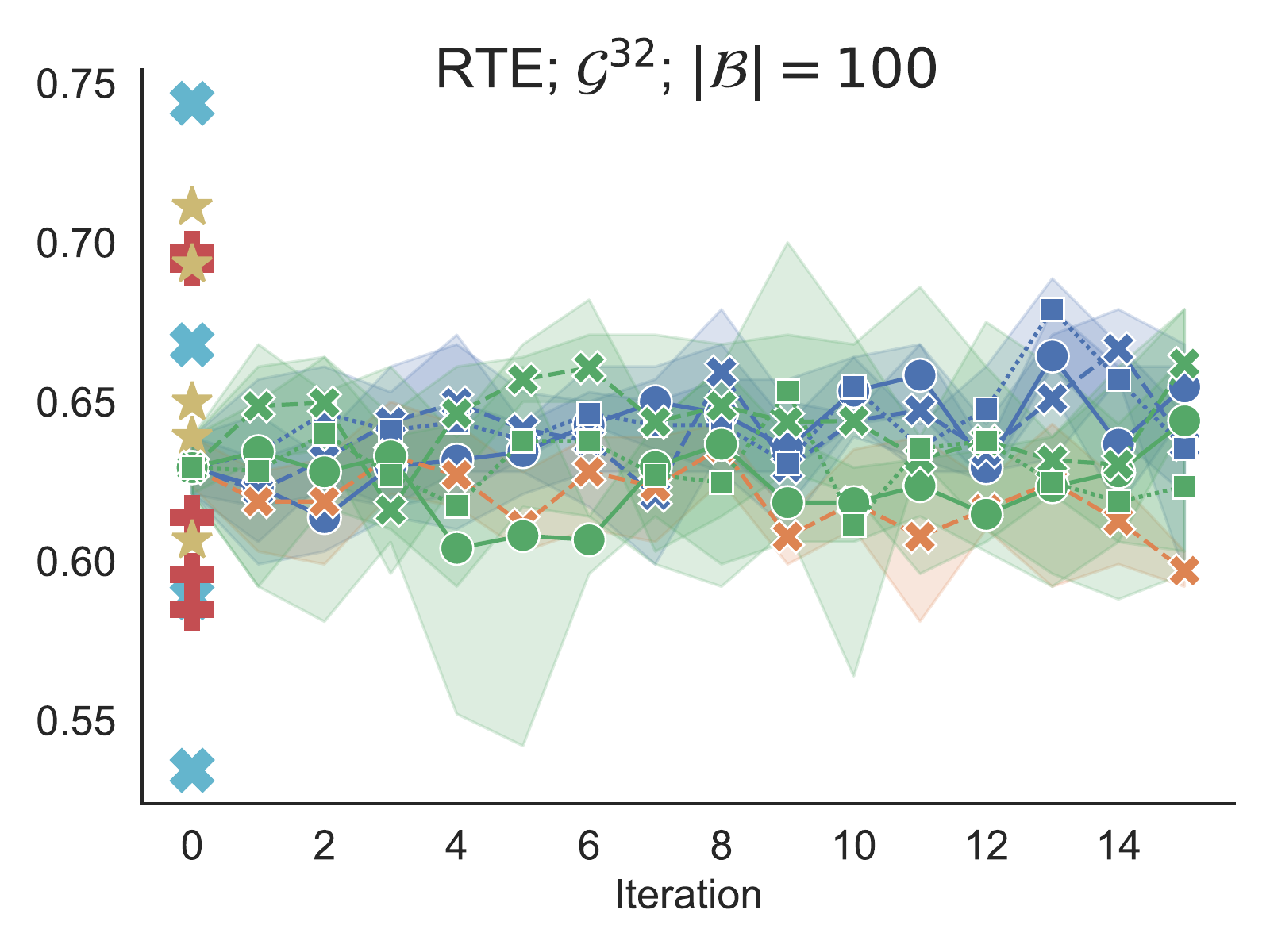}
}
\caption{
  Improving $\mathcal{S}$ with
  \textcolor{blue}{active learning (blue)},
  \textcolor{orange}{self training (orange)},
  and
  \textcolor{green}{\md (green)}.
  Free markers at step zero show \mdr performances;
  colors distinguish random seeds.
  Three acquisition functions are:
  Entropy ($\bullet$),
  LeastConfident (\tiny$\blacksquare$\normalsize),
  random sampling (\tiny\XSolidBold\normalsize).
  At iteration $j$, each experiment is repeated
  three times; we show
  mean and standard deviation.
  Appendix \figref{appendix:completeiteratives}
  visualizes more results.
}
\figlabel{iterative}
\end{figure}

\subsection{Iterative training}
We investigate
the effectiveness
of \md
by simulating
scenarios
analogous to
active learning.
Concretely,
we compare three
schemes of annotating
the sampled data
$\mathcal{B}$
at each annotation iteration $j$:

\begin{itemize}
\item \textcolor{blue}{Active learning (AL)}. We use $\mathcal{B}$'s
  \emph{gold labels} to show how $\mathcal{S}$ performs with expert
  annotations.  Gold labels are ideal, but costly because expert
  annotators need to be employed.

\item   \textcolor{orange}{Self training (ST)}.
  We use $\mathcal{S}^{j-1}$,
  the  model trained in
  the previous iteration,
  to annotate
  $\mathcal{B}$ \citep{yarowsky-1995-unsupervised,abney-2004-understanding,lee2013pseudo}.
  ST trades supervision
  quality for
  annotation cost;
  no extra cost is introduced.
  Because there is
  no external
  supervision,
  ST is expected to be a baseline.

\item   \textcolor{green}{\md}.
  We query the \mdrs
  to annotate $\mathcal{B}$.
  \mdrs
  are machine learning models, so there is no human labor.
  Based on
  the findings in \figref{fewshotperf}, \mdr supervisions
  are expected to have better quality
  than those of ST.
  Yet \md could fall behind
  AL because \mdr labels are not gold labels.
\end{itemize}

When sampling
$\mathcal{B}$ from $\mathcal{U}$ at
each iteration $j$,
we consider the strategies
described in
\secref{methodtrainingsmall}.
We employ Random for all three schemes and
Entropy/LeastConfident for
AL/\md.
Entropy and LeastConfident 
rely on $\mathcal{S}^{j-1}$.
Regarding the number of sampled examples,
we experiment with
$|\mathcal{B}|$=100 and $|\mathcal{B}|$=400
for SST2, SST5, AGNews,  CoLA.
Due to RTE's small size, we use
$|\mathcal{B}|$=20 and $|\mathcal{B}|$=100.
We run for 15 iterations of improving $\mathcal{S}$.
To aggregate annotations from \mdrs,
we use MajorityVoting
(\secref{subsec:aggregateannotations}),
which is widely used in crowdsourcing.
See \secref{varyaggregations} for a comparison
of various  aggregation methods.

\figref{iterative}
\textbf{compares AL, ST, and \md.}
ST (orange)
noticeably helps $\mathcal{S}$ to perform
progressively better on AGNews, e.g., when
comparing
$\mathcal{S}^{15}$ to $\mathcal{S}^{0}$
shown in
the first row, especially when $|\mathcal{B}|$=400.
However, we do not
identify clear
improvements
when looking at other tasks.
Except for RTE-$\mathcal{G}^{8}$,
ST clearly falls behind AL and \md.
This inferior performance
meets our expectation because
there is no
external supervision assisting
$\mathcal{S}$ to perform better
on \tasksymbol.
In what follows,
we omit ST for
clearer visualization and discussion.

AL (blue) performs
the best in most experiments.
However, this comes with extra
costs that are
not negligible: \emph{At each iteration},
human annotators need to
annotate 100--400 sentences.

\md (green) holds
a position
between AL and ST
on AGNews, SST2, SST5, and CoLA.
Somehow surprisingly,
\md performs almost comparably
to AL on SST2.
Unlike AL, \md
requires very little human labor;
the only human annotation
throughout the
entire process is
the
few-shot gold dataset
$\mathcal{G}$.
In contrast,
AL has high
human annotation cost, e.g.,
1000--4000 examples
by iteration ten.
\md also shows clear performance
improvements over ST.

Results on RTE are noisy;
we conjecture this is
due to its very small test set
(277 examples).
We do not observe
performance
improvement of $\mathcal{S}$
along the iterations
in experiment
RTE-$\mathcal{G}^{32}$-$|\mathcal{B}|$=$100$,
likely due to 
saturated task performance:
TinyBERT-General-4L-312D ($\mathcal{S}$)
achieves 66.6\% on RTE for the full train
set \citep{jiao-etal-2020-tinybert}.

\textbf{Comparing sampling strategies}.
Entropy ($\bullet$) and
LeastConfident (\tiny$\blacksquare$\normalsize)
outperform random sampling (\tiny\XSolidBold\normalsize)
in AGNews and SST2 with noticeable margins --
for both AL and \md, especially when $|\mathcal{B}|$=400.
They also surpass random sampling
when using \md for SST5 and CoLA with $\mathcal{G}^8$. 
In other words, Entropy and LeastConfident
assist \md to achieve the same
performance as of
using random sampling,
but with fewer annotations.
For example
in AGNews-$\mathcal{G}^8$-$|\mathcal{B}|$=$100$,
LeastConfident at iteration six already
achieves comparable performance as random
sampling at iteration eleven.
This is
economically and environmentally
beneficial because
the number of queries made to \mdrs,
i.e., the cost of running
inference passes on
the array of large PLMs,
is significantly reduced.

Overall, we show that
\md can be used to create
datasets for training a specialized
model $\mathcal{S}$ of solving
$\mathcal{T}$ in practical scenarios.
To reduce computational cost, we use only
Entropy in what follows.

\begin{figure}[t]
\centering
\subfloat{
\includegraphics[width=.5\linewidth,height=0.2\textwidth]{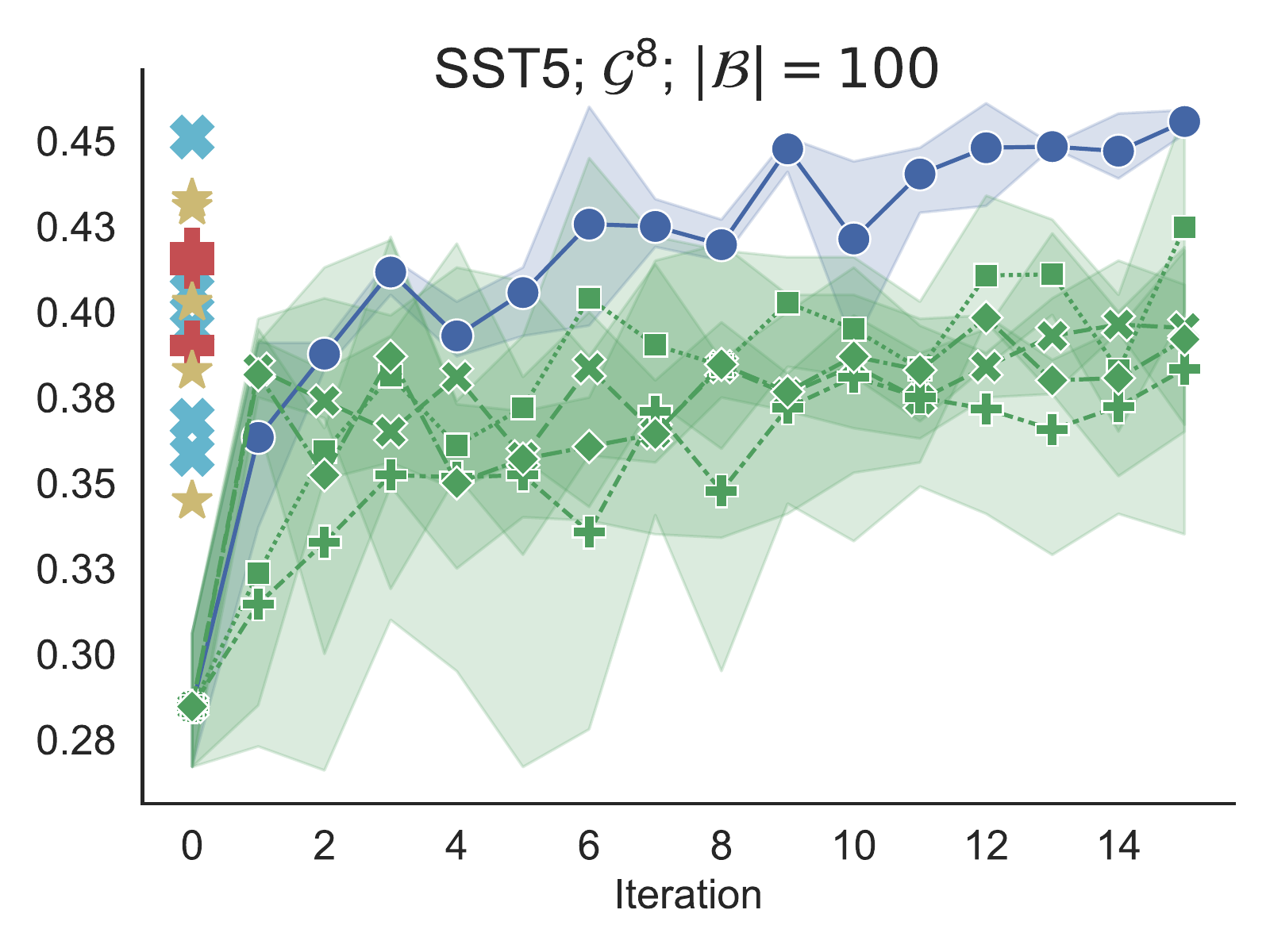}
}
\vspace{-.2cm}\subfloat{
\includegraphics[width=.5\linewidth,height=0.2\textwidth]{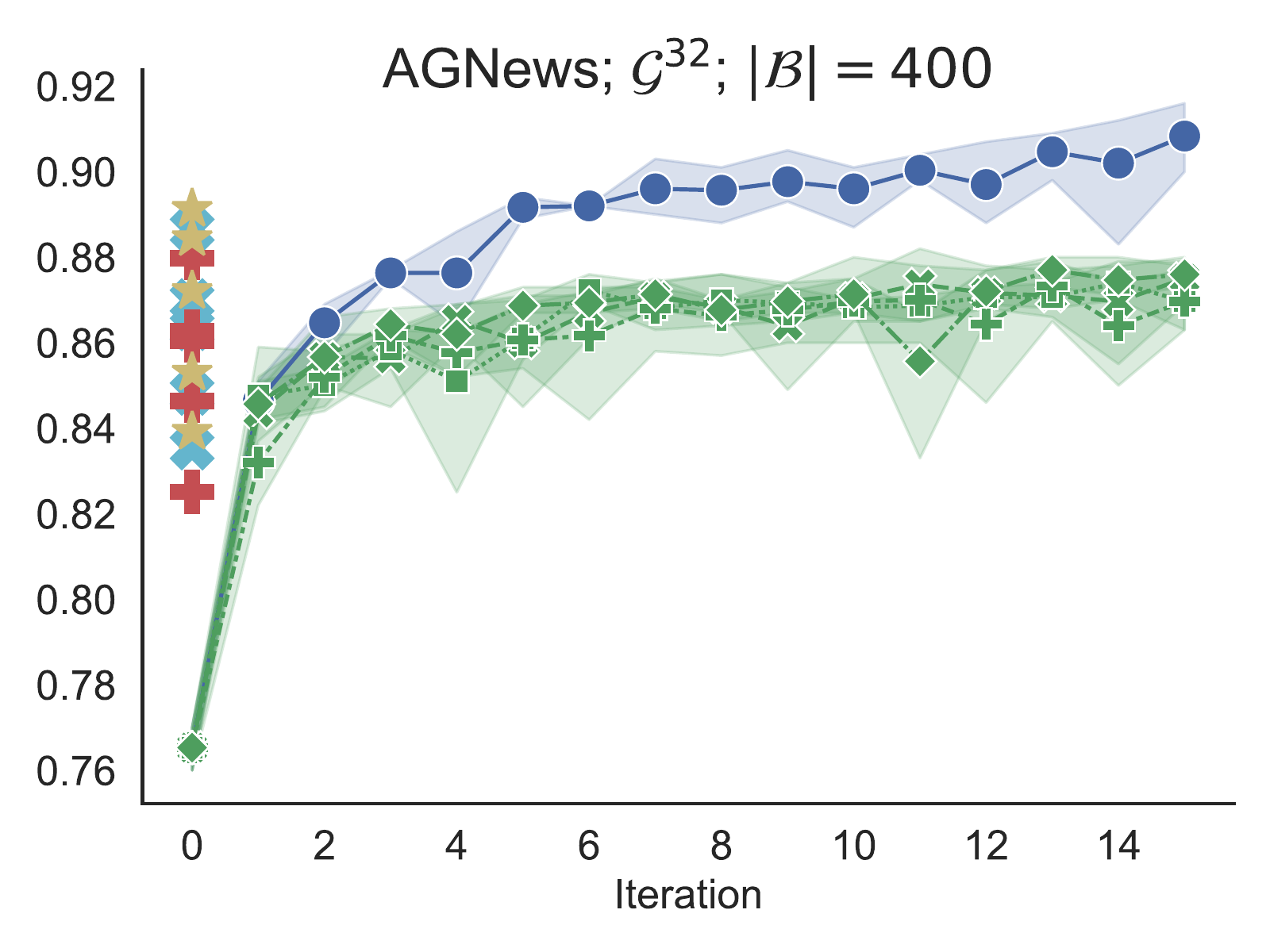}
}
\\
\vspace{-.2cm}\subfloat{
\includegraphics[width=.5\linewidth,height=0.2\textwidth]{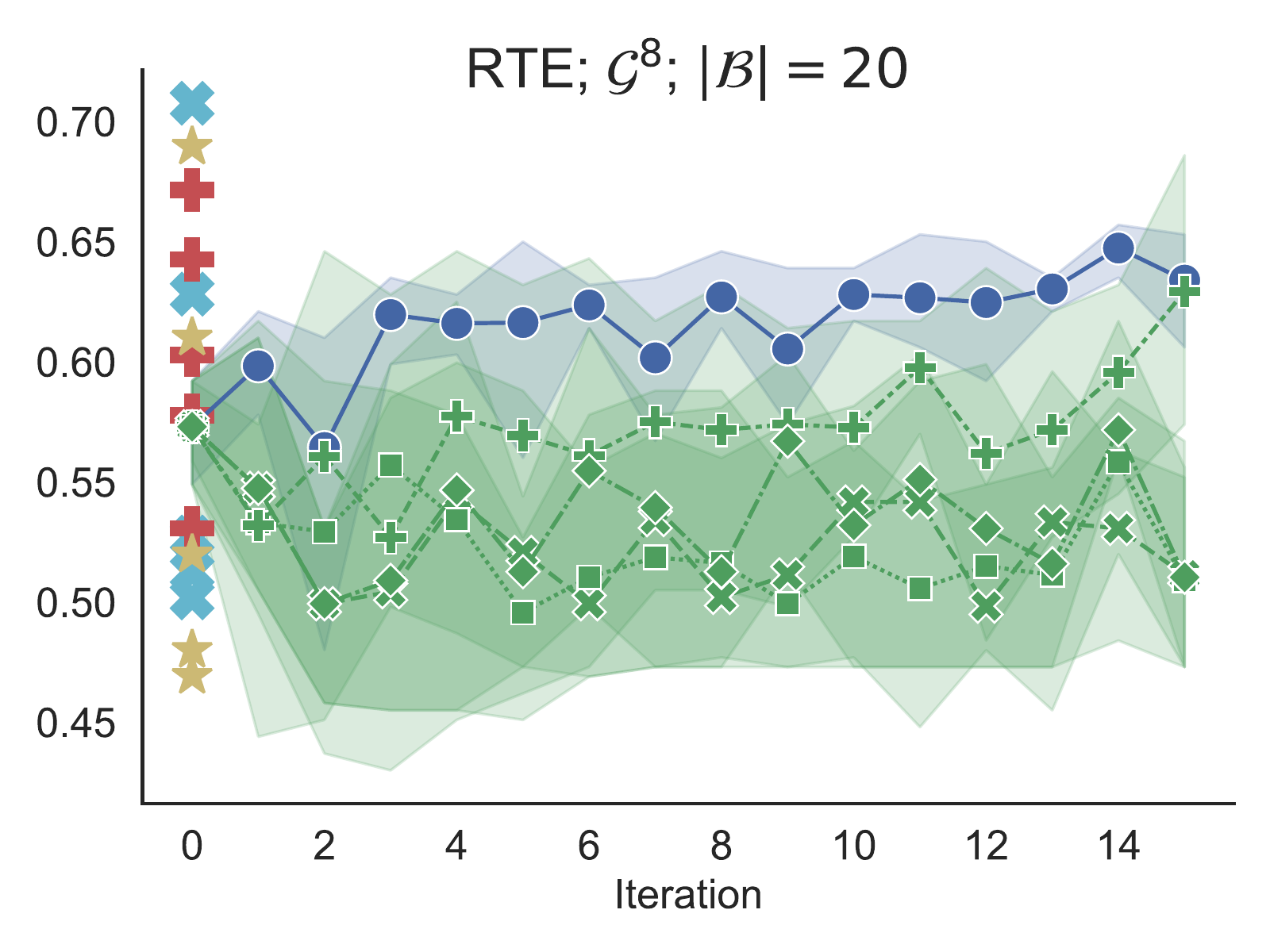}
}
\vspace{-.2cm}\subfloat{
\includegraphics[width=.5\linewidth,height=0.2\textwidth]{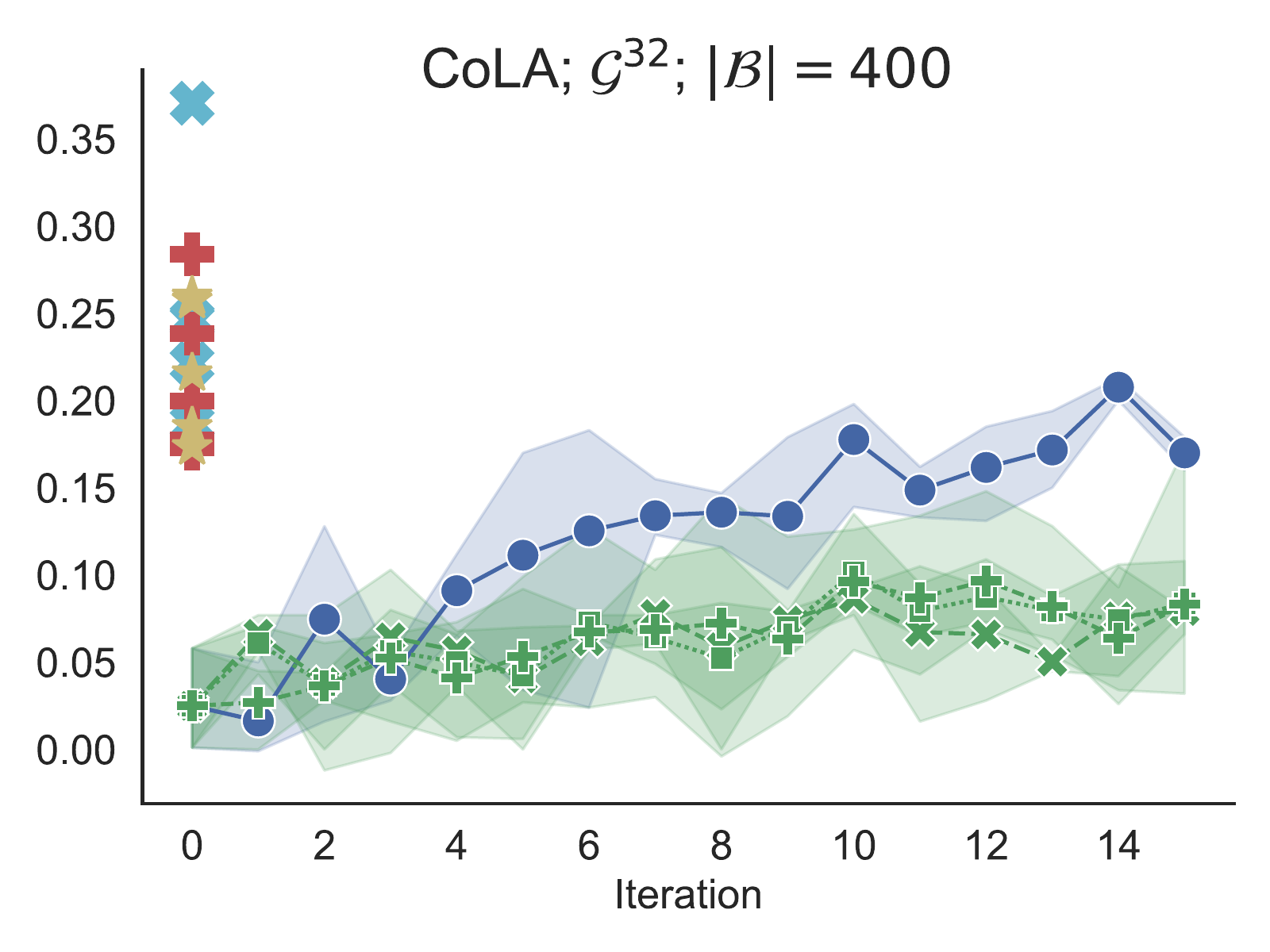}
}
\caption{
  Comparing 
  strategies of aggregating \mdr
  annotations.
  We compare \textcolor{green}{\md (green)} with
  \textcolor{blue}{AL (blue)}.
  Strategies: 
  LogitVoting (\tiny\XSolidBold\normalsize),
  MajorityVoting (\tiny$\blacksquare$\normalsize),
  WeightedLogitVoting (\scriptsize$\blacklozenge$\normalsize),
  BestWorker (\scriptsize\ding{58}\normalsize).  
  AL uses gold labels without aggregation ($\bullet$).
}
\figlabel{varyaggs}
\end{figure}

\subsection{Design choice 1: Aggregation strategies}
\seclabel{varyaggregations}
\figref{varyaggs} compares effectiveness
of different strategies of
aggregating
\mdr annotations
(\secref{subsec:aggregateannotations}).
Looking at 
SST5 and AGNews results (top two images),
we observe that
committee-style aggregation
(LogitVoting (\tiny\XSolidBold\normalsize),
MajorityVoting (\tiny$\blacksquare$\normalsize),
and WeightedLogitVoting (\scriptsize$\blacklozenge$\normalsize))
generally outperforms
BestWorker (\scriptsize\ding{58}\normalsize), which
simply relies on
the
\mdr performing best on $\mathcal{G}_{dev}$.
\mdrs perform well
on these two datasets
as shown by the free markers
at iteration zero;
ensembling their
predictions results in
higher-quality datasets.

In contrast,
BestWorker (\scriptsize\ding{58}\normalsize)
has stellar performance
on RTE (bottom-left),
outperforming
committee-style aggregation.
Note that
even the \mdrs
do not perform
really well in this experiment, as shown
by the free markers at iteration zero --
some \mdrs even
perform worse than $\mathcal{S}$.
Ensembling these low-quality annotations
seems a worse option than simply
relying on the best \mdr.
For CoLA, we observe comparable
performance of different aggregation
strategies.

\begin{figure}[t]
\centering
\subfloat{
\includegraphics[width=.5\linewidth,height=0.2\textwidth]{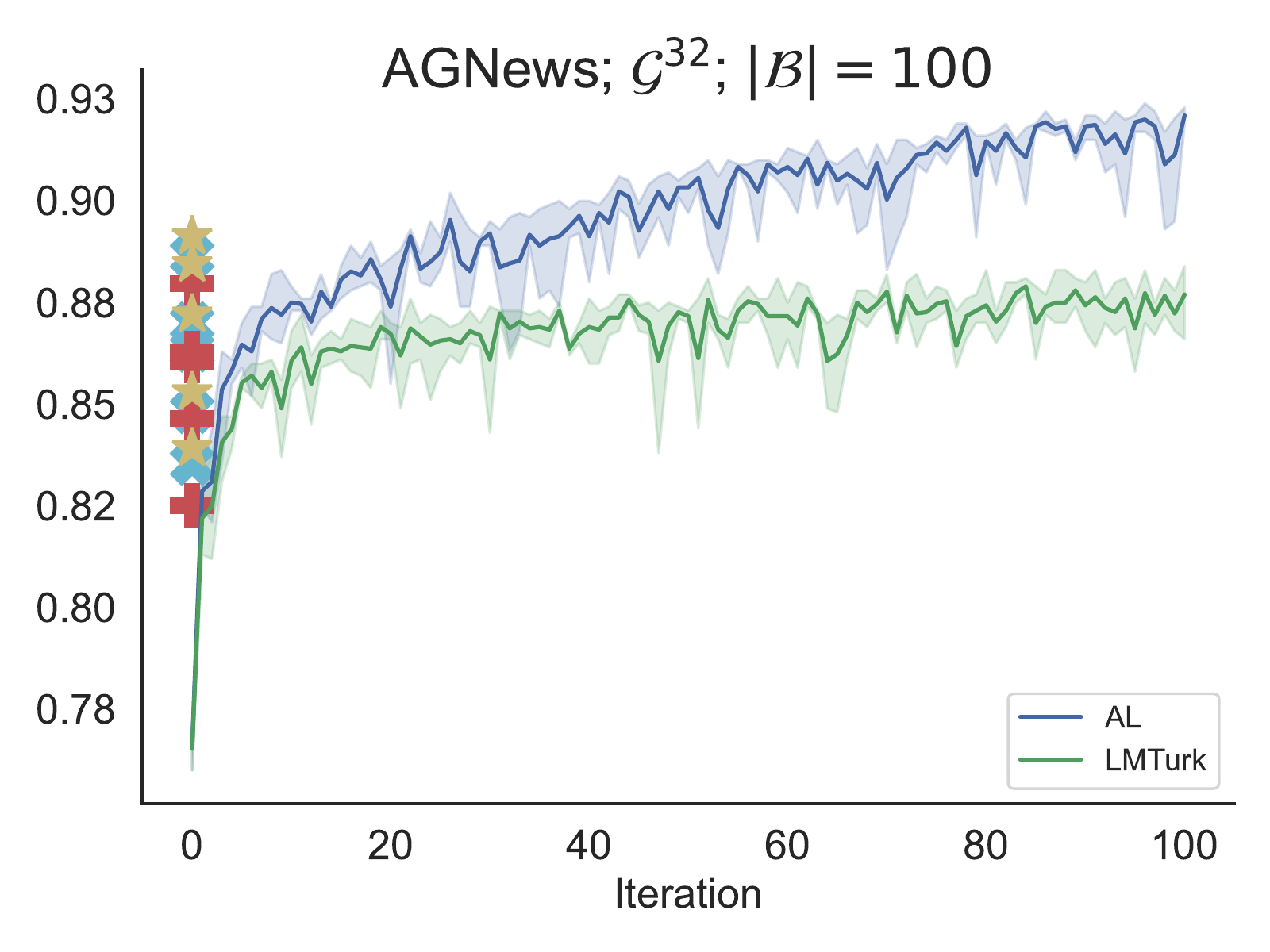}
}
\vspace{-.15cm}\subfloat{
\includegraphics[width=.5\linewidth,height=0.2\textwidth]{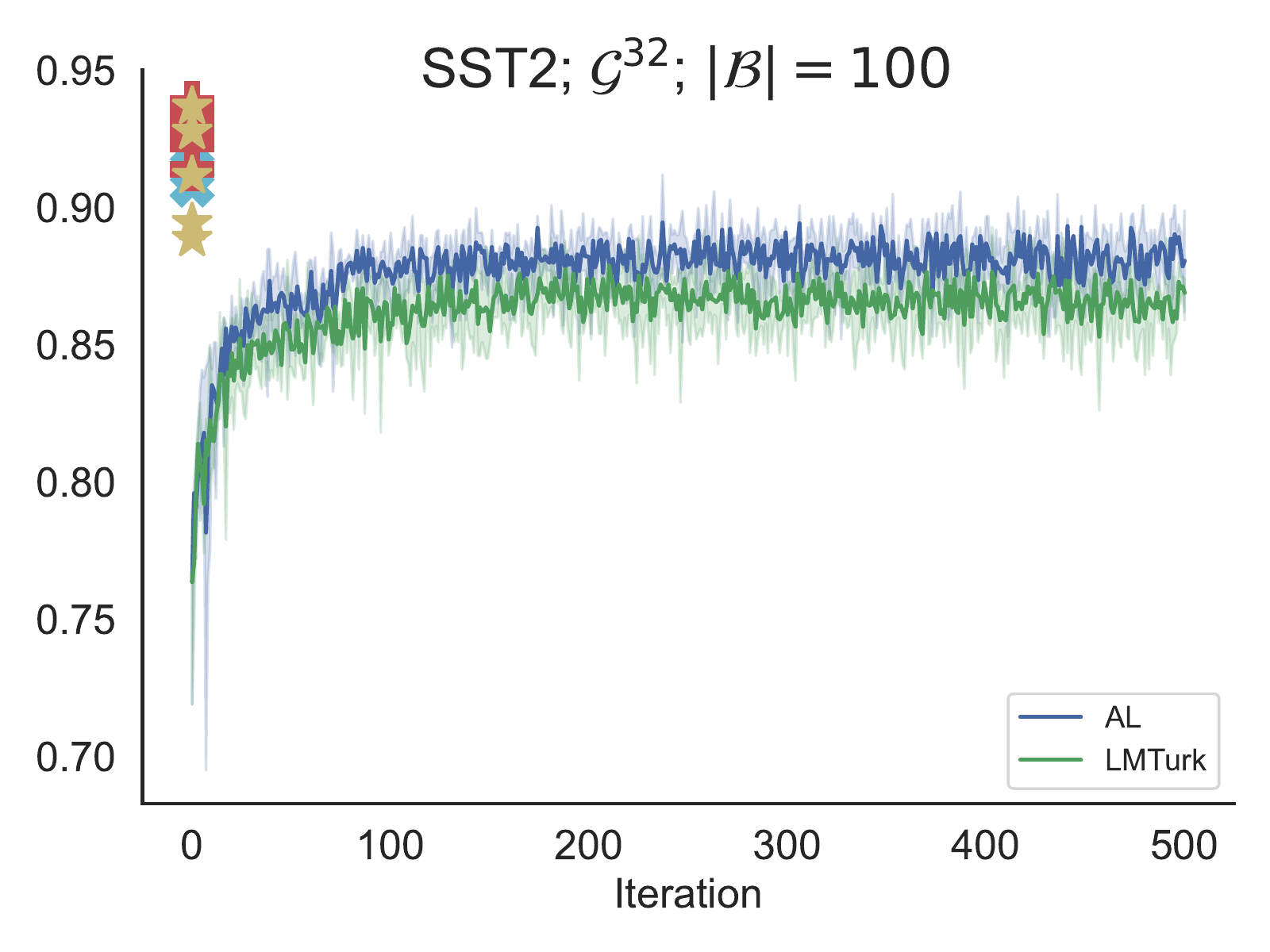}
}
\caption{
  Running more iterations 
  of improving $\mathcal{S}$ with AL and \md.
  Sampling strategy Entropy is used for both methods;
  WeightedLogitVoting is used for aggregating
  \mdr annotations.
}
\figlabel{moreiters}
\end{figure}

\subsection{Design choice 2: More iterations}
We hypothesize that AL performance is an upper bound for
performance when $\mathcal{S}$
is trained with
\mdr annotations -- recall that the AL annotations are gold labels.
\figref{moreiters} compares
AL and \md when running
100 iterations of improving $\mathcal{S}$ on AGNews
and
500 iterations on SST2.
As expected, AL
outperforms \md because
the pool of human-annotated data expands.
The performance of
$\mathcal{S}$ progressively 
approaches that of
the \mdrs; \md performs
comparably to AL in SST2,
however, no human labor is required.

\begin{figure}[t]
\centering
\subfloat{
\includegraphics[width=.5\linewidth,height=0.2\textwidth]{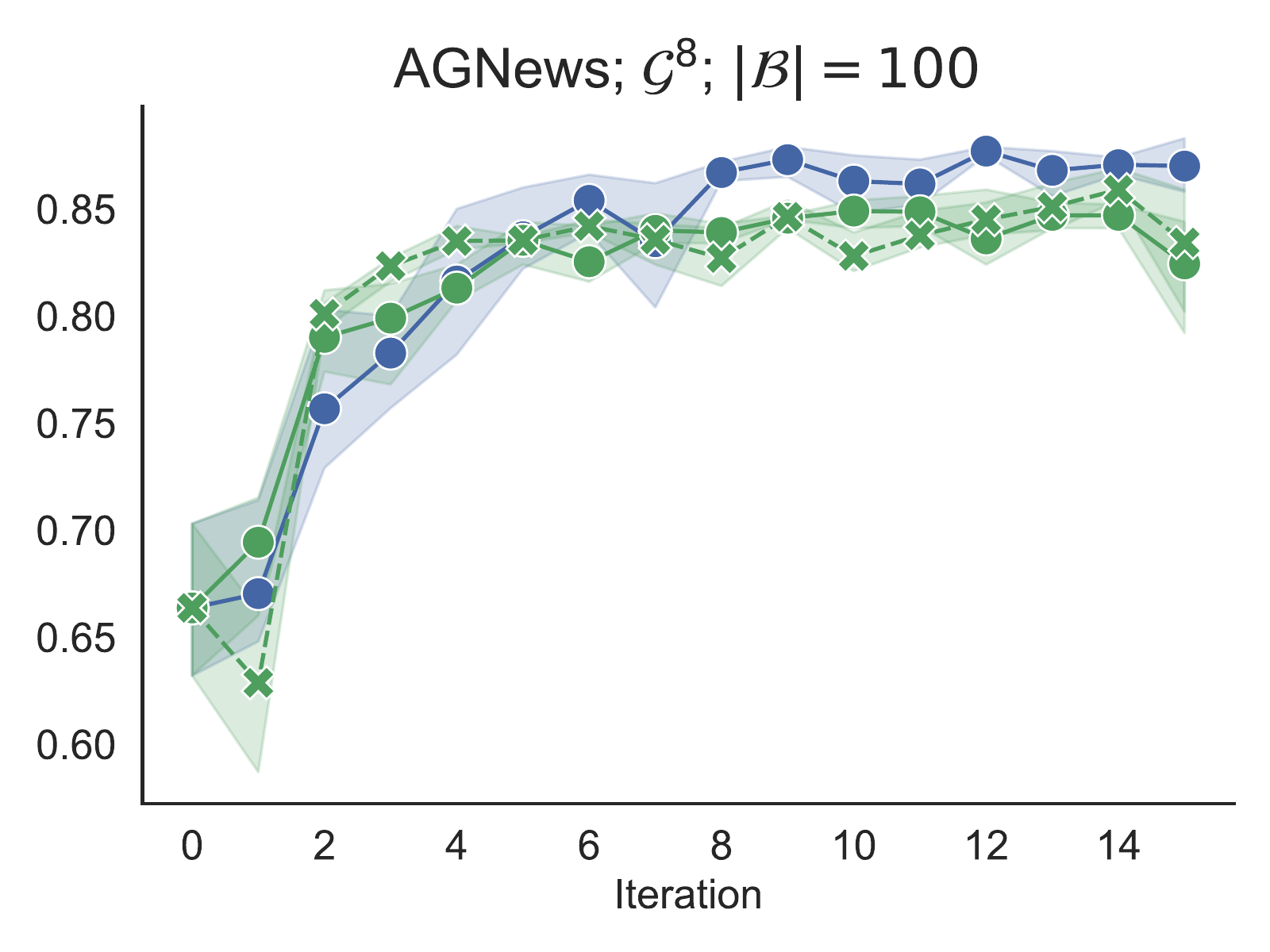}
}
\vspace{-.2cm}\subfloat{
\includegraphics[width=.5\linewidth,height=0.2\textwidth]{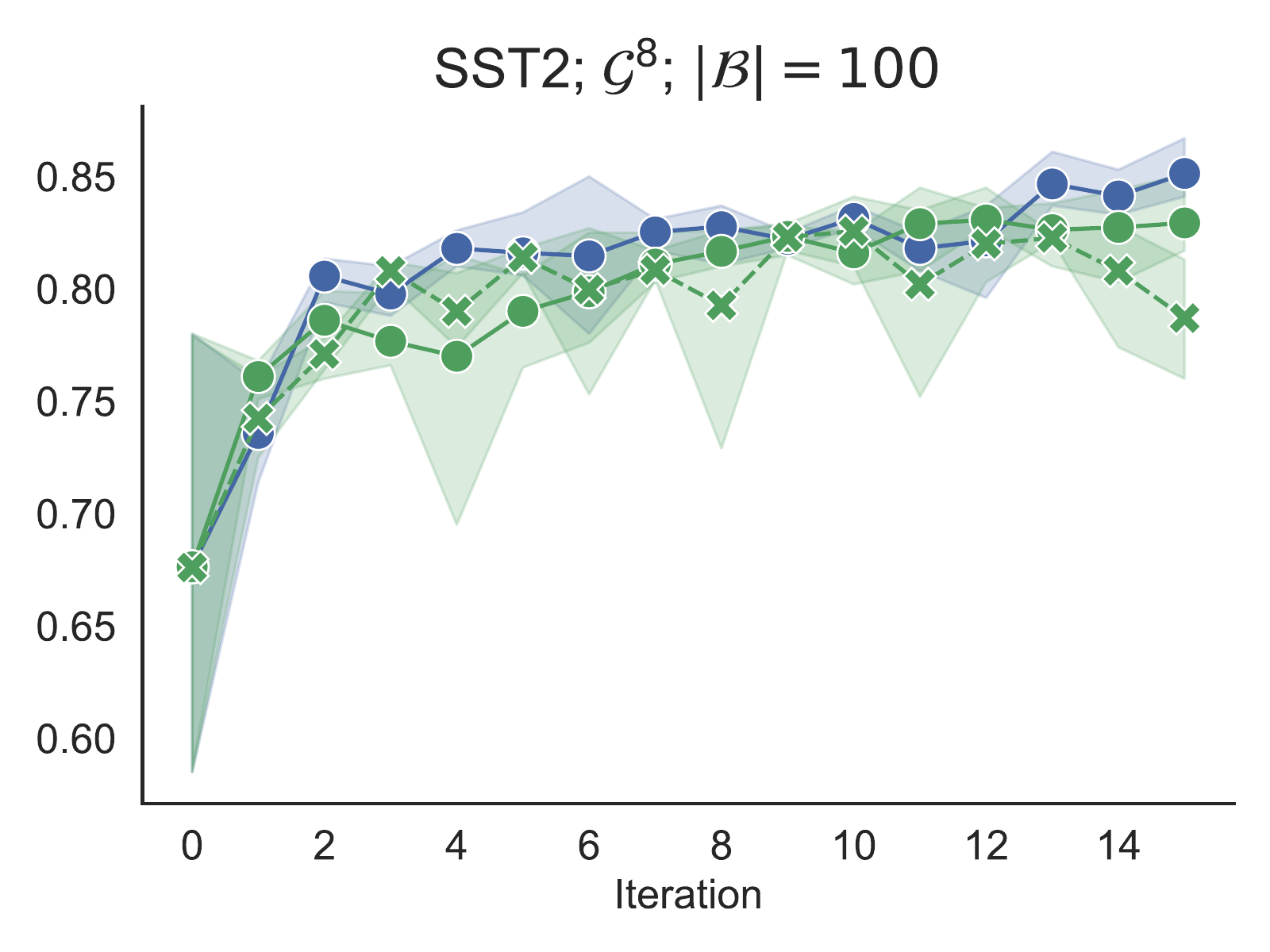}
}
\\
\vspace{-.2cm}\subfloat{
\includegraphics[width=.5\linewidth,height=0.2\textwidth]{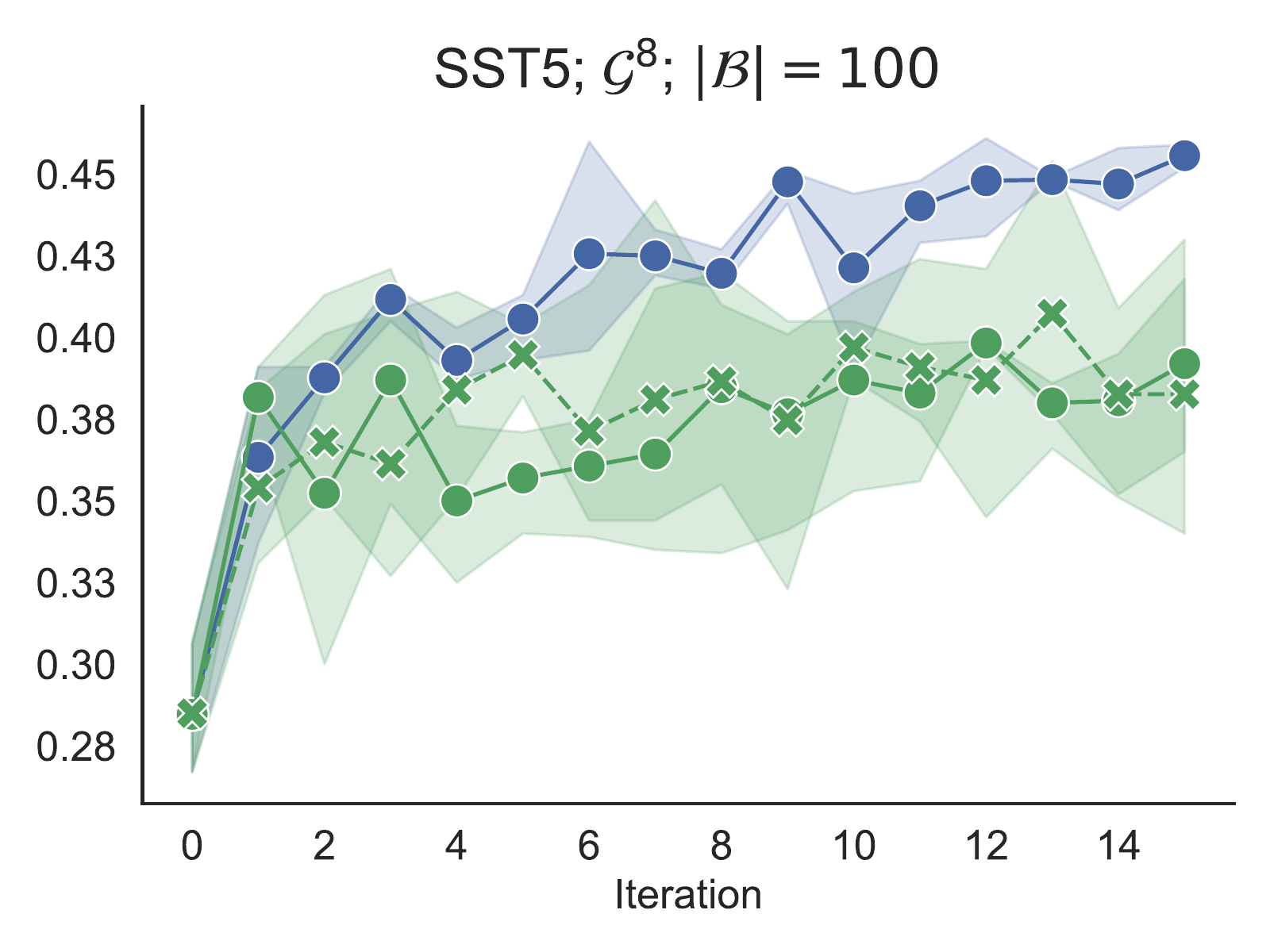}
}
\vspace{-.2cm}\subfloat{
\includegraphics[width=.5\linewidth,height=0.2\textwidth]{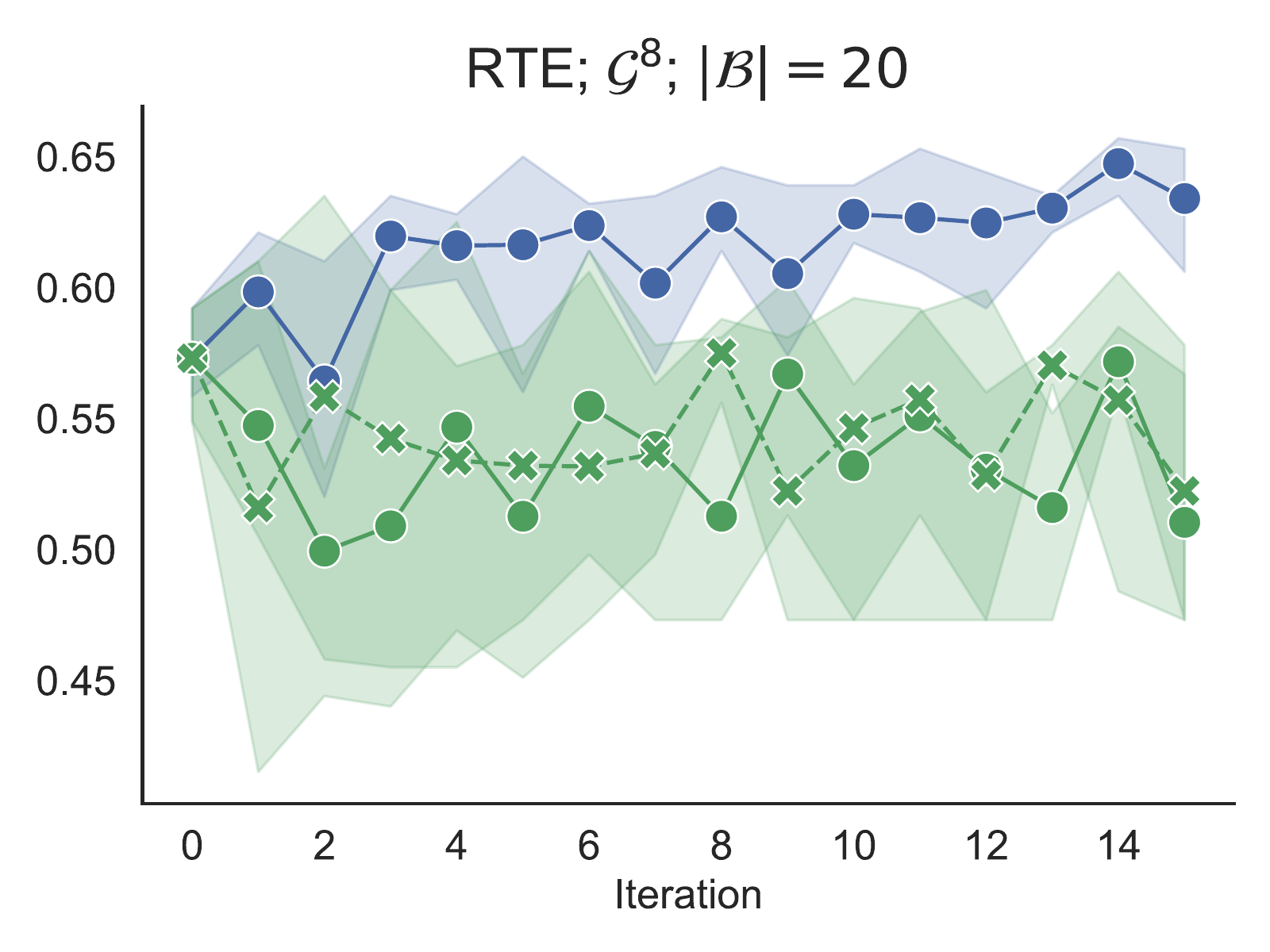}
}
\caption{
  Performance of \textcolor{blue}{AL}
  and \textcolor{green}{\md}
  with  discrete labels ($\bullet$) vs.\
  with KL divergence (\tiny\XSolidBold\normalsize).
  Entropy is used as the sampling strategy
  and WeightedLogitVoting is used 
  to aggregate worker annotations.
}
\figlabel{kllogits}
\end{figure}

\subsection{Design choice 3: Distilling logits}
We can view \md as a kind of distillation \citep{hinton2015distilling}:
The ability of
\mdrs to solve $\mathcal{T}$ is progressively transferred 
to $\mathcal{S}$.
In this section, we explore 
the utility of distillation:
We train
$\mathcal{S}$
with predicted logits\footnote{
Distilling with intermediate activations likely
to further improve performance of $\mathcal{S}$.
However, note that PLM intermediate activations
are not always available in a
Language-Model-as-a-Service framework.
}
instead of discrete labels
from \mdrs.
Concretely, we 
train
$\mathcal{S}$ by reducing the
KL divergence between its
predicted probability distribution
(over the label set)
and the
probability distribution
from \mdrs.

\figref{kllogits} shows that
training
$\mathcal{S}$ with KL divergence
noticeably improves
over discrete labels
on AGNews and SST5.
This is expected:
AGNews and SST5 have larger
label set size (four and five)
such that the
probability distribution over the label set
is more informative than
that of
the binary classification tasks
SST2 and RTE.

\begin{figure}[t]
\centering
\subfloat{
\includegraphics[width=.5\linewidth,height=0.2\textwidth]{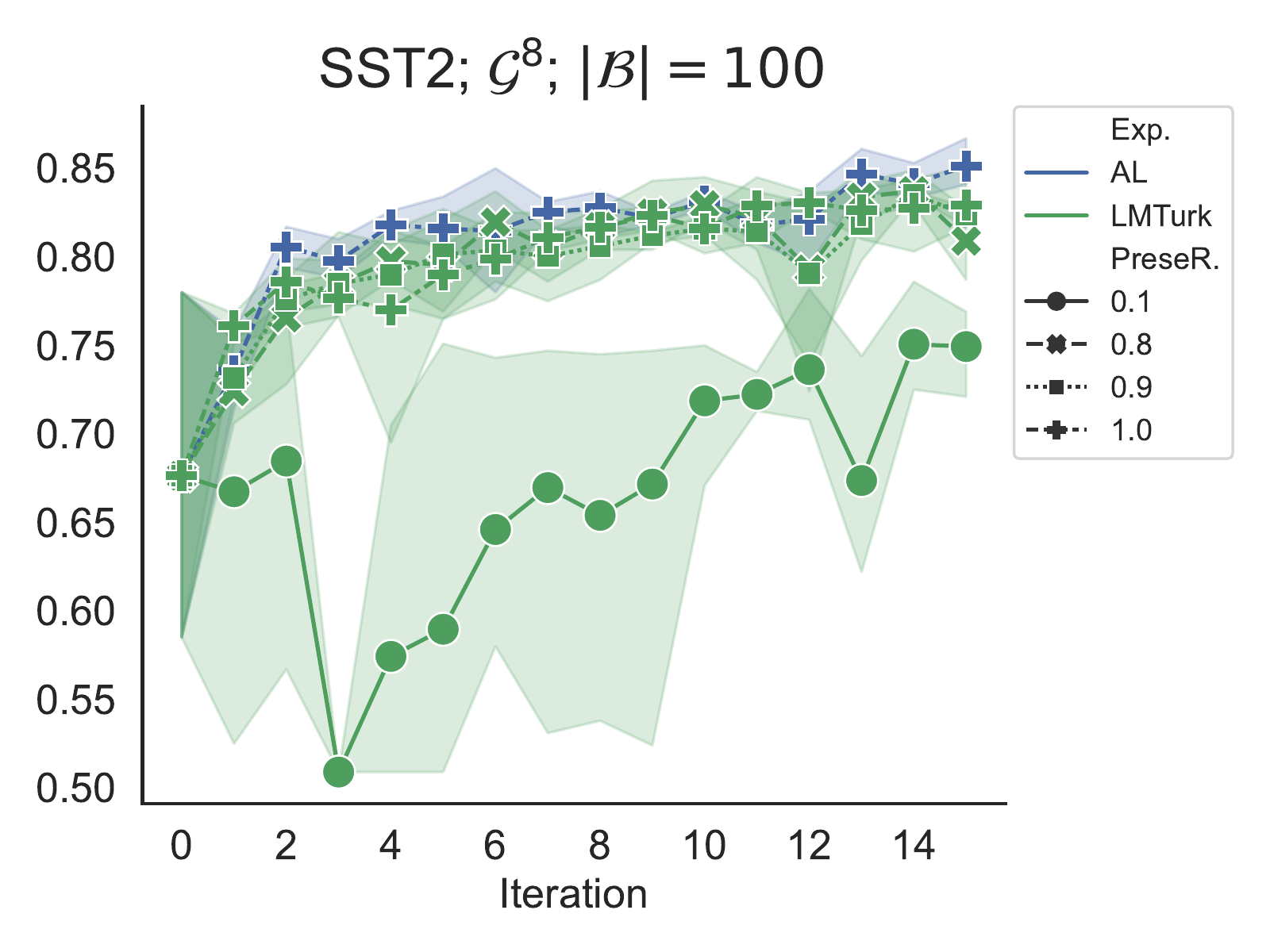}
}
\vspace{-.25cm}\subfloat{
\hspace{-.25cm}
\includegraphics[width=.5\linewidth,height=0.2\textwidth]{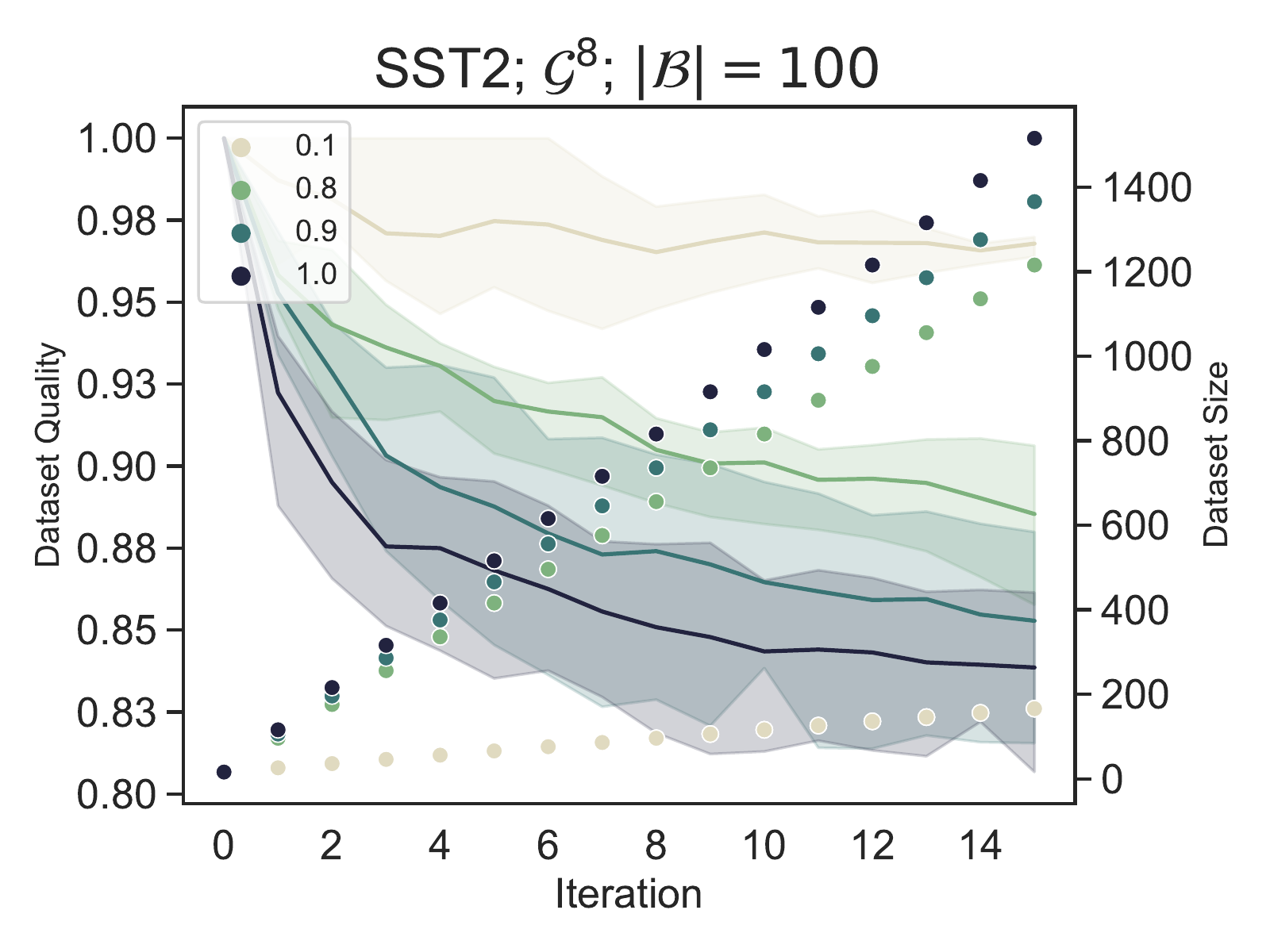}
}
\\
\vspace{-.25cm}\subfloat{
\includegraphics[width=.5\linewidth,height=0.2\textwidth]{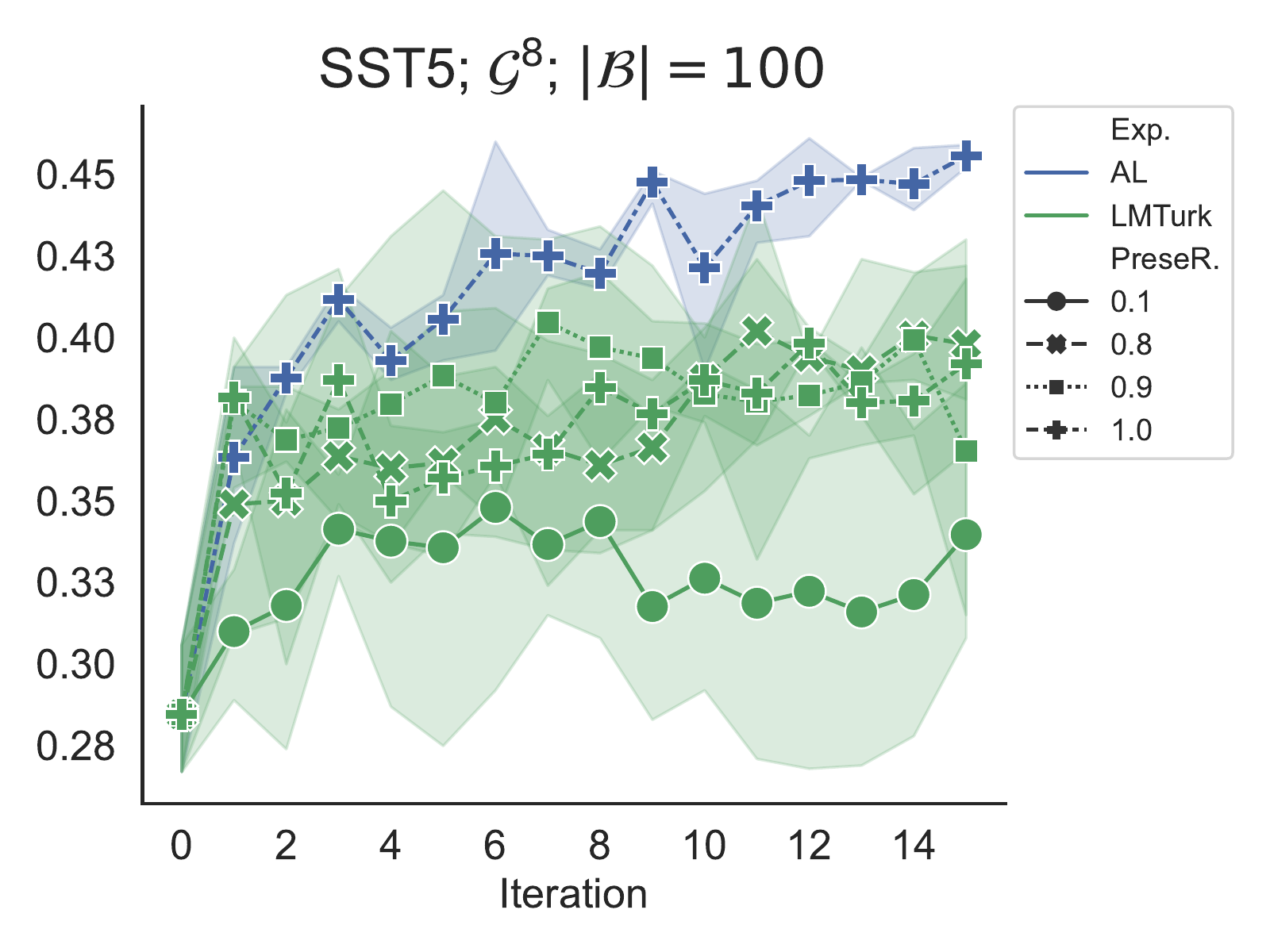}
}
\vspace{-.25cm}\subfloat{
\hspace{-.25cm}    
\includegraphics[width=.5\linewidth,height=0.2\textwidth]{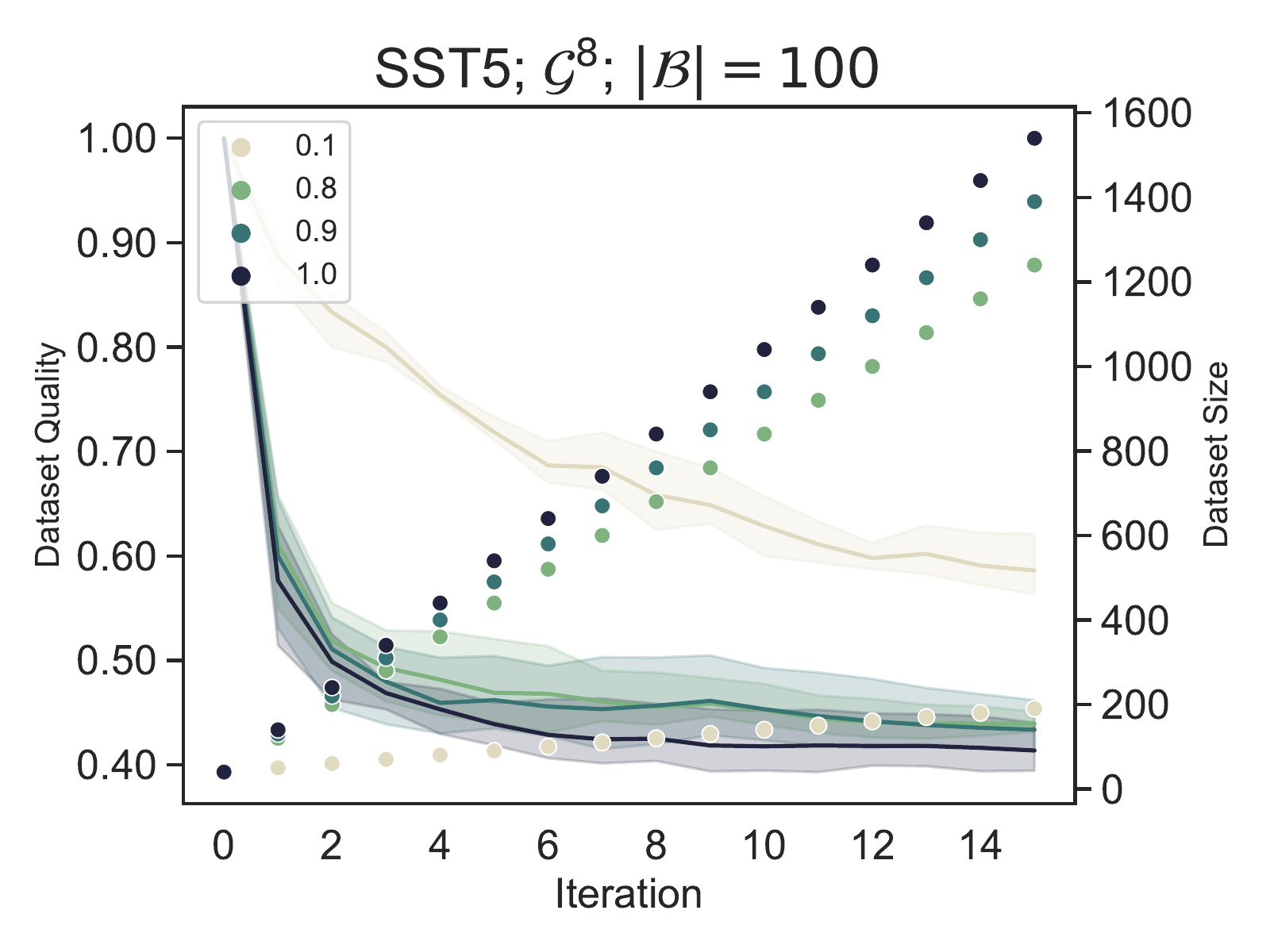}
}
\\
\vspace{-.25cm}\subfloat{
\includegraphics[width=.5\linewidth,height=0.2\textwidth]{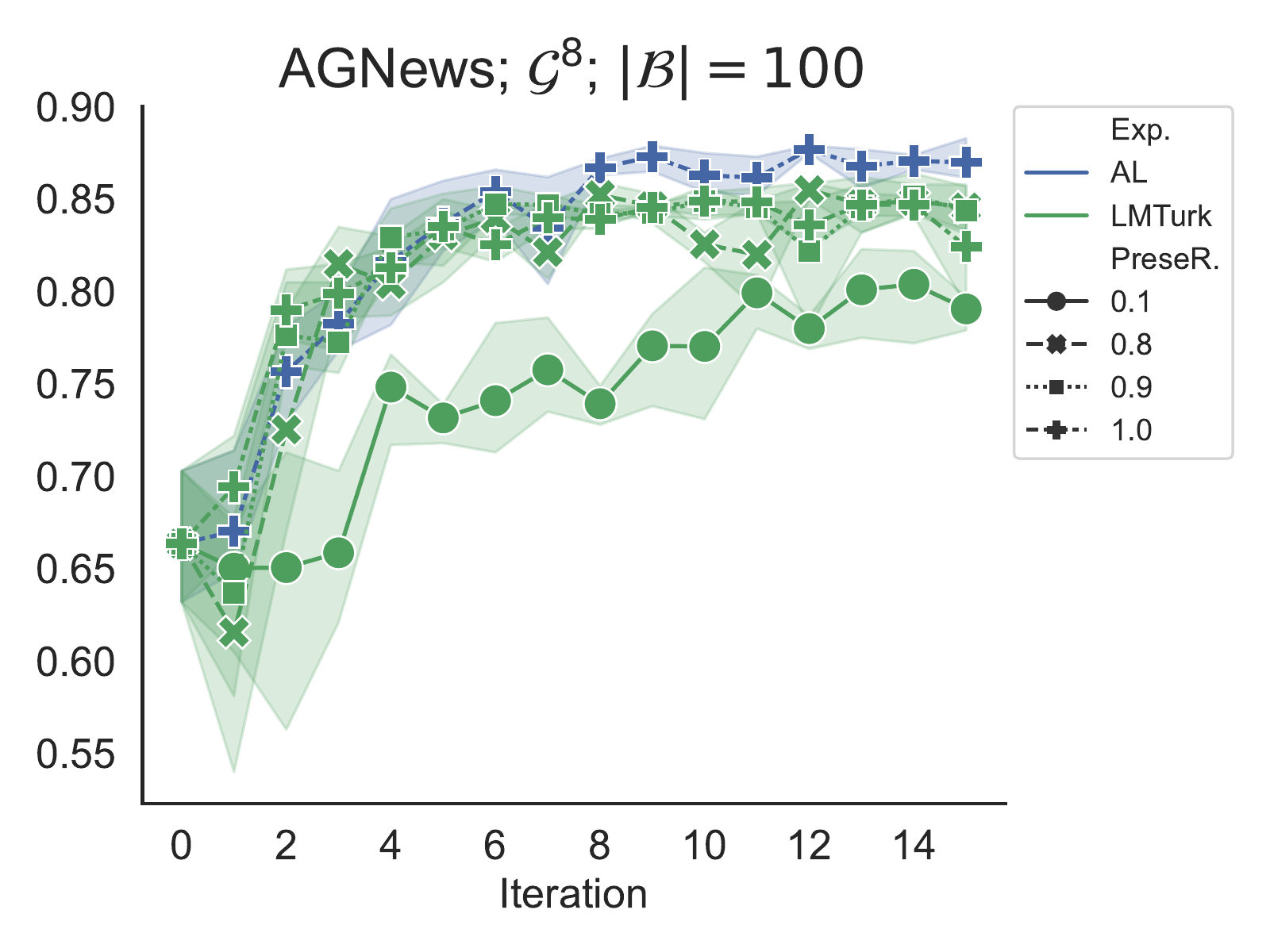}
}
\vspace{-.25cm}\subfloat{
\hspace{-.25cm}  
\includegraphics[width=.5\linewidth,height=0.2\textwidth]{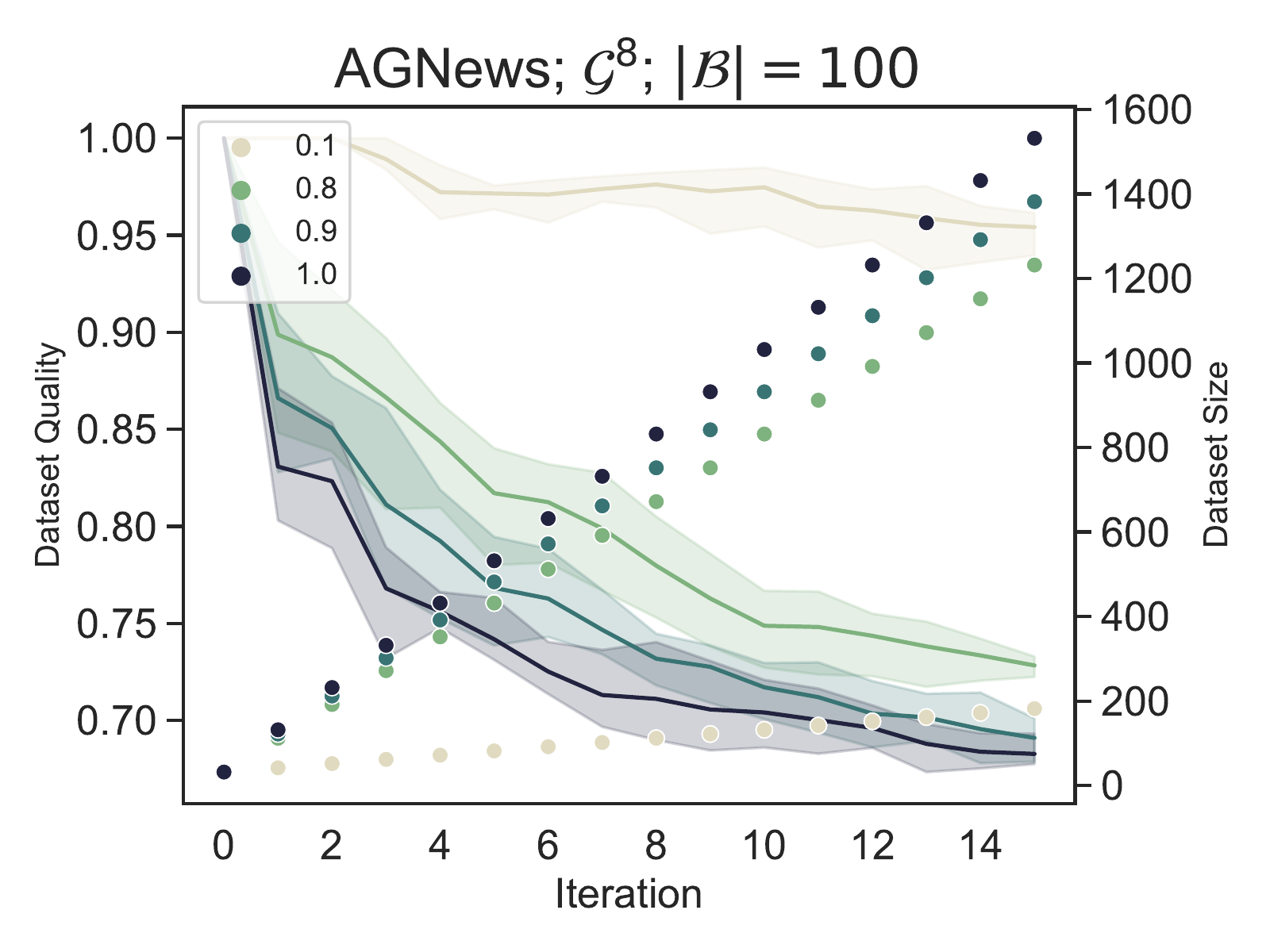}
}
\caption{
  Training $\mathcal{S}$
  with
  examples for which \mdrs have low entropy.
  We report performance of $\mathcal{S}$ (left),
  number and quality (measured by accuracy) of the
  preserved examples (right) at each iteration.
}
\figlabel{filtering}
\end{figure}

\subsection{Design choice 4: Quality-based filtering}
\seclabel{poolfilter}
One key difference between AL and \md
is that \mdrs are not oracles:
Their labels are not perfect.
Hence, it is reasonable to
consider processing the training
data, denoted as $\mathcal{D}^{j}$, for $\mathcal{S}^{j}$, instead
of using it indiscriminately as in AL.

\textbf{InstanceTresholding} (\secref{instancetresholdingintro})
preserves
annotations
in $\mathcal{D}^{j}$
for which \mdrs
have the smallest
prediction entropy.
Concretely,
we rank all annotations
$(\mathbf{x}, \hat{y}, \mathbf{l}) \in \mathcal{D}^{j}$
by \emph{entropy}($\mathbf{l}$)
and then keep the
$\tau$ percent smallest.
Note that we always preserve the human-labeled
few-shot data $\mathcal{G}_{train}$.
We experiment
with $\tau \in$
\{10\%, \ldots, 90\%, 100\%\}.

\figref{filtering} left shows the
performance of $\mathcal{S}$;
\figref{filtering} right 
tracks the status of
$\mathcal{D}^{j}$.
To measure quality, we compute the accuracy
of \mdr annotations
on $\mathcal{D}^{j}$ (compared to gold labels);
see the lineplots and the left y-axis.
We also report the size
of $\mathcal{D}^{j}$ as
scatter plots (right y-axis).

We observe 
that $\tau$=10\%, i.e., keeping only
the 10\% most certain examples,
gives the worst performance.
This is most obvious at iteration three
for  SST2: The performance
drops to near the majority baseline ($\approx$50\%).
This is because $\mathcal{D}^{3}$
is small and unbalanced: It has
eight negative (from $\mathcal{G}^{train}$)
and 38 positive examples.
However, using
all the \mdr annotations
($\tau$=100\%) may not
be optimal either.
This is noticeable
when looking at SST5:
$\tau$=90\% and $\tau$=80\%
are better options.

We see that there is a trade-off between
$\mathcal{D}^{j}$'s quality and size
from \figref{filtering} right.
Being conservative, i.e.,
preserving
only a handful of
annotations from
\mdrs, results in a small, but high-quality
$\mathcal{D}^{j}$;
using all the annotations
indiscriminately leads to a 
large $\mathcal{D}^{j}$ with low quality.
This experiment
highlights a key difference
between \md and AL:
\mdr annotations 
are not perfect and
taking the annotation quality
into consideration when training 
$\mathcal{S}$ is crucial.

\section{Conclusion}
In this work, our focus is the research
question: \emph{
  How to make effective use
  of current few-shot learners?
} We propose \md,
a simple yet effective method 
that considers PLM-based few-shot learners
as non-expert annotators in crowdsourcing;
active learning strategies are incorporated
to reduce the cost of annotation.
We further show that
processing the annotations from \mdrs
can  be beneficial.

Future work may combine
\mdr annotations with 
human annotators in a human-in-the-loop setup \cite{monarch2021human} to increase the overall utility
of invested resources \citep{bai2021pre}.
Scaling up to even larger PLMs
likely to further
boost model performances \citep{kaplan2020scaling,GPT3paper}
Applying \md to
multilingual few-shot
learners \citep{zhao-etal-2021-closer,winata2021language,lin2021few}
is also
promising.

\section*{Acknowledgements}
We thank the anonymous reviewers for their insightful comments and suggestions.
MZ and HS were supported by the European Research Council
(ERC\# 740516)
and the 
German Federal Ministry of Education and
Research (BMBF, grant \#01IS18036A).

\bibliography{custom}
\clearpage
\appendix
\section{Reproducibility Checklist}
\seclabel{appendix:checklist}
\subsection{Computing infrastructure}
We use four Tesla V100 GPUs to prompt
each of the \mdrs, and
a single Tesla V100 GPU
is used when finetuning the small model
$\mathcal{S}$.

\subsection{Datasets}
For SST2, CoLA, and RTE, we use the official datasets
available on the benchmark website
\url{gluebenchmark.com}.
We download
SST5 dataset from
\url{nlp.stanford.edu/sentiment}
and AGNews
from the link provided by \citet{agdataset}.

The number of testing examples of each dataset
is shown in \tabref{appendix:numtestegs}.
Note that for SST2, CoLA, and RTE,
$\mathcal{G}^{dev}$ is sampled from
the training set, and the dev set is used as the test set.

\begin{table}[h]
\small\centering
\begin{tabular}{|c|c|c|c|c|}
\hline
CoLA & SST5 & RTE & AGNews & SST2 \\ \hline
1042 & 2210 & 277 & 7600   & 872  \\ \hline
\end{tabular}
\caption{Number of testing examples.}
\tablabel{appendix:numtestegs}
\end{table}

\begin{table*}[h]
\centering\scriptsize\renewcommand{\arraystretch}{1.2}\setlength{\tabcolsep}{4pt}
\begin{tabular}{c|cc|cc|cc|}
\cline{2-7}
                             & \multicolumn{2}{c|}{$\mathcal{G}^8$}                    & \multicolumn{2}{c|}{$\mathcal{G}^{16}$}                 & \multicolumn{2}{c|}{$\mathcal{G}^{32}$}                    \\ \cline{2-7} 
                             & \multicolumn{1}{c|}{Workers}        &  $\mathcal{S}$    & \multicolumn{1}{c|}{Workers}        & $\mathcal{S}$     & \multicolumn{1}{c|}{Workers}        & $\mathcal{S}$             \\ \hline
\multicolumn{1}{|c|}{}       & \multicolumn{1}{c|}{91.13$\pm$0.52} &                   & \multicolumn{1}{c|}{91.93$\pm$1.09} &                   & \multicolumn{1}{c|}{91.97$\pm$0.83} &                   \\
\multicolumn{1}{|c|}{}       & \multicolumn{1}{c|}{91.63$\pm$0.68} &                   & \multicolumn{1}{c|}{93.08$\pm$0.62} &                   & \multicolumn{1}{c|}{91.70$\pm$1.78} &                   \\
\multicolumn{1}{|c|}{SST2}   & \multicolumn{1}{c|}{90.18$\pm$1.00} & 67.63$\pm$8.01    & \multicolumn{1}{c|}{91.74$\pm$1.04} & 75.83$\pm$1.35    & \multicolumn{1}{c|}{91.21$\pm$1.83} & 76.37$\pm$3.16 \\
\multicolumn{1}{|c|}{}       & \multicolumn{1}{c|}{90.83$\pm$0.58} &                   & \multicolumn{1}{c|}{90.79$\pm$0.47} &                   & \multicolumn{1}{c|}{91.13$\pm$0.24} &                   \\
\multicolumn{1}{|c|}{}       & \multicolumn{1}{c|}{90.52$\pm$1.84} &                   & \multicolumn{1}{c|}{91.67$\pm$1.36} &                   & \multicolumn{1}{c|}{93.23$\pm$0.37} &                   \\ \hline
\multicolumn{1}{|c|}{}       & \multicolumn{1}{c|}{41.37$\pm$1.55} &                   & \multicolumn{1}{c|}{45.16$\pm$2.13} &                   & \multicolumn{1}{c|}{45.91$\pm$0.96} &                   \\
\multicolumn{1}{|c|}{}       & \multicolumn{1}{c|}{42.32$\pm$2.04} &                   & \multicolumn{1}{c|}{45.96$\pm$2.12} &                   & \multicolumn{1}{c|}{48.64$\pm$0.59} &                   \\
\multicolumn{1}{|c|}{SST5}   & \multicolumn{1}{c|}{40.57$\pm$2.70} & 28.47$\pm$1.61    & \multicolumn{1}{c|}{46.70$\pm$0.93} & 34.97$\pm$1.51    & \multicolumn{1}{c|}{50.53$\pm$0.94} & 33.47$\pm$2.79 \\
\multicolumn{1}{|c|}{}       & \multicolumn{1}{c|}{37.69$\pm$1.34} &                   & \multicolumn{1}{c|}{42.53$\pm$2.43} &                   & \multicolumn{1}{c|}{43.32$\pm$3.42} &                   \\
\multicolumn{1}{|c|}{}       & \multicolumn{1}{c|}{38.05$\pm$2.60} &                   & \multicolumn{1}{c|}{42.96$\pm$0.69} &                   & \multicolumn{1}{c|}{45.72$\pm$1.43} &                   \\ \hline
\multicolumn{1}{|c|}{}       & \multicolumn{1}{c|}{68.95$\pm$1.47} &                   & \multicolumn{1}{c|}{68.35$\pm$2.29} &                   & \multicolumn{1}{c|}{71.72$\pm$1.96} &                   \\
\multicolumn{1}{|c|}{}       & \multicolumn{1}{c|}{54.99$\pm$3.76} &                   & \multicolumn{1}{c|}{57.64$\pm$3.23} &                   & \multicolumn{1}{c|}{58.48$\pm$3.59} &                   \\
\multicolumn{1}{|c|}{RTE}    & \multicolumn{1}{c|}{62.70$\pm$1.33} & 57.30$\pm$1.79    & \multicolumn{1}{c|}{70.88$\pm$1.70} & 61.50$\pm$0.78    & \multicolumn{1}{c|}{68.47$\pm$1.19} & 62.93$\pm$0.74 \\
\multicolumn{1}{|c|}{}       & \multicolumn{1}{c|}{50.42$\pm$2.07} &                   & \multicolumn{1}{c|}{58.60$\pm$1.62} &                   & \multicolumn{1}{c|}{59.33$\pm$4.72} &                   \\
\multicolumn{1}{|c|}{}       & \multicolumn{1}{c|}{51.99$\pm$4.45} &                   & \multicolumn{1}{c|}{57.88$\pm$2.83} &                   & \multicolumn{1}{c|}{60.41$\pm$2.47} &                   \\ \hline
\multicolumn{1}{|c|}{}       & \multicolumn{1}{c|}{75.39$\pm$5.25} &                   & \multicolumn{1}{c|}{83.06$\pm$0.83} &                   & \multicolumn{1}{c|}{84.92$\pm$0.28} &                   \\
\multicolumn{1}{|c|}{}       & \multicolumn{1}{c|}{85.40$\pm$1.43} &                   & \multicolumn{1}{c|}{87.71$\pm$0.07} &                   & \multicolumn{1}{c|}{87.79$\pm$1.08} &                   \\
\multicolumn{1}{|c|}{AGNews} & \multicolumn{1}{c|}{78.83$\pm$4.77} & 66.37$\pm$2.95    & \multicolumn{1}{c|}{83.59$\pm$2.96} & 69.40$\pm$0.93    & \multicolumn{1}{c|}{87.39$\pm$1.29} & 76.53$\pm$0.41 \\
\multicolumn{1}{|c|}{}       & \multicolumn{1}{c|}{85.07$\pm$1.09} &                   & \multicolumn{1}{c|}{87.69$\pm$0.04} &                   & \multicolumn{1}{c|}{87.17$\pm$0.67} &                   \\
\multicolumn{1}{|c|}{}       & \multicolumn{1}{c|}{79.95$\pm$0.86} &                   & \multicolumn{1}{c|}{80.15$\pm$3.38} &                   & \multicolumn{1}{c|}{83.32$\pm$0.59} &                   \\ \hline
\multicolumn{1}{|c|}{}       & \multicolumn{1}{c|}{ 0.14$\pm$1.43} &                   & \multicolumn{1}{c|}{11.81$\pm$7.82} &                   & \multicolumn{1}{c|}{19.88$\pm$3.30} &                   \\
\multicolumn{1}{|c|}{}       & \multicolumn{1}{c|}{ 2.42$\pm$4.84} &                   & \multicolumn{1}{c|}{15.23$\pm$7.07} &                   & \multicolumn{1}{c|}{22.51$\pm$0.96} &                   \\
\multicolumn{1}{|c|}{CoLA}   & \multicolumn{1}{c|}{ 7.40$\pm$8.12} & 0.97$\pm$4.40     & \multicolumn{1}{c|}{19.71$\pm$1.89} & 4.27$\pm$3.26     & \multicolumn{1}{c|}{26.34$\pm$1.54} & 2.50$\pm$2.41 \\
\multicolumn{1}{|c|}{}       & \multicolumn{1}{c|}{ 9.91$\pm$7.98} &                   & \multicolumn{1}{c|}{17.14$\pm$2.48} &                   & \multicolumn{1}{c|}{18.15$\pm$0.63} &                   \\
\multicolumn{1}{|c|}{}       & \multicolumn{1}{c|}{15.33$\pm$2.15} &                   & \multicolumn{1}{c|}{19.66$\pm$0.48} &                   & \multicolumn{1}{c|}{27.58$\pm$7.09} &                   \\ \hline
\end{tabular}
\caption{Few-shot performance of the five \mdrs
  and the small model $\mathcal{S}$.
  Each experiment is repeated three times and we report mean and standard deviation.}
\tablabel{appendix:fewshotnumeric}
\end{table*}


\section{Numerical Results}
\seclabel{appendix:numeric}
\tabref{appendix:fewshotnumeric}
reports the numerical
value of \figref{fewshotperf}.

\section{Prompting Details}
\seclabel{appendix:promptingdetails}
For each task, 
we list the five prompts
employed to adapt a PLM  to a \mdr.
``[v]'' is a prompting token whose
trainable embedding vector is 
randomly initialized.

For \textbf{SST5}, we use following prompts:
\begin{itemize}
\item ``[v] \textbf{x} It is [MASK].''
\item ``[v] \textbf{x} Such a [MASK] movie.''
\item ``\textbf{x} [v] It is pretty [MASK].''
\item ``It is [MASK] because \textbf{x} [v]''
\item ``\textbf{x} So it is [MASK]. [v]''
\end{itemize}
and the PLM picks a word from
\{``crap'', ``bad'', ``normal'', ``good'', ``perfect''\}.
to fill the position of ``[MASK]''.
The mapping
\{``crap''$\,\to\,$1,
``bad''$\,\to\,$2,
``normal''$\,\to\,$3,
``good''$\,\to\,$4,
``perfect''$\,\to\,$5
\} is used to convert
model predictions to numerical values.

For \textbf{SST2}, we use following prompts:
\begin{itemize}
\item ``[v] \textbf{x} It is [MASK].''
\item ``[v] \textbf{x} Such a [MASK] movie.''
\item ``\textbf{x} [v] It is pretty [MASK].''
\item ``It is [MASK] because \textbf{x} [v]''
\item ``\textbf{x} So it is [MASK]. [v]''
\end{itemize}
and the PLM picks a word from
\{``bad'', ``good''\}
to fill the position of ``[MASK]''.
The mapping
\{``bad''$\,\to\,$0, ``good''$\,\to\,$1\}
is used.

For \textbf{AGNews}, we use following prompts:
\begin{itemize}
\item ``[v] \textbf{x} It is about [MASK].''
\item ``\textbf{x} [v] Topic: [MASK].''
\item ``\textbf{x} [v] The text is about [MASK].''
\item ``\textbf{x} Topic: [MASK]. [v]''
\item ``\textbf{x} [v] [MASK].''
\end{itemize}
and the PLM picks a word from
\{``world'', ``sports'', ``economy'', ``technology''\}
to fill the position of ``[MASK]''.
The mapping
\{``world''$\,\to\,$1,
``sports''$\,\to\,$2,
``economy''$\,\to\,$3,
``technology''$\,\to\,$4
\} is used.

For \textbf{CoLA}, we use following prompts:
\begin{itemize}
\item ``[v] \textbf{x} It sounds [MASK].''
\item ``[v] \textbf{x} The sentence is [MASK].''
\item ``[v] \textbf{x} It is a [MASK] sentence.''
\item ``\textbf{x} [v] [MASK].''
\item ``[v] \textbf{x} [MASK].''
\end{itemize}
and the PLM picks a word from
\{``wrong'', ``ok''\}
to fill the position of ``[MASK]''.
The mapping
\{``wrong''$\,\to\,$0,
``okay''$\,\to\,$1\}
is used.

For \textbf{RTE}, we use following prompts:
\begin{itemize}
\item ``\textbf{p} Question: \textbf{h}? [v] Answer: [MASK].''
\item ``\textbf{p} [SEP] \textbf{h}? [MASK]. [v]''
\item ``\textbf{p} [SEP] \textbf{h}? [v] answer: [MASK].''
\item ``\textbf{p} [SEP] In short \textbf{h}. [MASK]. [v]''
\item ``[v] \textbf{p} [SEP] In short \textbf{h}. [MASK].''
\end{itemize}
where \textbf{p} and \textbf{h}
refer to premise and hypothesis.
The PLM picks a word from
\{``No'', ``Yes''\}
to fill the position of ``[MASK]''.
The mapping
\{``No''$\,\to\,$0,
``Yes''$\,\to\,$1\}
is used.

\begin{figure}[t]
\centering
\vspace{-.25cm}\subfloat{
\includegraphics[width=.5\linewidth,height=0.18\textwidth]{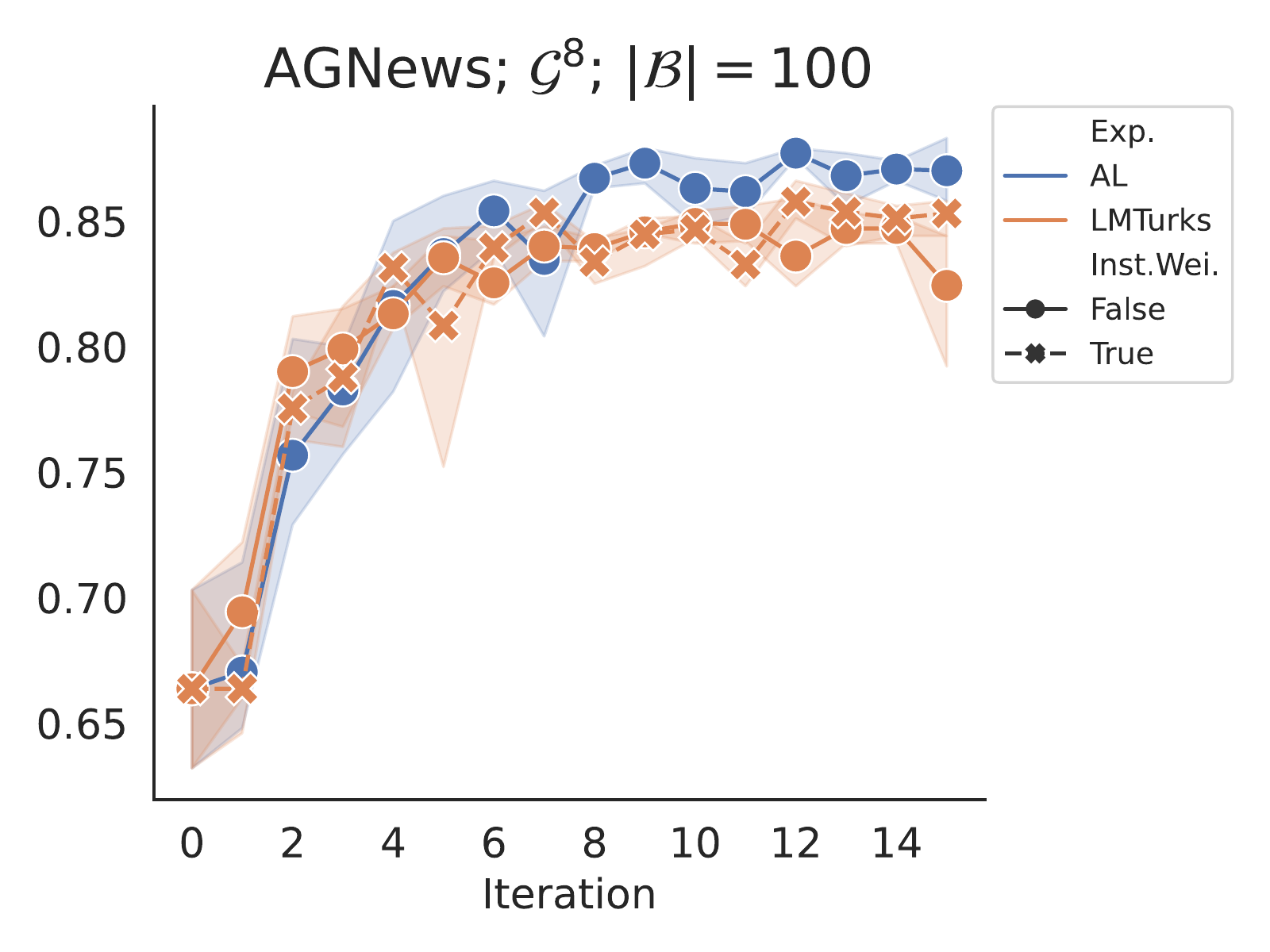}
}
\vspace{-.25cm}\subfloat{
\hspace{-.25cm}
\includegraphics[width=.5\linewidth,height=0.18\textwidth]{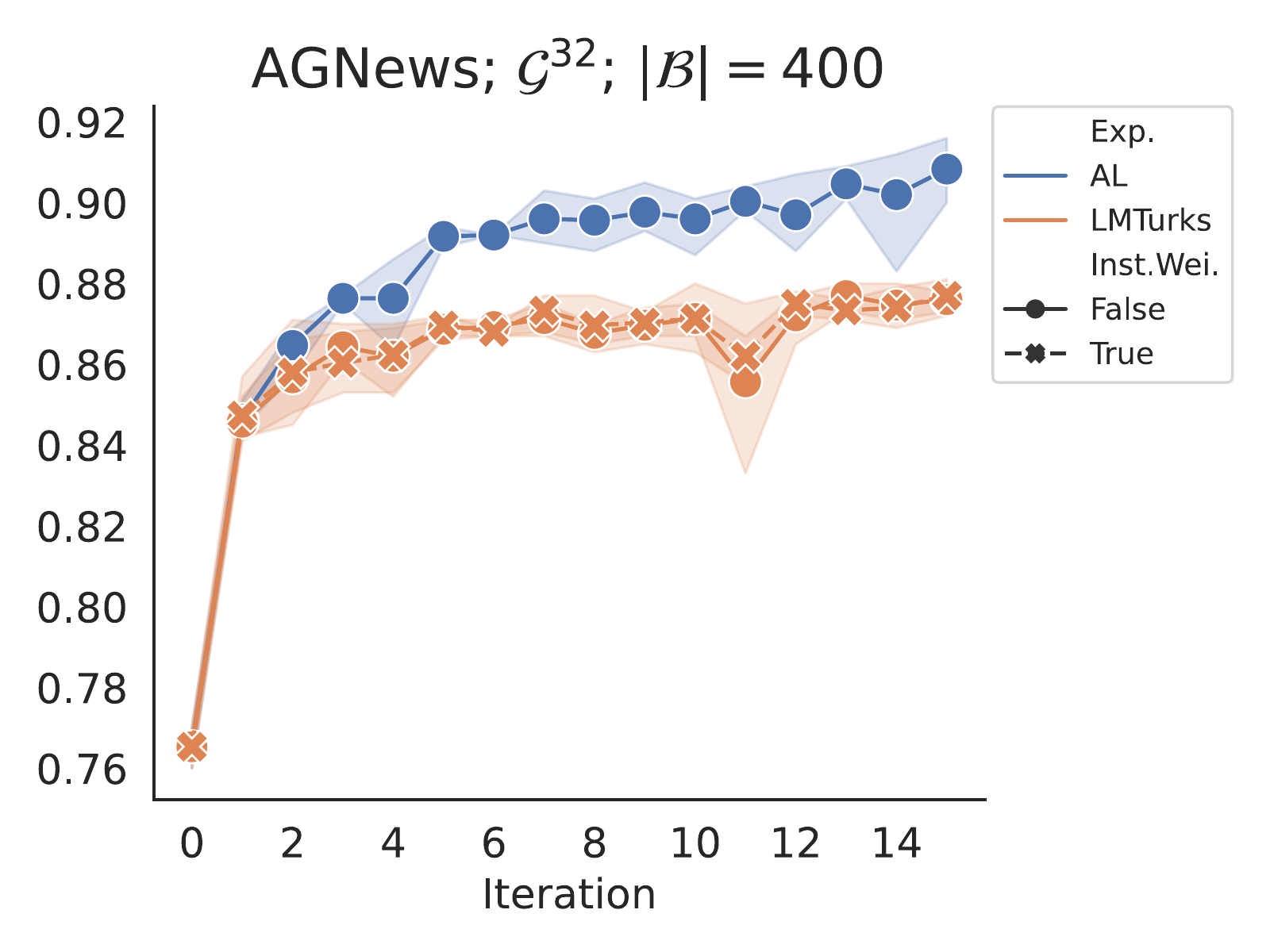}
}
\\
\vspace{-.25cm}\subfloat{
\includegraphics[width=.5\linewidth,height=0.18\textwidth]{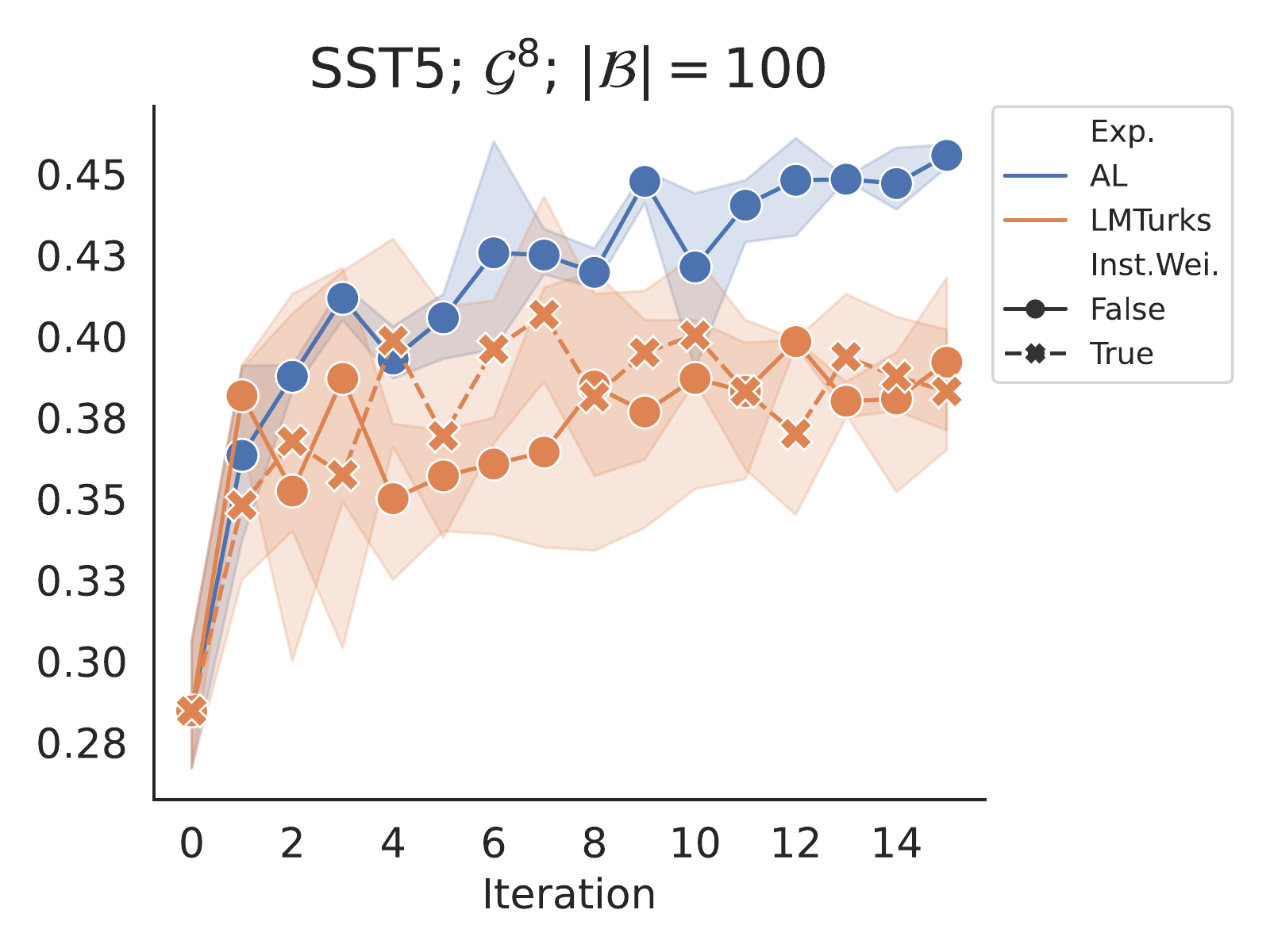}
}
\vspace{-.25cm}\subfloat{
\hspace{-.25cm}    
\includegraphics[width=.5\linewidth,height=0.18\textwidth]{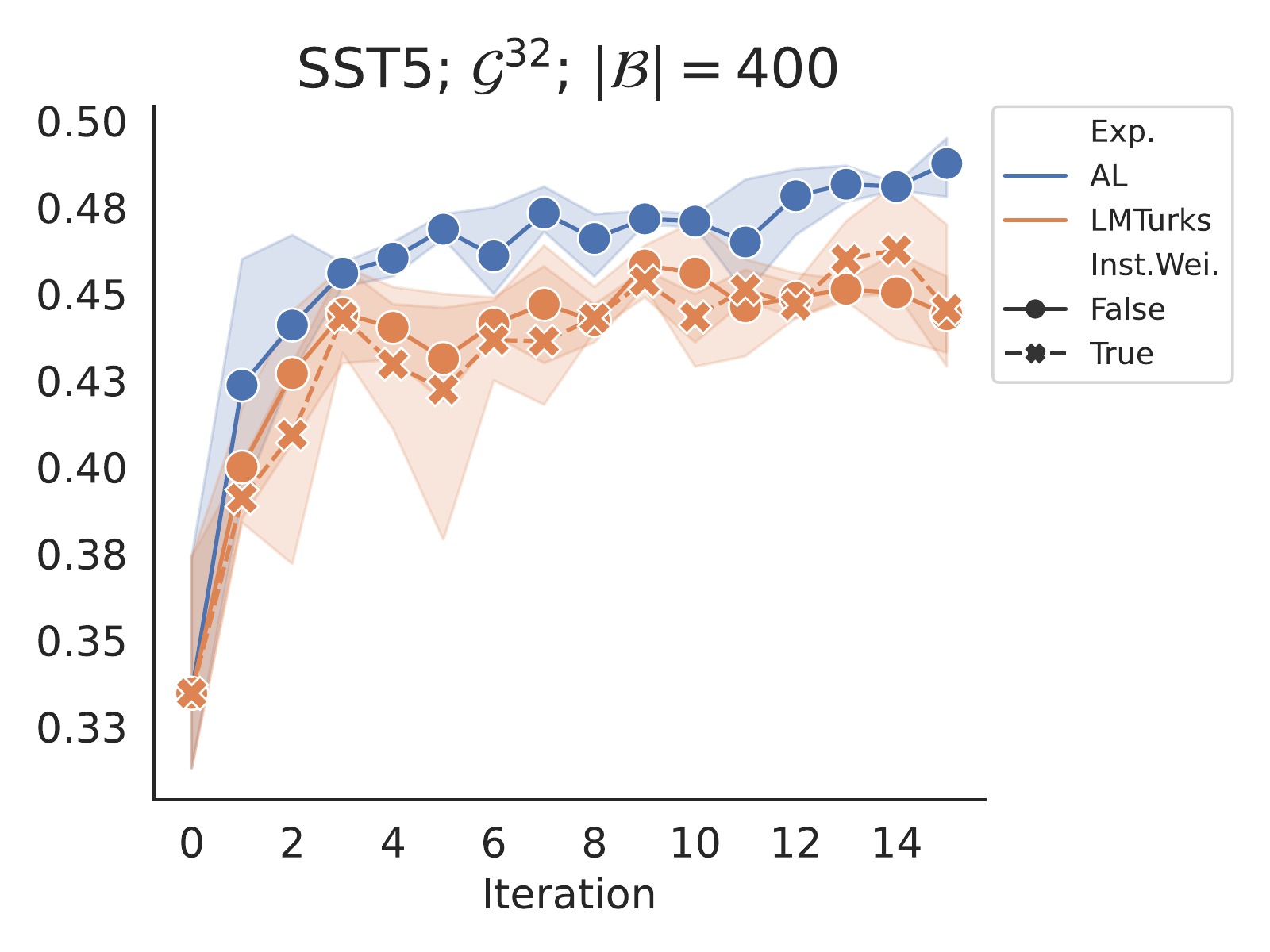}
}
\\
\vspace{-.25cm}\subfloat{
\includegraphics[width=.5\linewidth,height=0.18\textwidth]{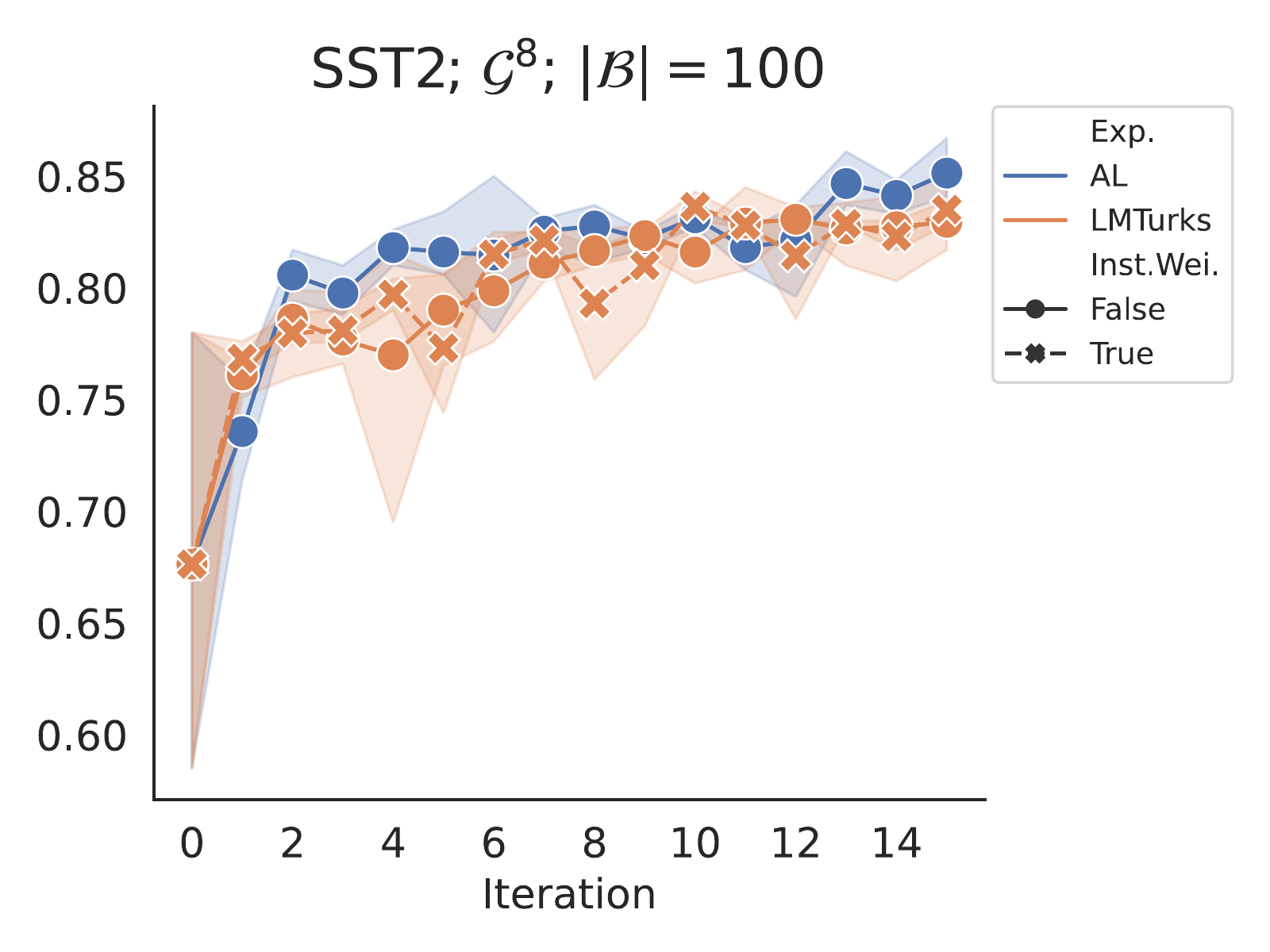}
}
\vspace{-.25cm}\subfloat{
\hspace{-.25cm}  
\includegraphics[width=.5\linewidth,height=0.18\textwidth]{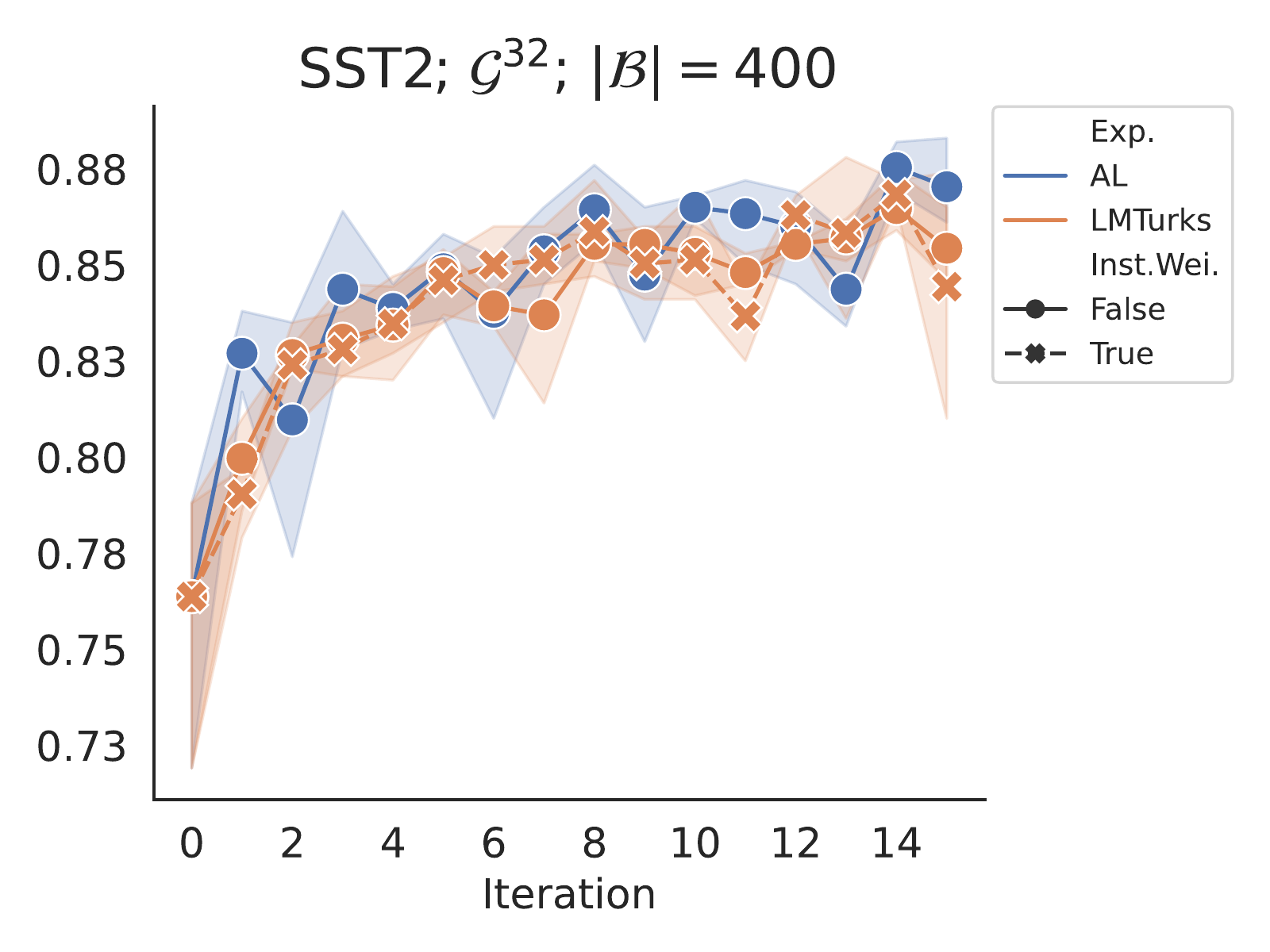}
}
\caption{
Weighting the training instances from \mdrs.
}
\figlabel{instanceweighting}
\end{figure}

\section{More Visualizations}
\seclabel{appendix:morevisualization}
\figref{appendix:completeiteratives} visualizes the performance
of $\mathcal{S}$ when different $|\mathcal{G}|$ and
$|\mathcal{B}|$ are used.

\begin{figure*}[t]
\centering
\hspace{-.2cm}\subfloat{
\includegraphics[width=.25\linewidth,height=0.2\textwidth]{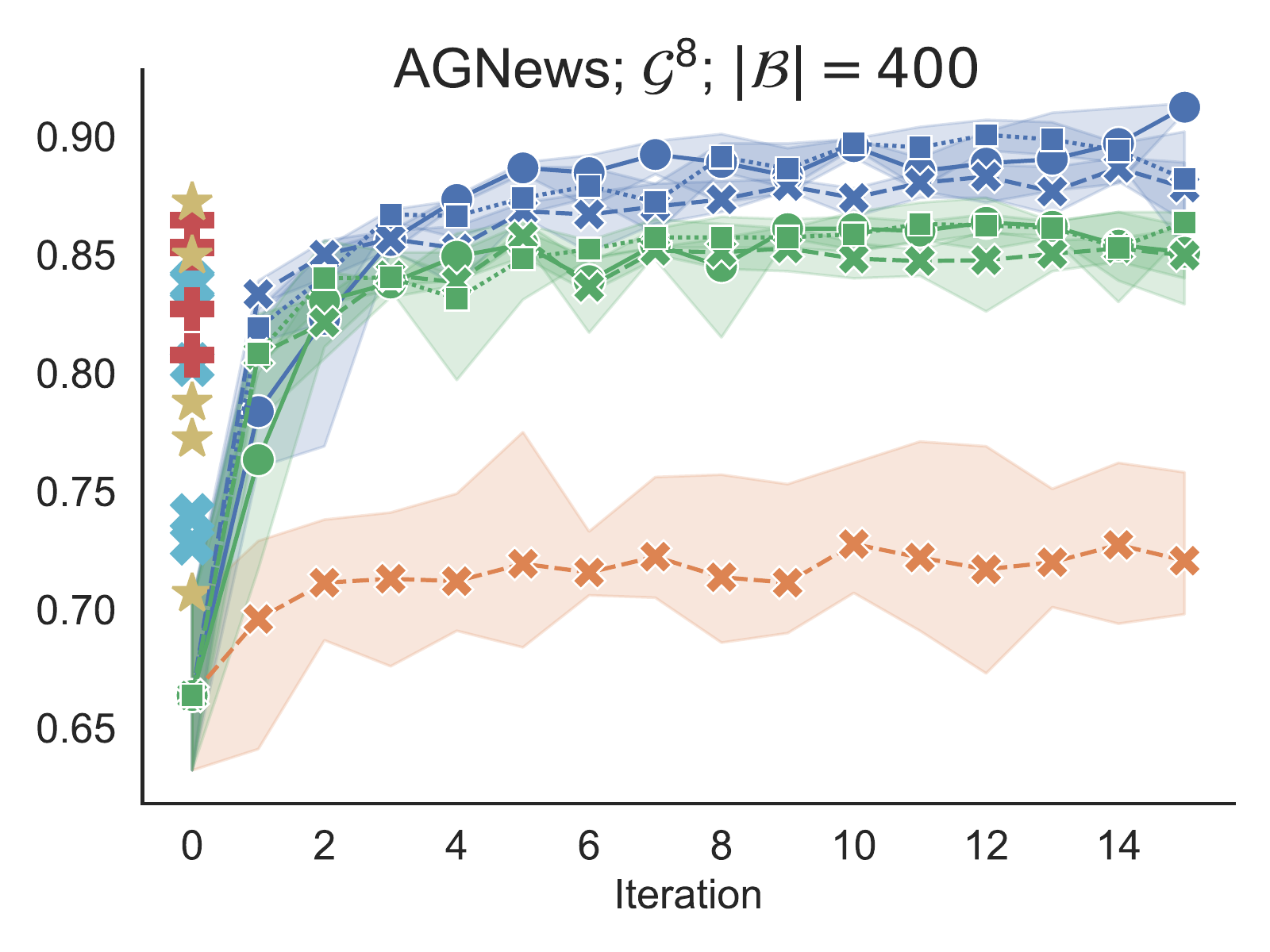}
}
\hspace{-.2cm}\subfloat{
\includegraphics[width=.25\linewidth,height=0.2\textwidth]{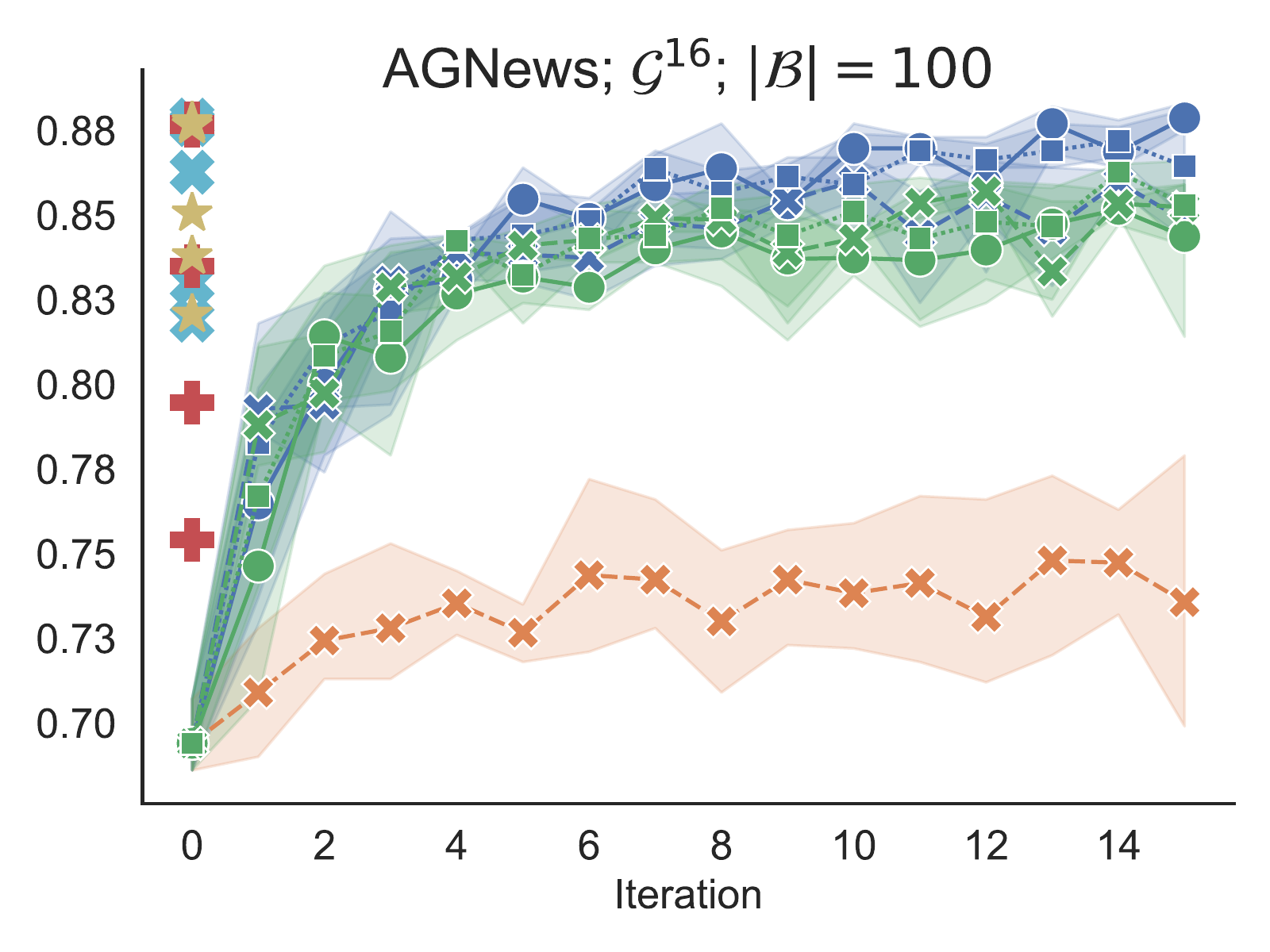}
}
\hspace{-.2cm}\subfloat{
\includegraphics[width=.25\linewidth,height=0.2\textwidth]{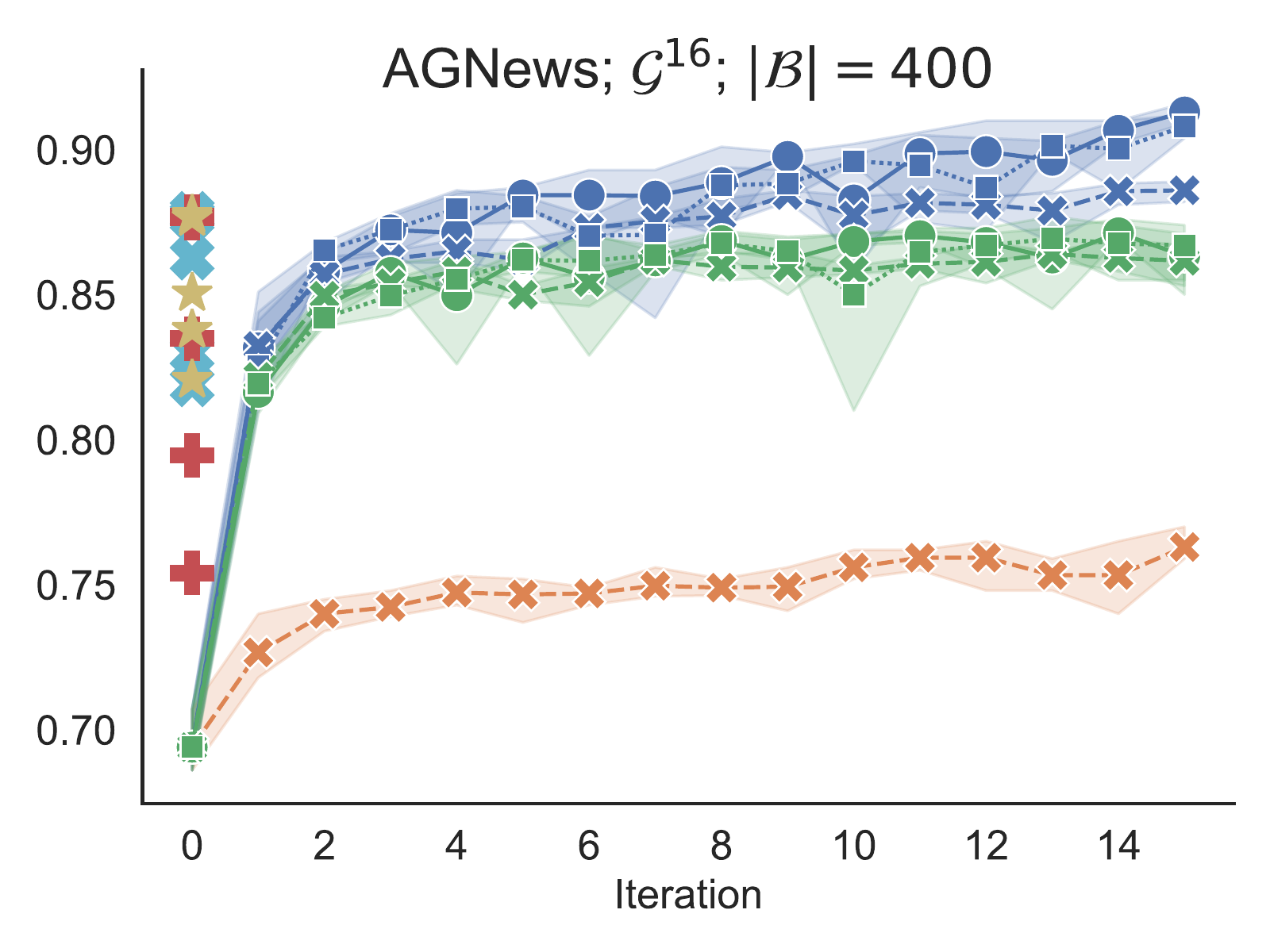}
}
\hspace{-.2cm}\subfloat{
\includegraphics[width=.25\linewidth,height=0.2\textwidth]{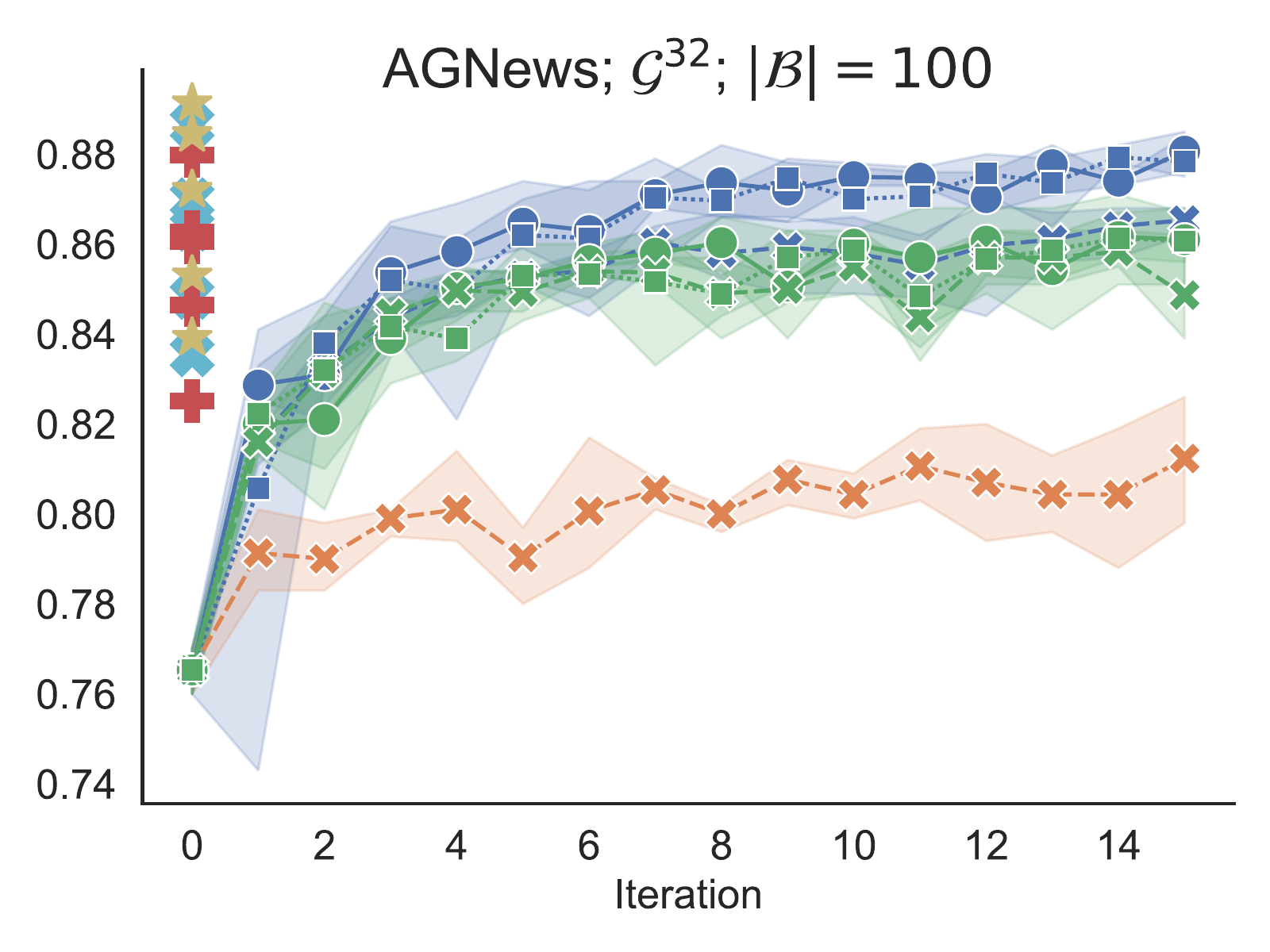}
}
\\
\hspace{-.2cm}\subfloat{
\includegraphics[width=.25\linewidth,height=0.2\textwidth]{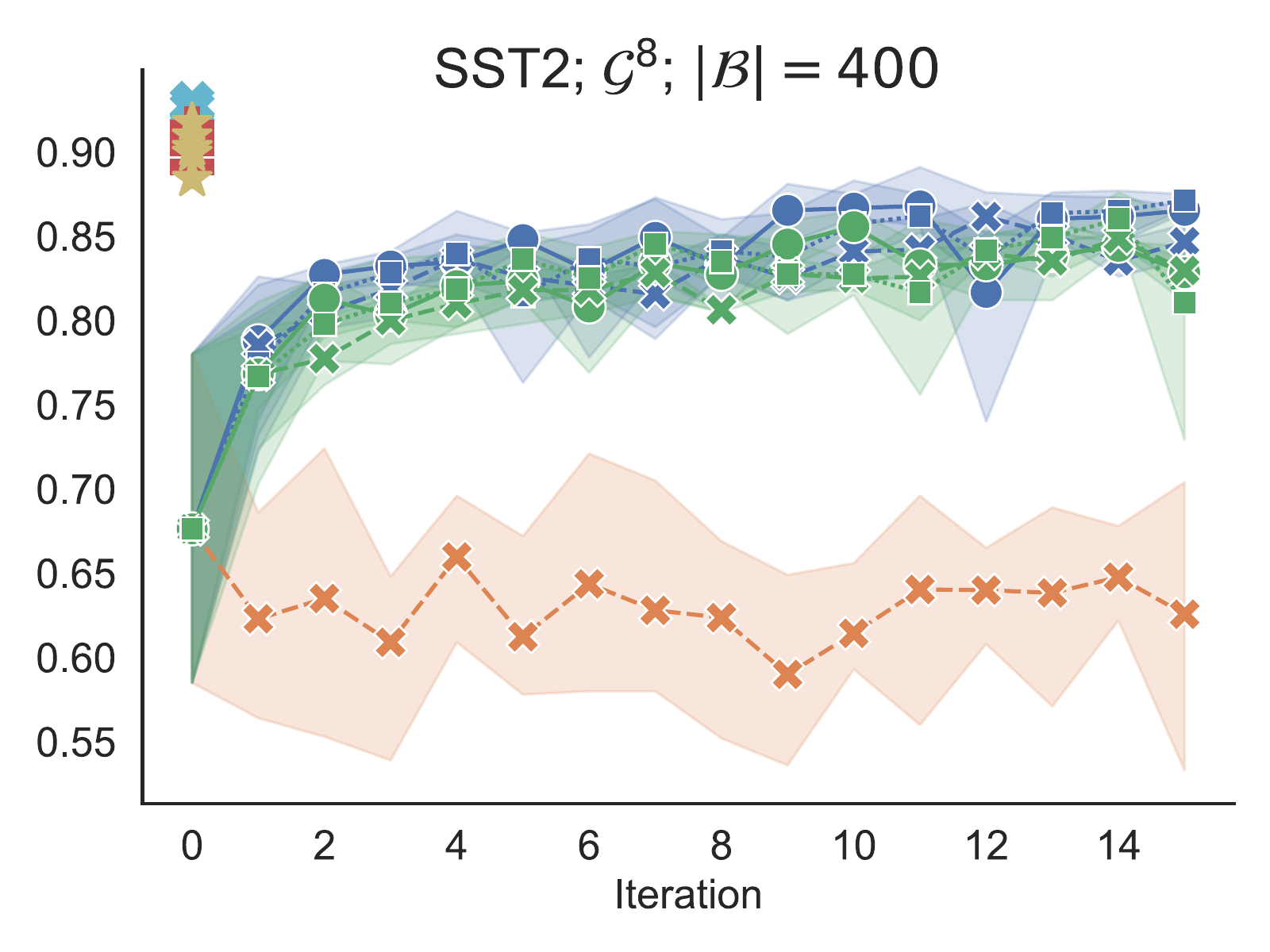}
}
\hspace{-.2cm}\subfloat{
\includegraphics[width=.25\linewidth,height=0.2\textwidth]{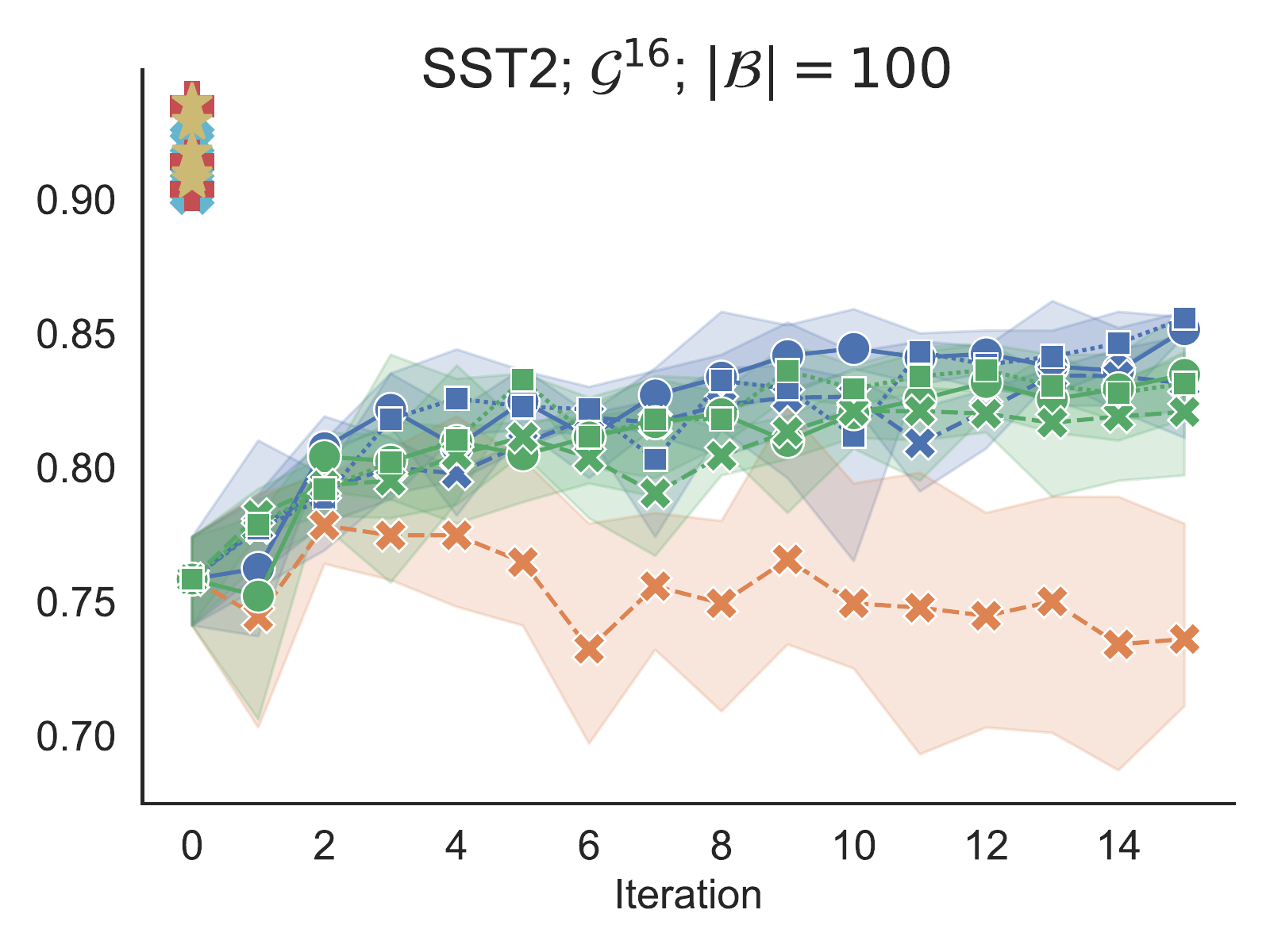}
}
\hspace{-.2cm}\subfloat{
\includegraphics[width=.25\linewidth,height=0.2\textwidth]{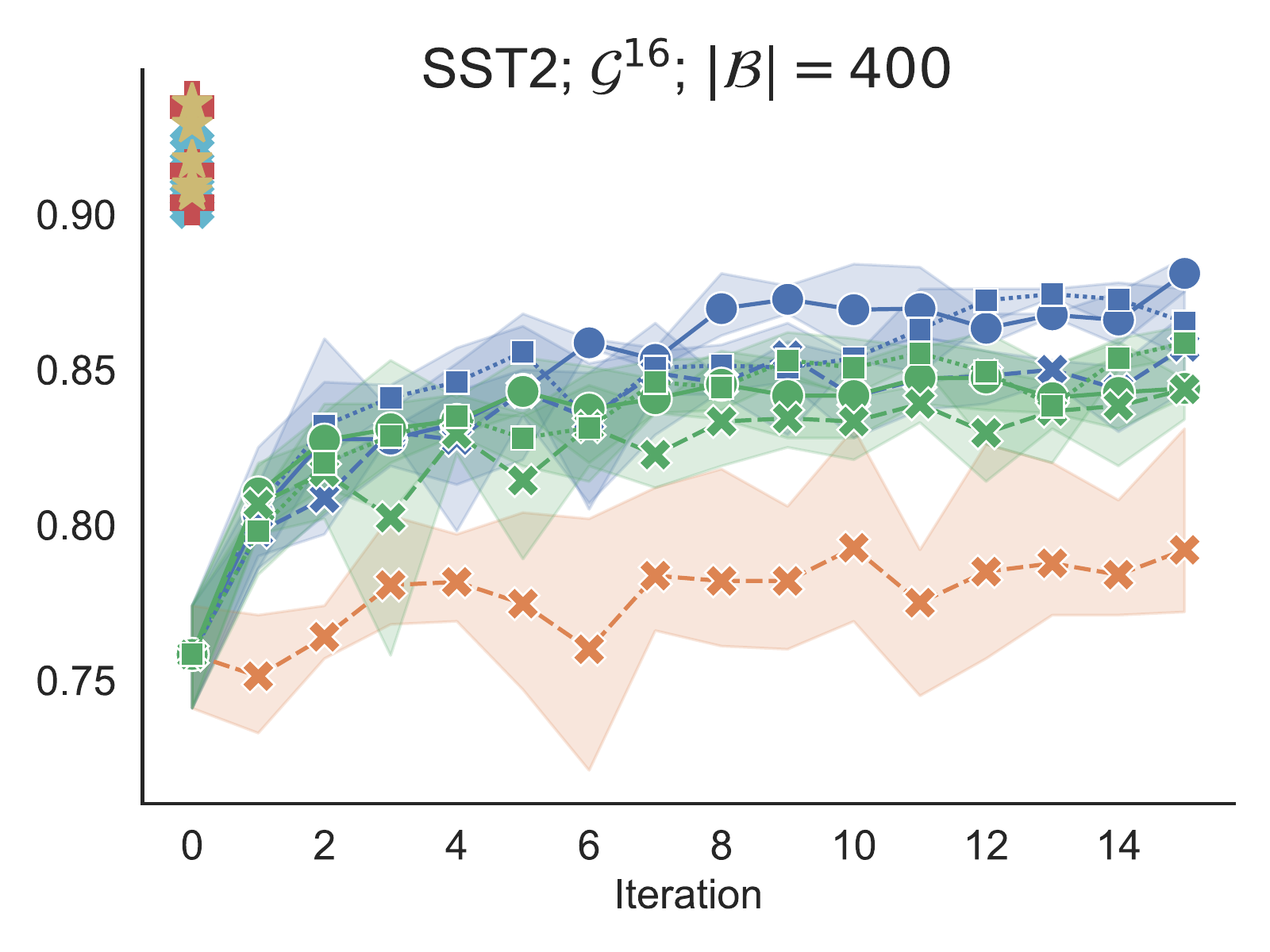}
}
\hspace{-.2cm}\subfloat{
\includegraphics[width=.25\linewidth,height=0.2\textwidth]{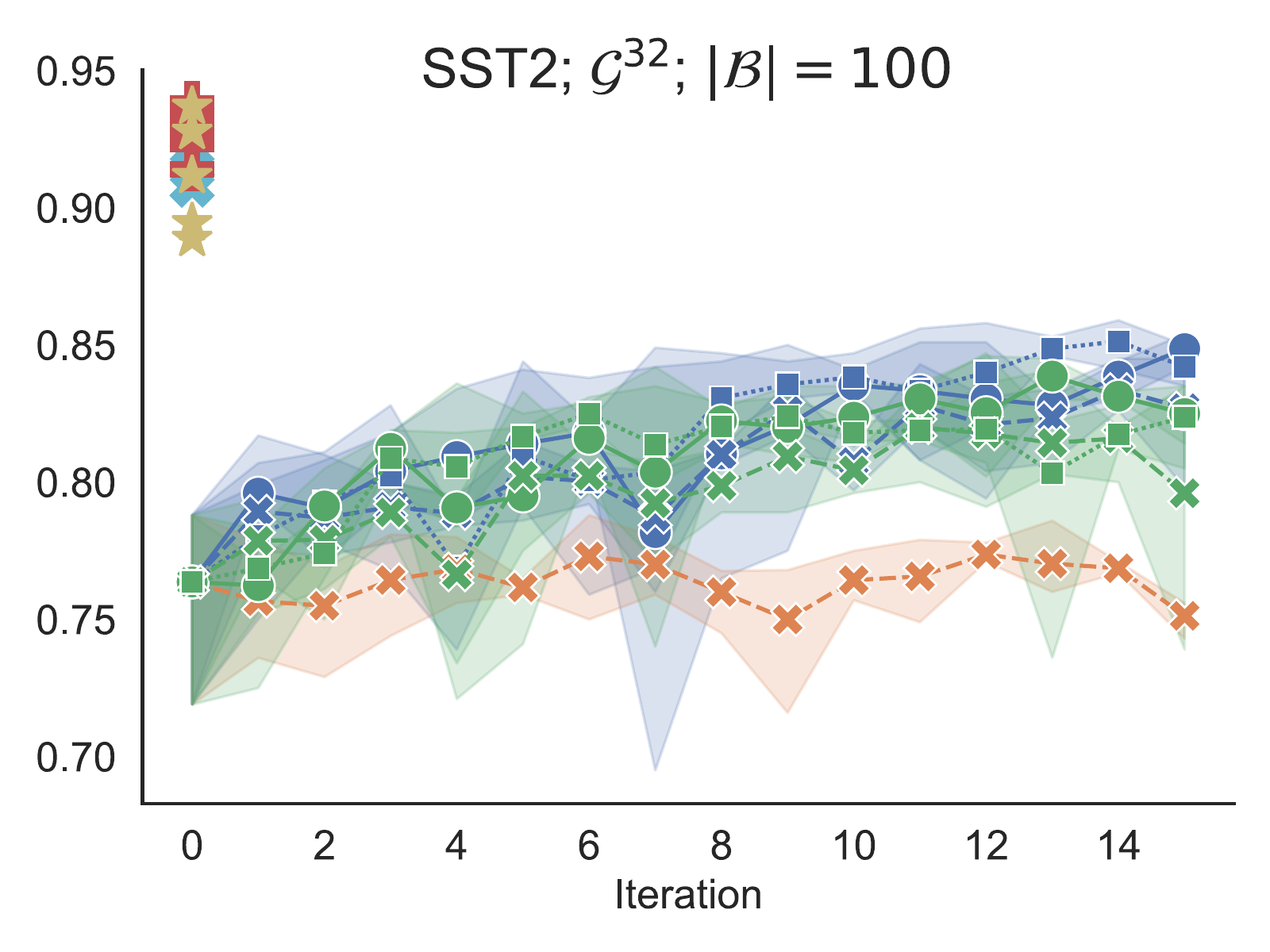}
}
\\
\hspace{-.2cm}\subfloat{
\includegraphics[width=.25\linewidth,height=0.2\textwidth]{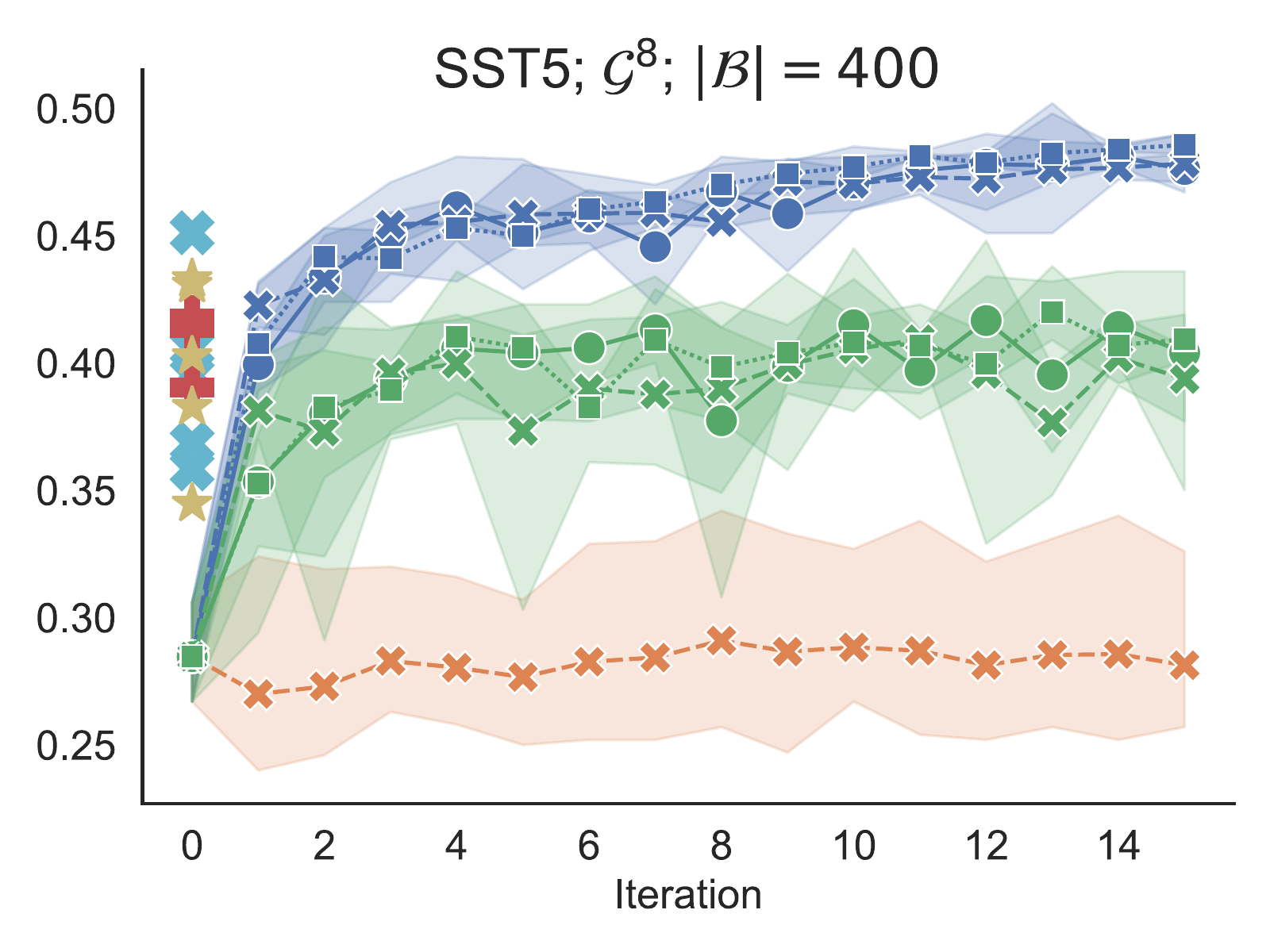}
}
\hspace{-.2cm}\subfloat{
\includegraphics[width=.25\linewidth,height=0.2\textwidth]{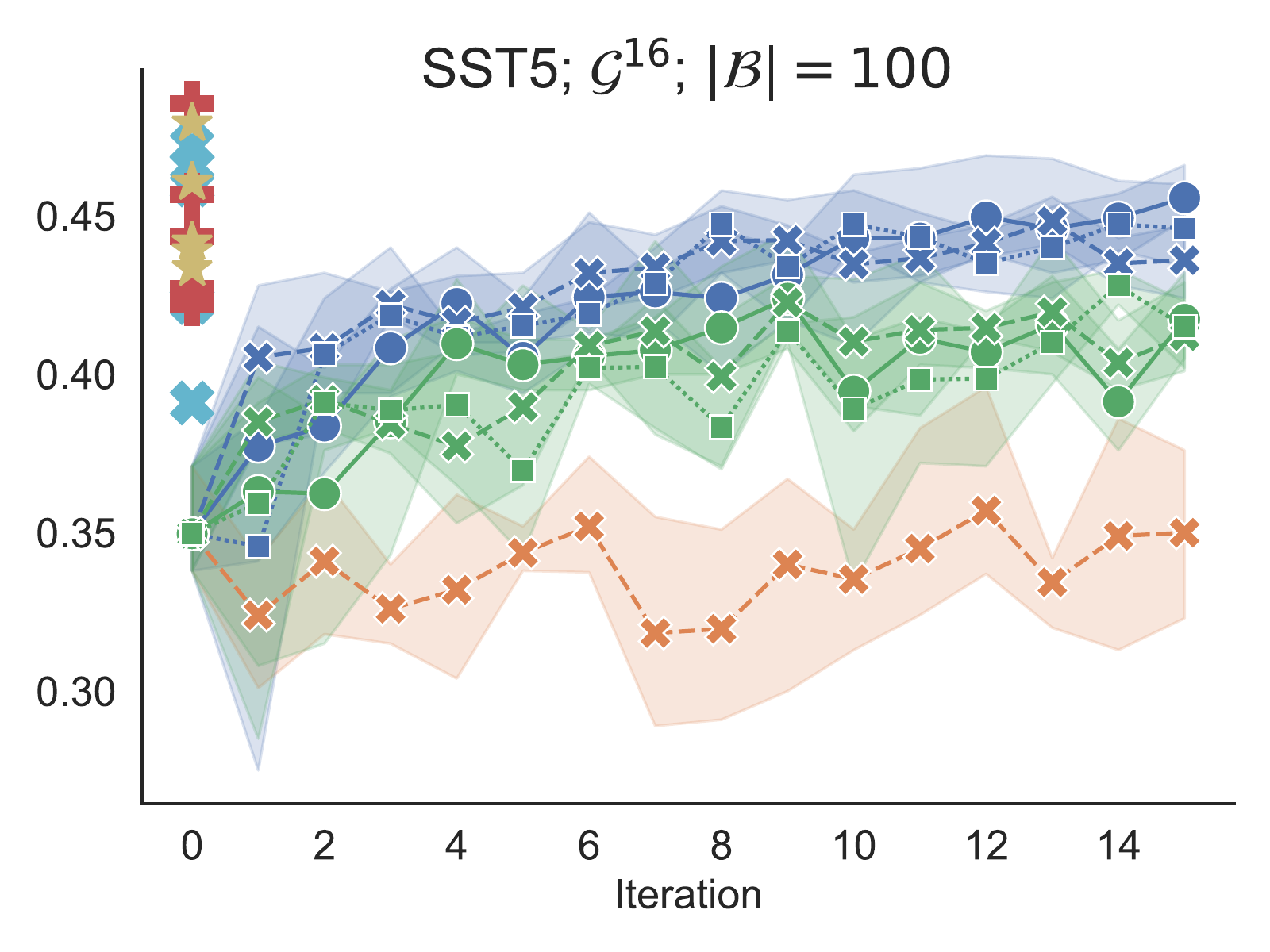}
}
\hspace{-.2cm}\subfloat{
\includegraphics[width=.25\linewidth,height=0.2\textwidth]{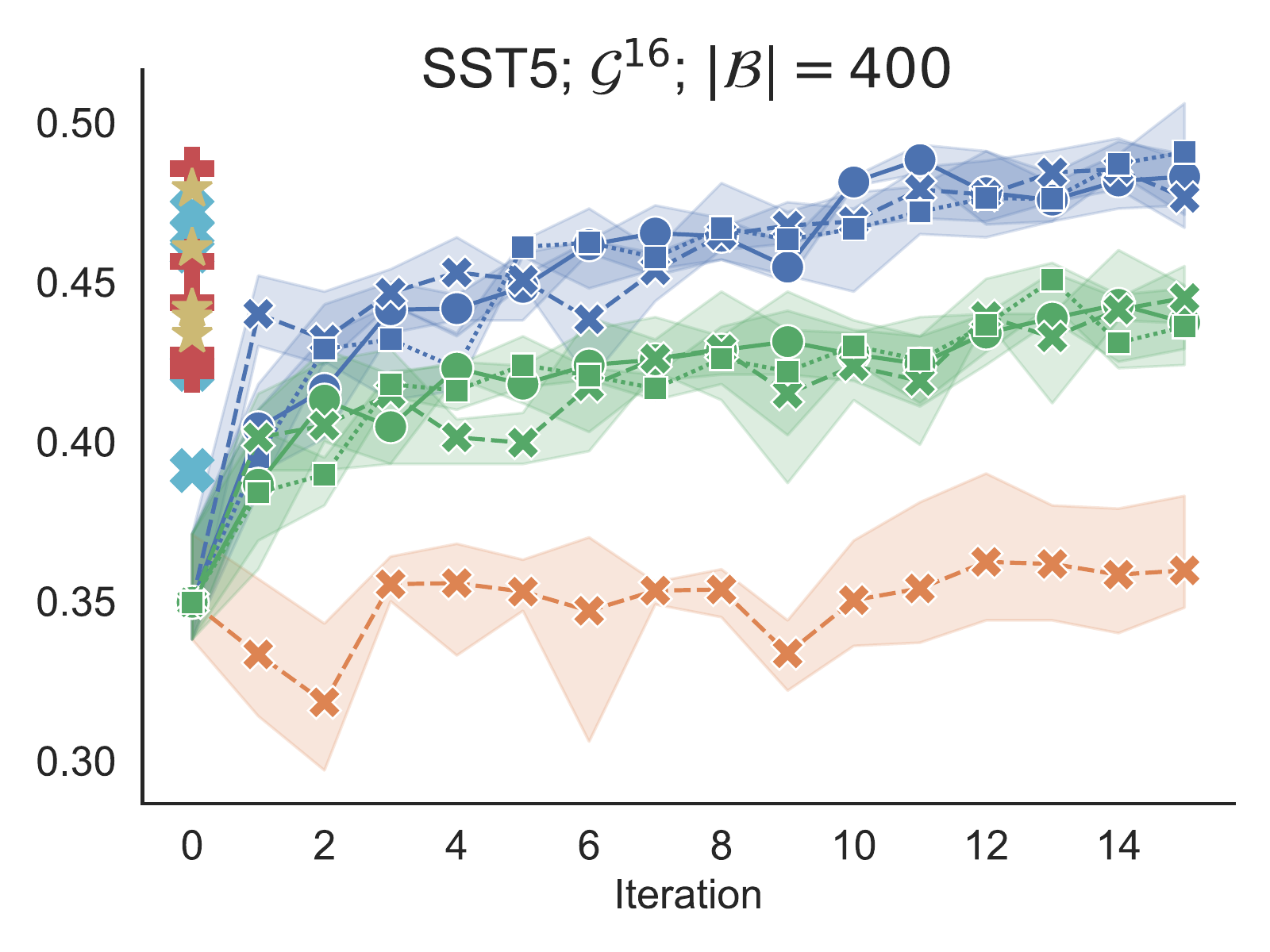}
}
\hspace{-.2cm}\subfloat{
\includegraphics[width=.25\linewidth,height=0.2\textwidth]{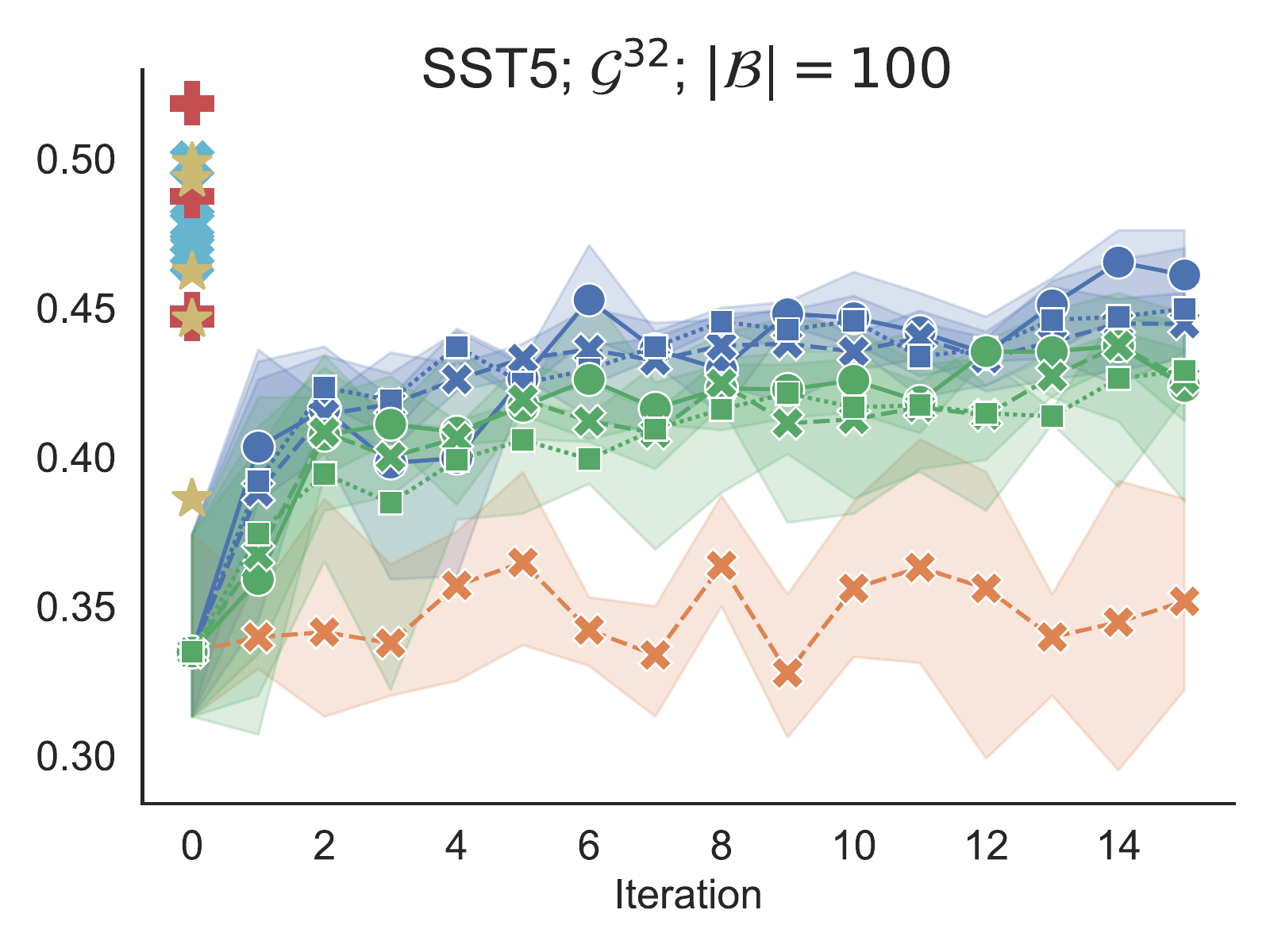}
}
\\
\hspace{-.2cm}\subfloat{
\includegraphics[width=.25\linewidth,height=0.2\textwidth]{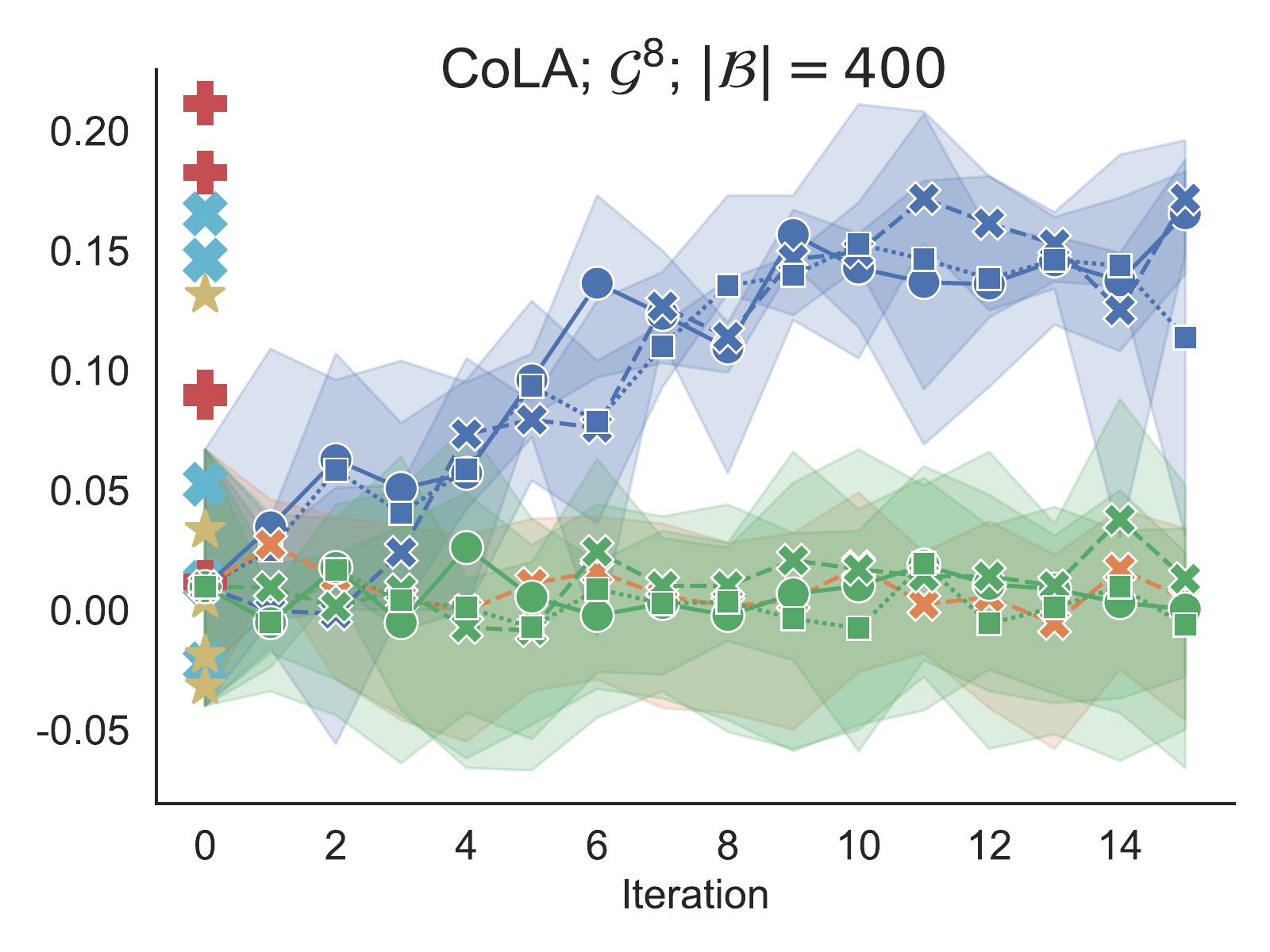}
}
\hspace{-.2cm}\subfloat{
\includegraphics[width=.25\linewidth,height=0.2\textwidth]{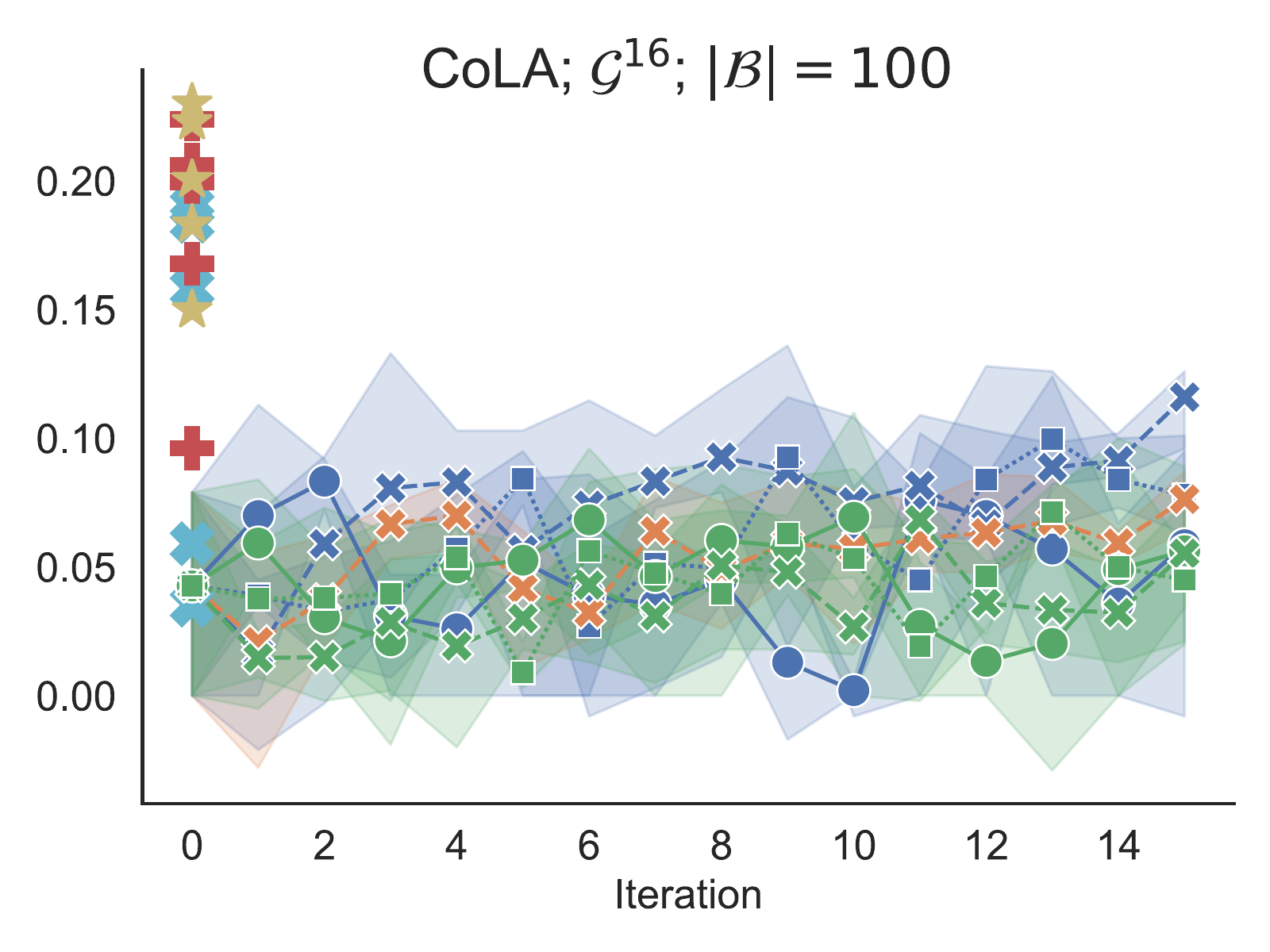}
}
\hspace{-.2cm}\subfloat{
\includegraphics[width=.25\linewidth,height=0.2\textwidth]{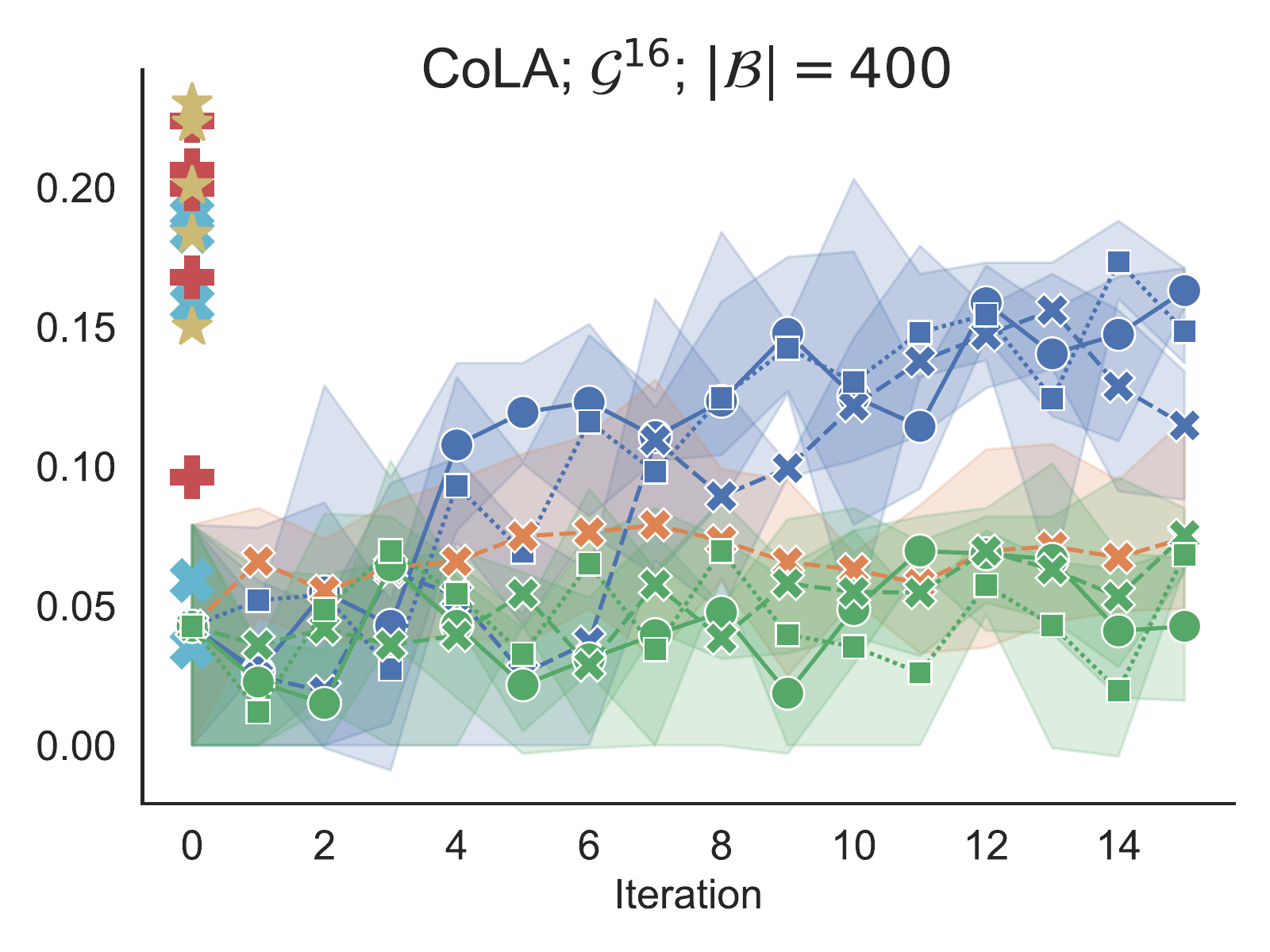}
}
\hspace{-.2cm}\subfloat{
\includegraphics[width=.25\linewidth,height=0.2\textwidth]{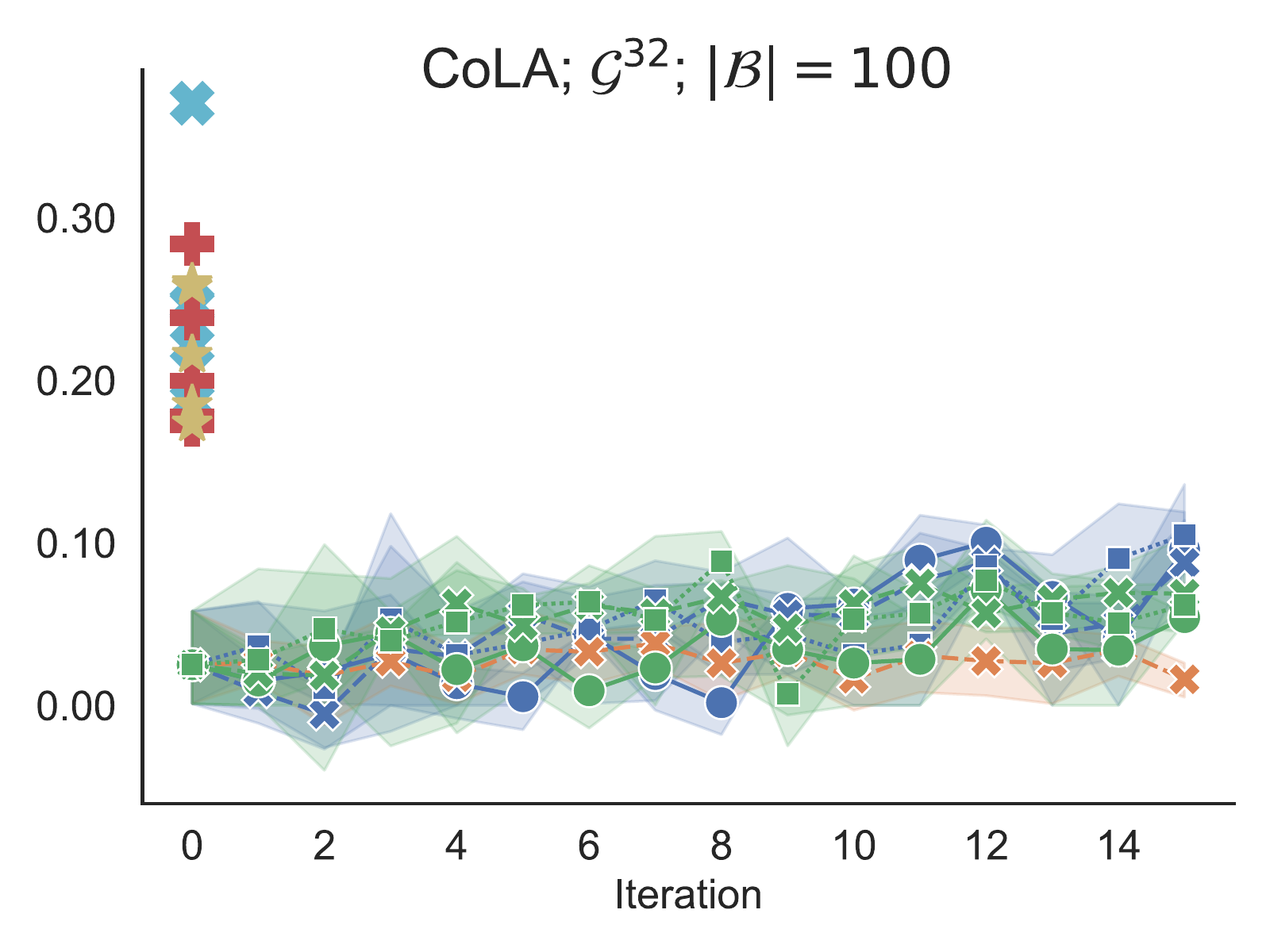}
}
\\
\hspace{-.2cm}\subfloat{
\includegraphics[width=.25\linewidth,height=0.2\textwidth]{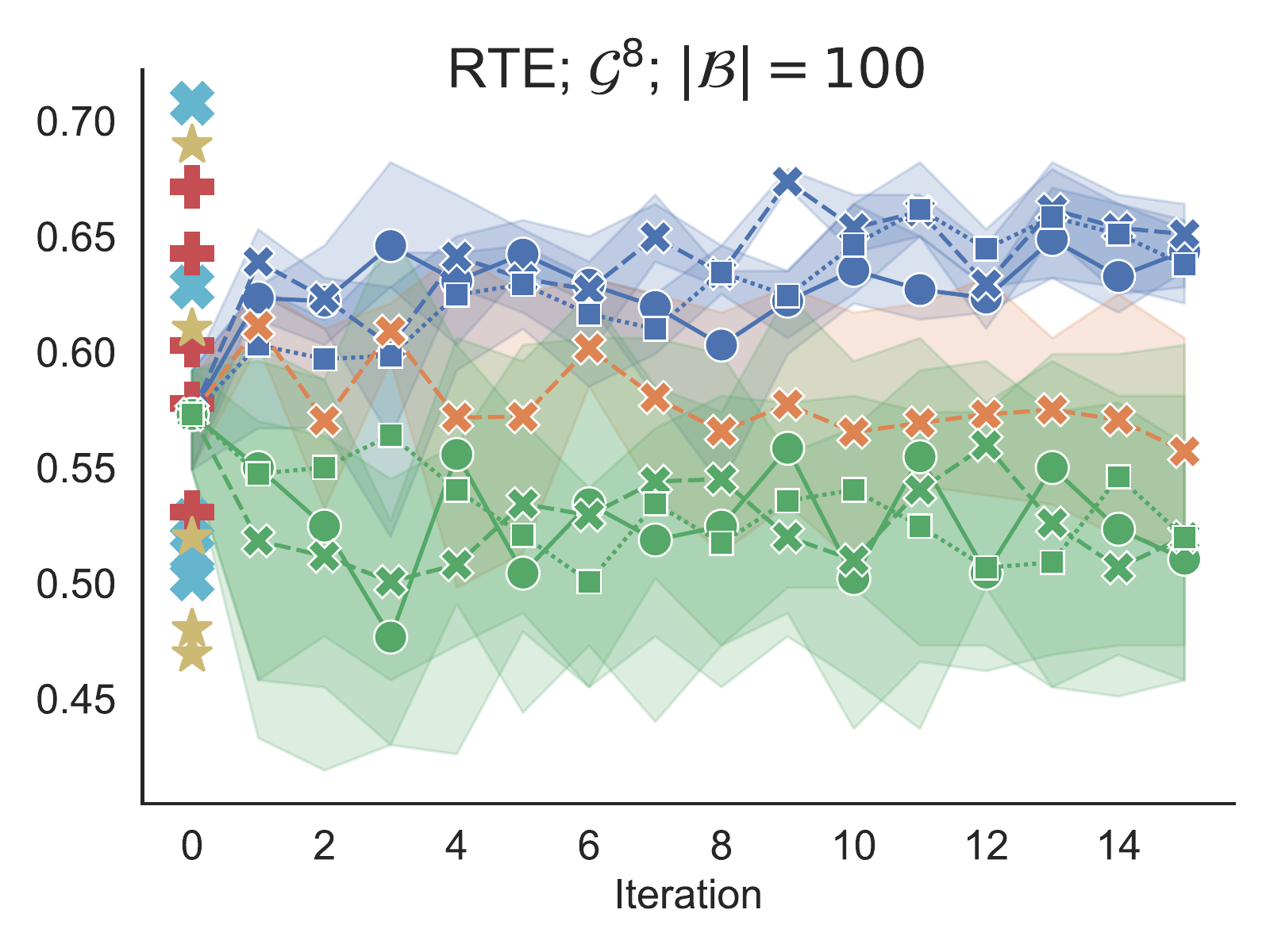}
}
\hspace{-.2cm}\subfloat{
\includegraphics[width=.25\linewidth,height=0.2\textwidth]{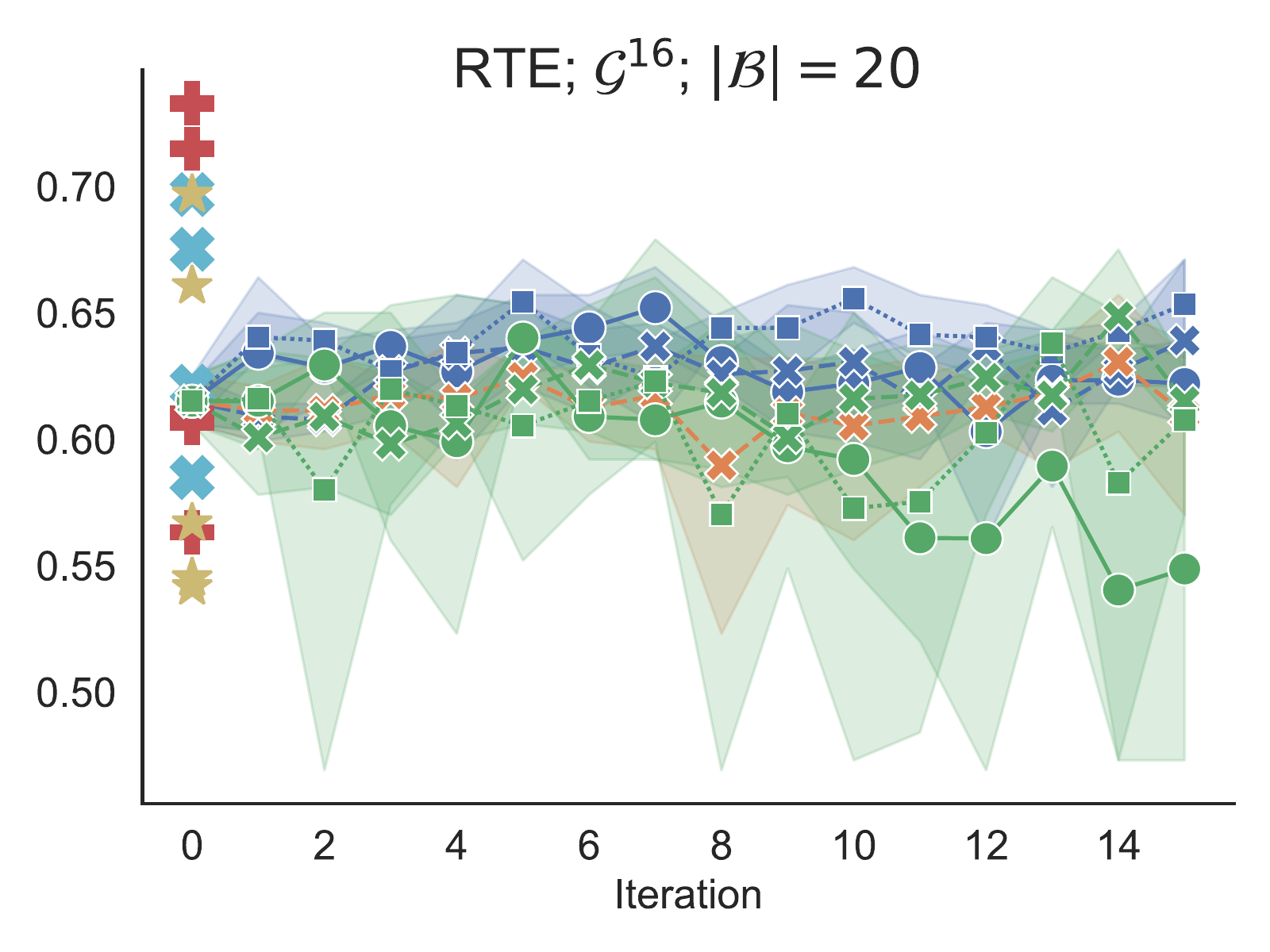}
}
\hspace{-.2cm}\subfloat{
\includegraphics[width=.25\linewidth,height=0.2\textwidth]{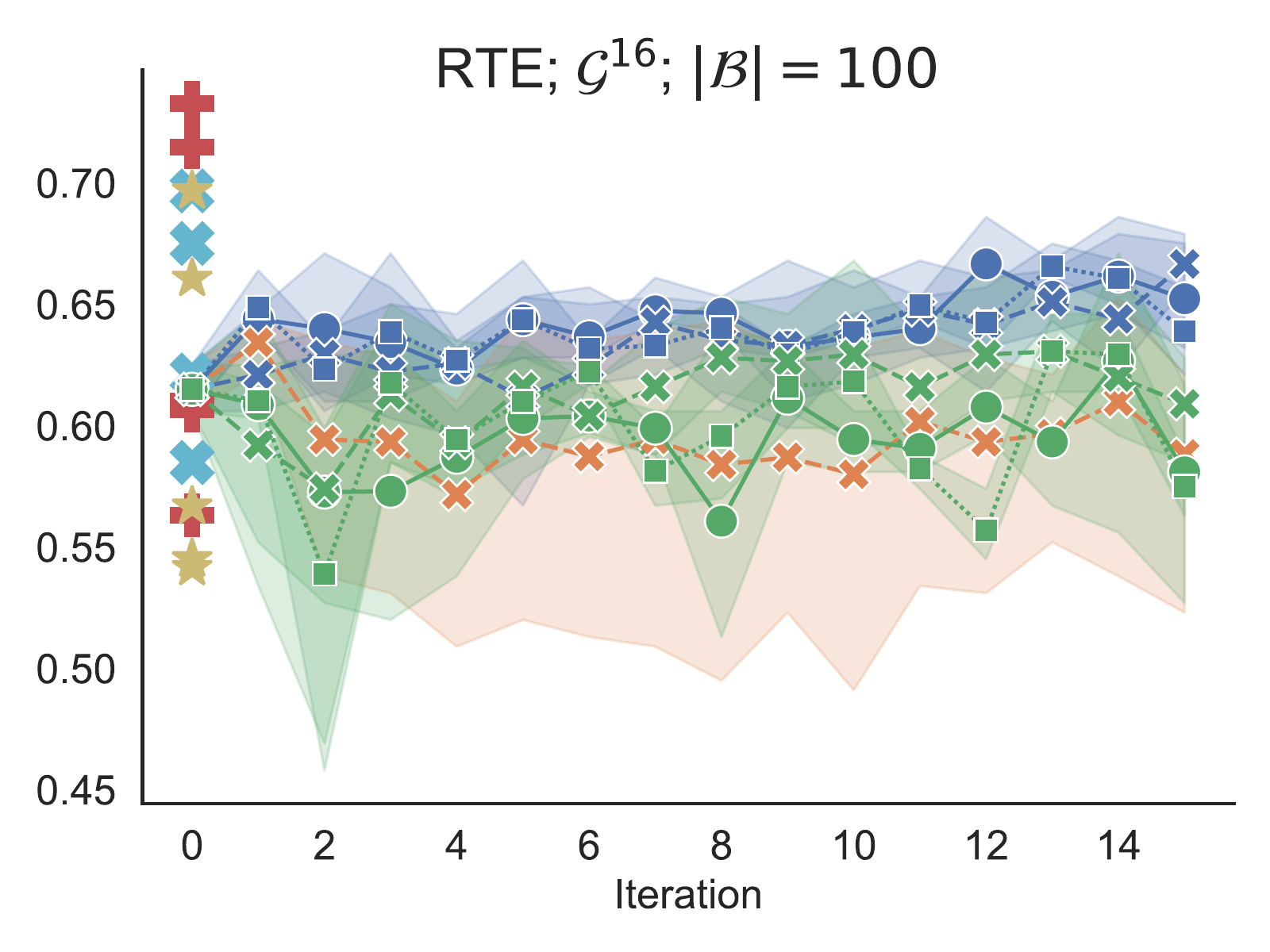}
}
\hspace{-.2cm}\subfloat{
\includegraphics[width=.25\linewidth,height=0.2\textwidth]{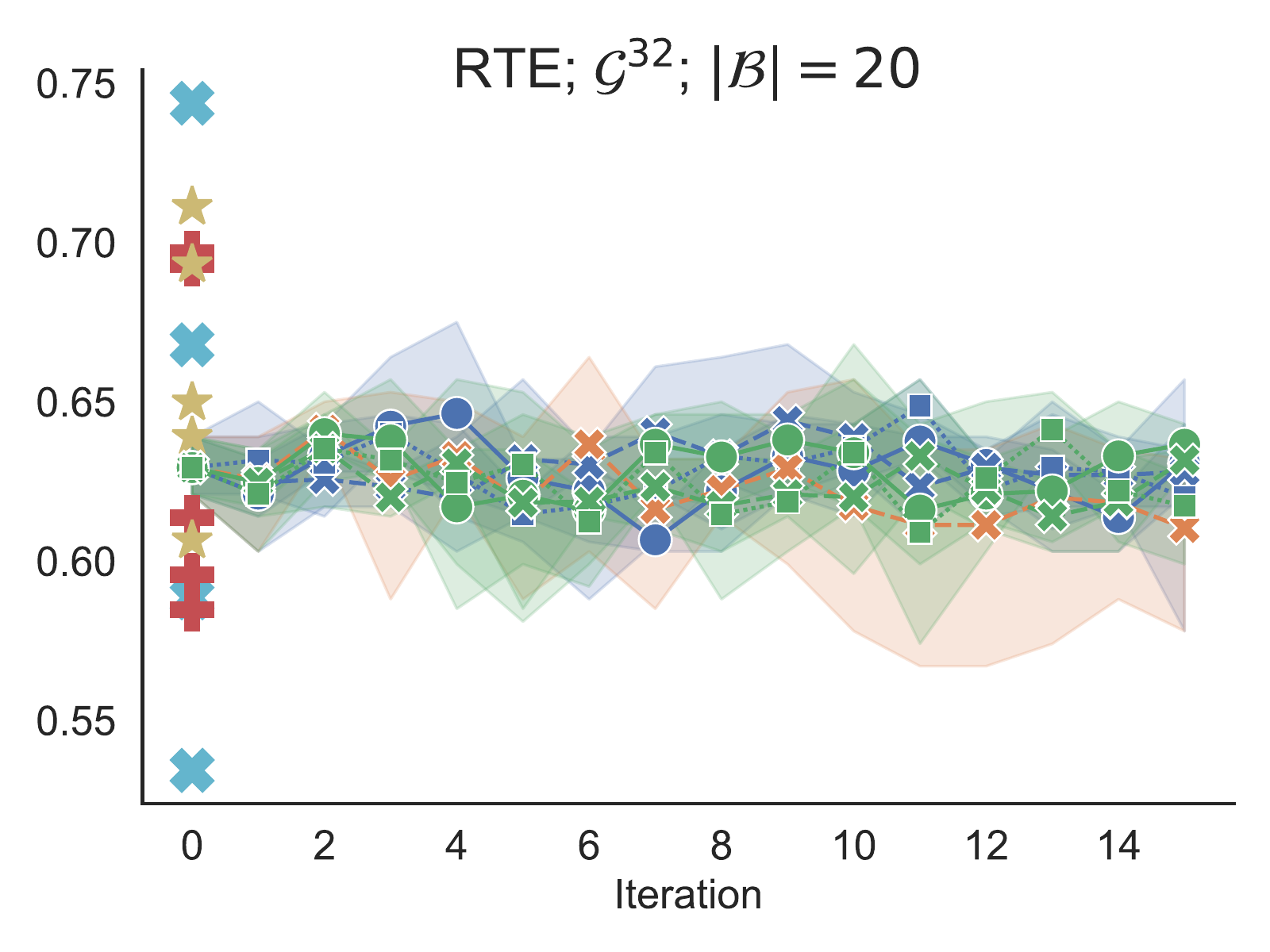}
}

\caption{
  Improving $\mathcal{S}$ with
  \textcolor{blue}{active learning (blue)},
  \textcolor{orange}{self training (orange)},
  and
  \textcolor{green}{\md (green)}.
  Free markers at step zero show \mdr performances;
  colors distinguish random seeds.
  Three acquisition functions are:
  Entropy ($\bullet$),
  LeastConfident (\tiny$\blacksquare$\normalsize),
  random sampling (\tiny\XSolidBold\normalsize).
  At iteration $j$, each experiment is repeated
  three times; we show
  mean and standard deviation.
  We evaluate different $|\mathcal{G}|$ and $|\mathcal{B}|$.
}
\figlabel{appendix:completeiteratives}
\end{figure*}

\section{Instance Weighting}
\seclabel{appendix:instanceweighting}
Following \citet{wang-etal-2017-instance},
we associate each example
$(\textbf{x}, \hat{y}, \textbf{l}) \in \mathcal{D}^{j}$
with weight
1-$entropy(\textbf{l})$ when
computing the loss during
training $\mathcal{S}^{j}$.
We can interpret this weight as a measure
of the certainty of the \mdrs ensemble.

\figref{instanceweighting} reports
the performance of $\mathcal{S}$
when using instance weighting,
however, the impacts are less noticeable.


\end{document}